\documentclass{article}

     \usepackage[preprint,nonatbib]{neurips_2020}

\usepackage[utf8]{inputenc} 
\usepackage[T1]{fontenc}    
\usepackage{hyperref}       
\usepackage{url}            
\usepackage{booktabs}       
\usepackage{amsfonts}       
\usepackage{nicefrac}       
\usepackage{microtype}      

\usepackage{graphicx}
\usepackage{comment}
\usepackage{amsmath,amssymb} 

\usepackage[multidot]{grffile}[=v1]
\usepackage{amsmath,amssymb} 

\usepackage{array}
\usepackage{tabu,multirow}
\usepackage{booktabs}
\usepackage{diagbox}
\usepackage{caption}
\usepackage{subcaption}
\usepackage[percent]{overpic}
\usepackage{xspace}
\usepackage{makecell}
\usepackage[table]{xcolor}

\usepackage[acronym]{glossaries}

\newcommand{\imtoim}{image-to-image}
\newcommand{\pargan}{ParGAN}

\newcommand{\domA}{X}
\newcommand{\domB}{Y}
\newcommand{\domPar}{P}
\newcommand{\sampA}{x}
\newcommand{\sampB}{y}
\newcommand{\sampP}{p}
\newcommand{\genAB}{G}
\newcommand{\genBA}{F}
\newcommand{\discB}{D_Y}
\newcommand{\discA}{D_X}
\newcommand{\genParAB}{G^P}
\newcommand{\genParBA}{F^P}
\newcommand{\discParB}{D^P_Y}
\newcommand{\discParA}{D^P_X}

\newcommand{\summertowinter}[0]{\textit{summer}$\rightarrow$\textit{winter}}
\newcommand{\summertosoftrain}[0]{\textit{summer}$\rightarrow$\textit{softrain}}
\newcommand{\summertofog}[0]{\textit{summer}$\rightarrow$\textit{fog}}
\newcommand{\summertodawn}[0]{\textit{summer}$\rightarrow$\textit{dawn}}

\newcommand{\summertofall}[0]{\textit{summer}$\rightarrow$\textit{fall}}
\newcommand{\summertospring}[0]{\textit{summer}$\rightarrow$\textit{spring}}

\newcommand{\summertosunset}[0]{\textit{summer}$\rightarrow$\textit{sunset}}
\newcommand{\summertonight}[0]{\textit{summer}$\rightarrow$\textit{night}}
\newcommand{\summertowinternight}[0]{\textit{summer}$\rightarrow$\textit{winter}+\allowbreak\textit{night}}
\newcommand{\summertowinternightsoftrain}[0]{\textit{summer}$\rightarrow$\allowbreak\textit{winter}\allowbreak+\textit{night}+\textit{softrain}}
\newcommand{\daytonight}[0]{\textit{day}$\rightarrow$\textit{night}}

\newcommand{\summertodawnsunsetnight}[0]{\textit{summer}$\rightarrow$\allowbreak\textit{dawn}\allowbreak+\textit{sunset}\allowbreak+\textit{night}}

\definecolor{metric_step0}{RGB}{87,187,138}
\definecolor{metric_step1}{RGB}{129,204,167}
\definecolor{metric_step2}{RGB}{171,221,197}
\definecolor{metric_step3}{RGB}{213,238,226}
\definecolor{metric_step4}{RGB}{255,255,255}

\makeatletter
\DeclareRobustCommand\onedot{\futurelet\@let@token\@onedot}
\def\@onedot{\ifx\@let@token.\else.\null\fi\xspace}

\def\eg{\emph{e.g}\onedot}

\def\ie{\emph{i.e}\onedot}

\newcommand{\synthia}{\textit{SYNTHIA}}
\newcommand{\init}{\textit{INIT}}

\newcommand{\bookCovers}{\textit{Book-Covers}}

\newcommand{\FID}{\textit{FID}}
\newcommand{\LPIPS}{\textit{LPIPS}}

\title{\pargan: Learning Real Parametrizable Transformations}

\author{%
  Diego Martin\\
  Google Switzerland\\
  \texttt{martinarroyo@google.com} \\
   \And
 Alessio Tonioni\\
 Google Switzerland\\
 \texttt{alessiot@google.com} \\
 \And
  Federico Tombari\\
  Google Switzerland\\
  Technical University of Munich\\
  \texttt{tombari@google.com} \\
}

\begin{document}

\maketitle

\begin{abstract}
    Current methods for \imtoim{} translation produce compelling results, however, the applied transformation is difficult to control, since existing mechanisms
    are often limited and non-intuitive.
    We propose \pargan{}, a generalization of the cycle-consistent GAN framework to learn image transformations with simple and intuitive controls.
    The proposed generator takes as input both an image and a parametrization of the transformation.
    We train this network to preserve the content of the input image while ensuring that the result is consistent with the given parametrization.
    Our approach does not require paired data and can learn transformations across several tasks and datasets.
    We show how, with disjoint image domains with no annotated parametrization, our framework can create smooth interpolations as well as learn multiple transformations simultaneously.
    
\end{abstract}

\section{Introduction}
\label{sec:introduction}

\newlength{\teaserlength}
\setlength{\teaserlength}{0.12\linewidth}
\begin{figure}[h!]
    \centering
    \setlength{\tabcolsep}{1pt}
    \begin{tabular}{c|c}
    \begin{tabular}{c}
    \includegraphics[width=0.25\linewidth]{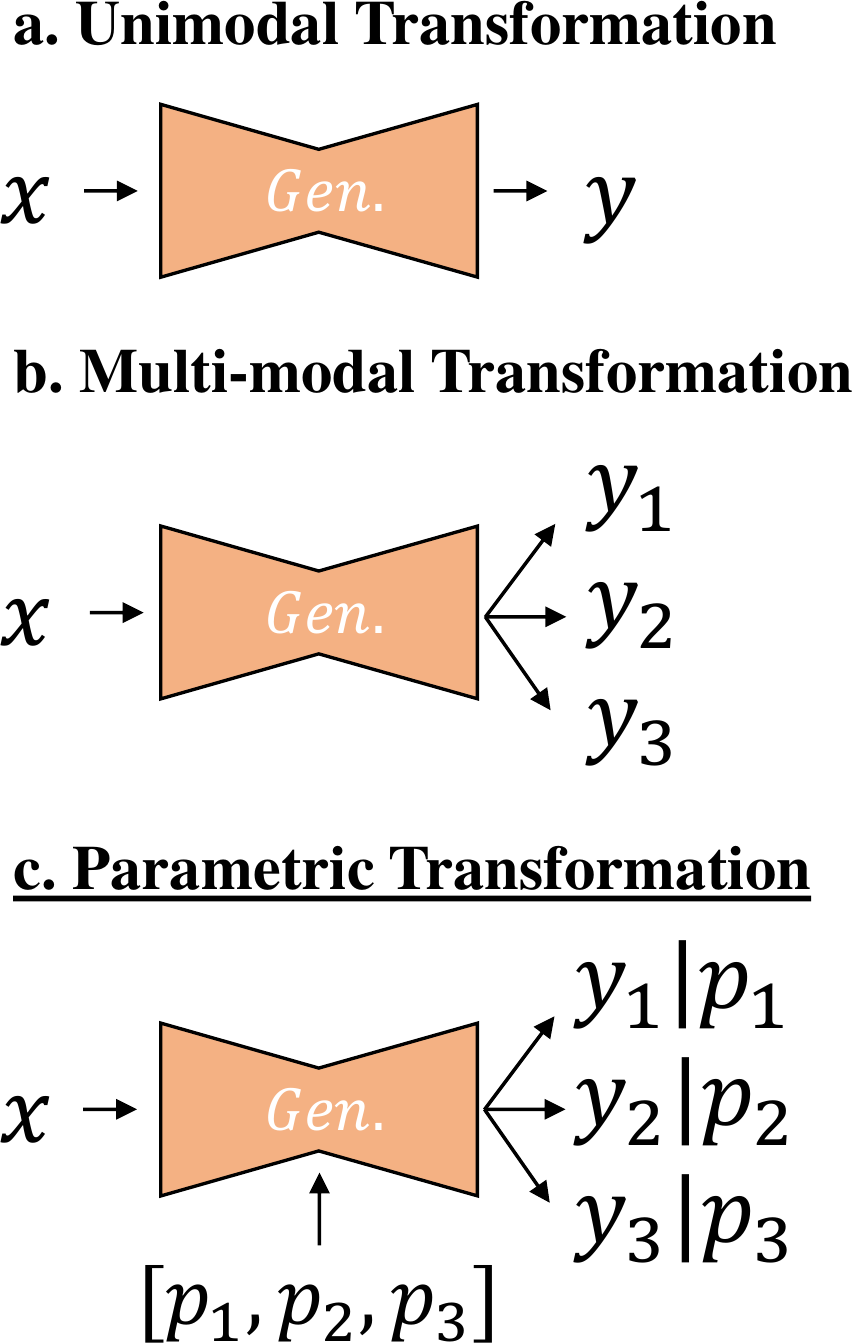}
    \end{tabular} &
    \begin{tabular}{ccccccc}
        \rotatebox{90}{\hspace{2mm}\textbf{\fontsize{7}{8} \daytonight{}}} & \includegraphics[width=\teaserlength]{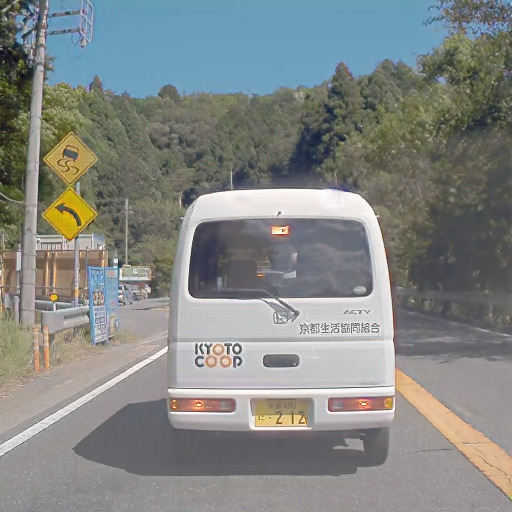} &
        \includegraphics[width=\teaserlength]{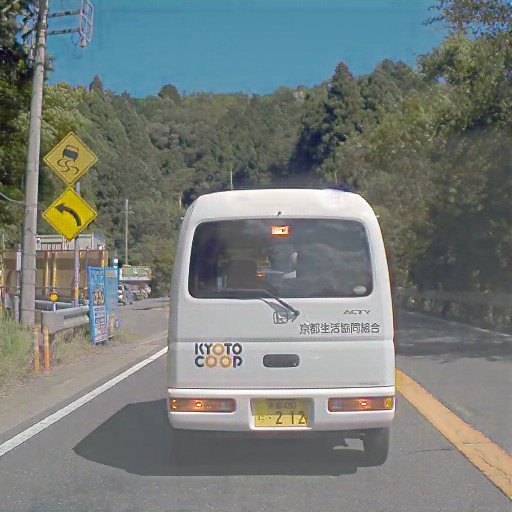} &
        \includegraphics[width=\teaserlength]{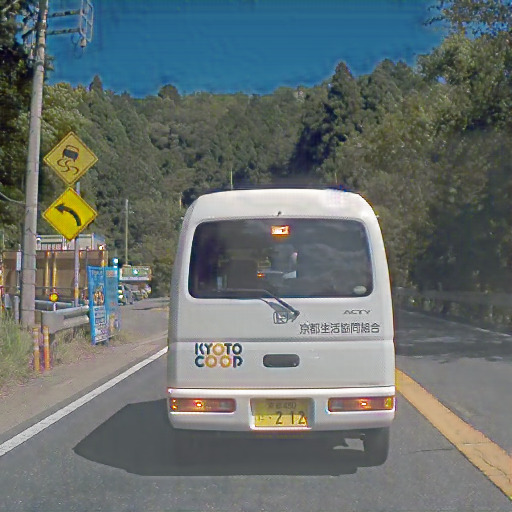} &
        \includegraphics[width=\teaserlength]{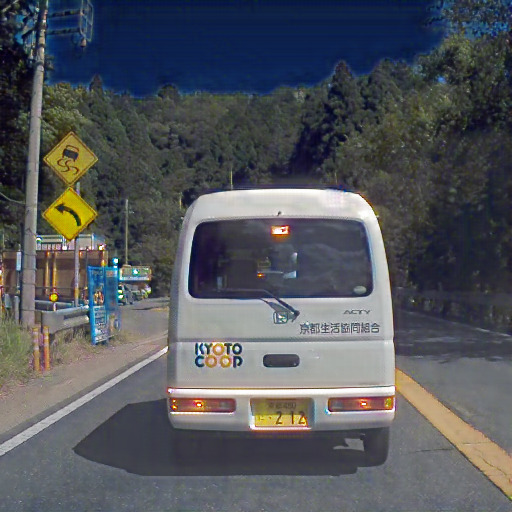} &
        \includegraphics[width=\teaserlength]{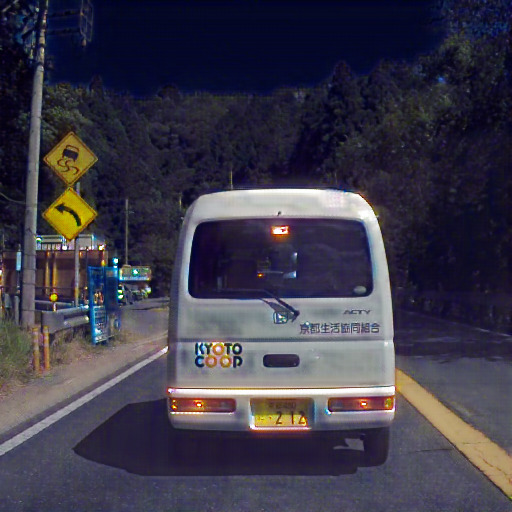} \\
        
        \rotatebox{90}{\hspace{-1mm}\textbf{\fontsize{7}{8} \summertosunset{}}} &\includegraphics[width=\teaserlength]{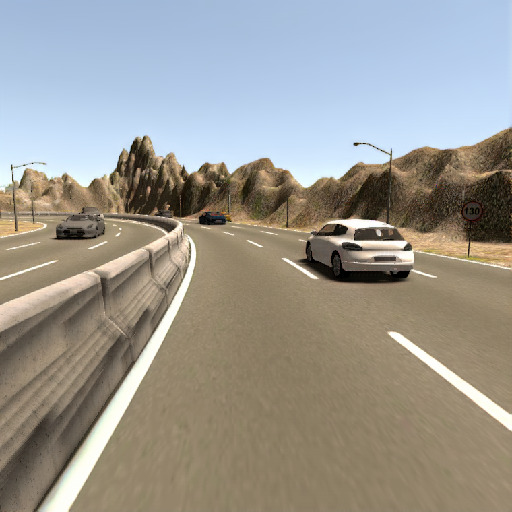} &
        \includegraphics[width=\teaserlength]{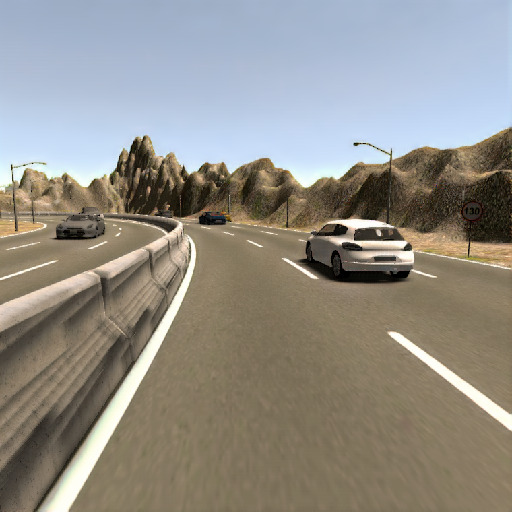} &
        \includegraphics[width=\teaserlength]{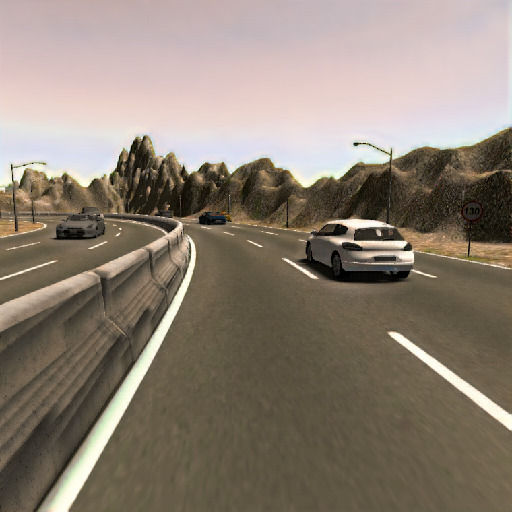} &
        \includegraphics[width=\teaserlength]{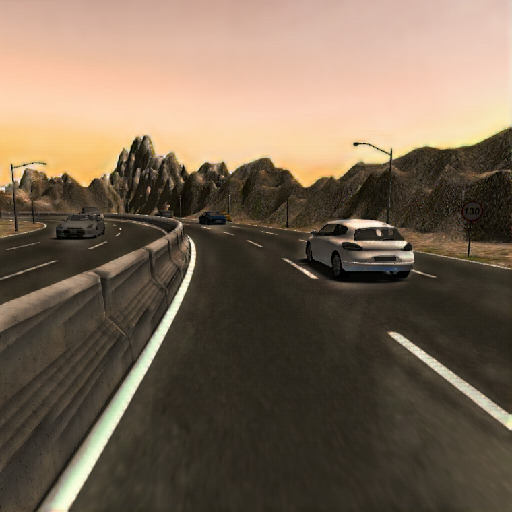} &
        \includegraphics[width=\teaserlength]{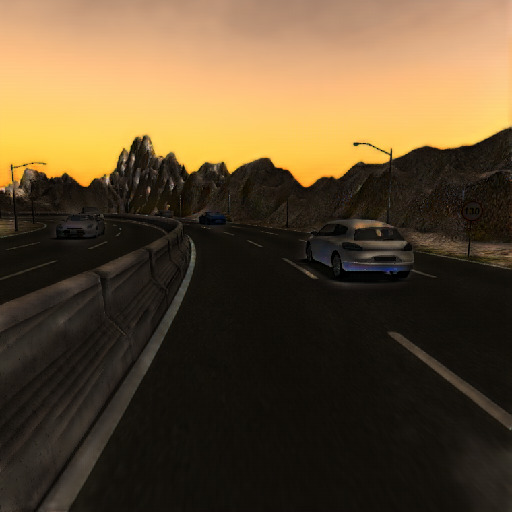} \\[1mm]
        
        \rotatebox{90}{\hspace{-2mm}\textbf{\fontsize{7}{8} \summertowinter{}}} &\includegraphics[width=\teaserlength]{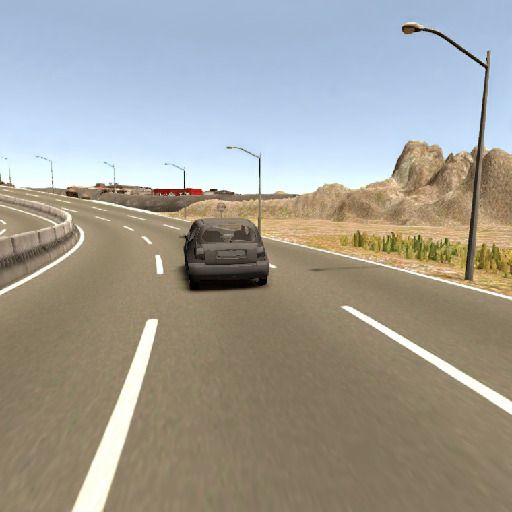} &
        \includegraphics[width=\teaserlength]{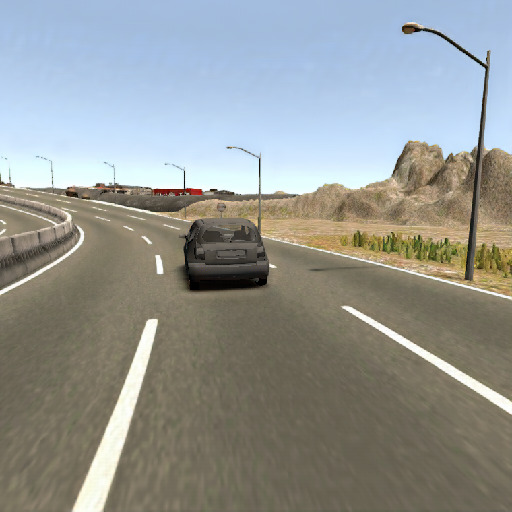} &
        \includegraphics[width=\teaserlength]{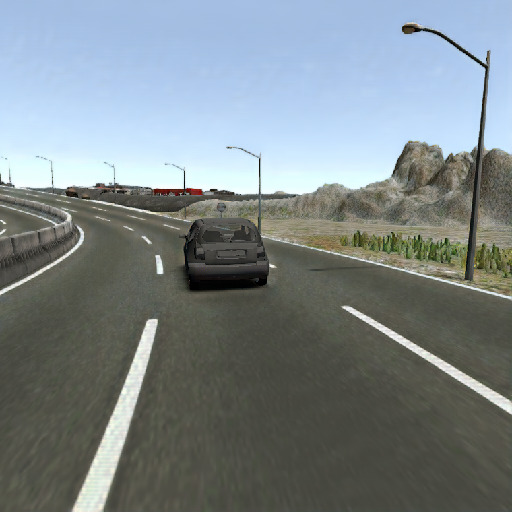} &
        \includegraphics[width=\teaserlength]{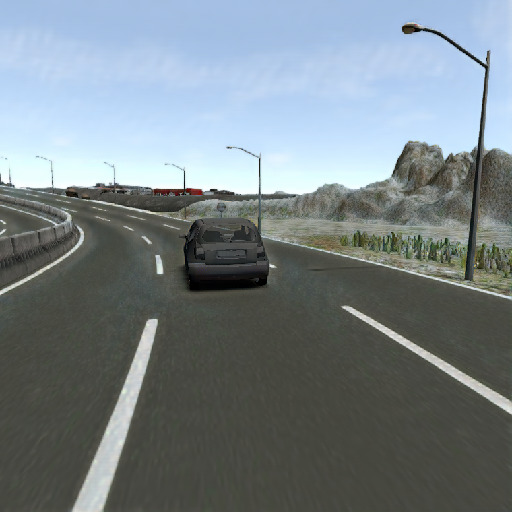} &
        \includegraphics[width=\teaserlength]{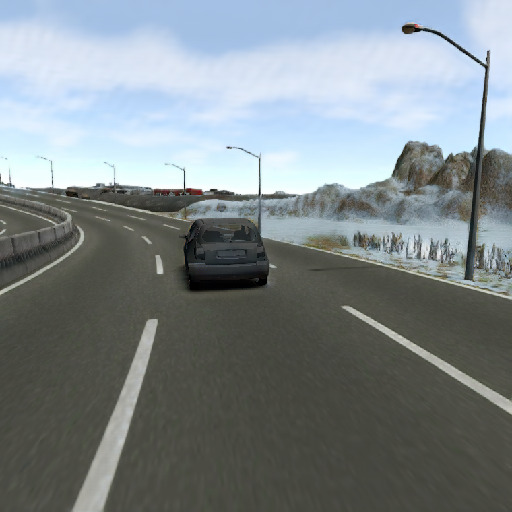} \\
        & {\scriptsize $p=0.00$} & {\scriptsize $p=0.25$} & {\scriptsize $p=0.50$} & {\scriptsize $p=0.75$} & {\scriptsize $p=1.00$} \\
        
    \end{tabular} \\

    \end{tabular}
    \captionof{figure}{
    \textbf{Left}: Unimodal transformations \cite{CycleGAN2017,pix2pix2016} (a) generate a unique image. Multimodal ones (b) often require sampling and lack explicit control. The output $y$ of our proposal (c) is driven by both the input $x$ and an \emph{intuitive parametrization} $p$. 
    \textbf{Right}: results of our method on different tasks parametrized by a scalar. Each row is generated from the same input $x$ and increasing values of $p$.
    }
    \label{fig:teaser}
\end{figure}%

Recent developments in Generative Adversarial Networks (GANs) have introduced a general framework to learn \imtoim{} mapping functions between two disjoint input and output domains (\eg, two sets of images) \cite{pix2pix2016,CycleGAN2017}.
While some computer vision tasks can be directly addressed by these models (\eg, colorization, image to segmentation mask, style transfer\dots), others cannot.
For example, transforming an outdoor image of a sunny day into the same scene under different weather conditions is certainly a useful application as a data augmentation tool for training autonomous driving systems. 
However, solving this task would require to learn a one-to-many mapping function or, even better, a continuous parametrizable transformation, since weather conditions can hardly be categorized as discrete classes \cite{hoffman2014continuous}. 
Additional examples of problems with similar continuous parametrizations might be learning the illumination changes caused by a moving light source on a certain object or scene, changing the focus point between foreground and background, etc. 

The most common GAN-based architectures for \imtoim{} translation \cite{pix2pix2016,CycleGAN2017} unfortunately cannot handle these tasks, as they learn a unimodal mapping function and do not provide any mechanism to explicitly control the transformation process.
Recent efforts have extended these formulations to learn multimodal outputs \cite{zhu2017toward,huang2018multimodal,DBLP:journals/corr/abs-1808-00948} by relying on latent space sampling to produce multiple images with the same content but different styles. 
Unfortunately, the sampling process is often not directly interpretable and hardly scales to practical applications. 
For example, given the sunny image from the aforementioned use case, it is difficult to control the generator to synthesize a range of images under different rainy conditions, or to explicitly control the amount of ``raininess''.

To address these issues we have developed a generalization of the GAN-based unpaired \imtoim{} translation framework that allows it to learn a parametrizable transformation between two or more domains without the need of explicit supervision.
We have designed our framework, which we named \pargan{}, for explicit control over the changes in the transformation: our generator network accepts as input a parameterization of the transformation encoded in real-world magnitudes (\eg, position in pixels, intensity change as a percentage\dots). These values are easy to understand and control by a human user.
The discriminator also receives this parametrization in addition to the images, in order to judge whether the generated image belongs to the target domain and whether it is consistent with the value of $p$.
Both components of our GAN are therefore forced to explicitly take into account the parametrization of the transformation when generating or discriminating samples.  

Feeding a parameterization of the transformation provides an easy and tunable solution to generate both multi-modal outputs (\ie, same input image with different parame\-tri\-zations) as well as continuous and human interpretable transformations.
Our framework can be applied when there exist annotations for a continuous parameterization of the transformation (\ie, supervised data), as well as in the case of only domain-level categorical parameterization (\ie, soft parameterization).
For example, given two disjoint image domains, we can use this information as soft supervision and define a categorical parameterization of the transformation between them (\ie, a single parameter with a value of either 0 or 1 according to the domain). 
Our framework can learn to map images from one domain to the other, as well as to smoothly interpolate between them according to the provided parameterization without the need for samples from the intermediate domains at training time.
The same principle can be applied to transformations between a source domain and multiple targets. 
In this scenario a single generator is able to smoothly interpolate between the source and each one of the target domains independently (\textit{disentanglement}), while also obtaining promising results when combining multiple transformations (\textit{mixing}). 

In the left part of \autoref{fig:teaser} we summarize our novel problem formulation (c) with respect to uni-modal (a) and multi-modal (b) \imtoim{} translation networks.
On the right part of \autoref{fig:teaser} we show results on several tasks obtained by our architecture using the soft parameterization described above.
To sum up the novel contributions of our work:
\begin{itemize}
\item We propose a generalization of the unsupervised \imtoim{} translation framework to learn explicitly parametrizable transformations. Existing works either do not provide any control  \cite{CycleGAN2017,pix2pix2016,wang2018pix2pixHD} or rely on hardly interpretable sampling from a latent space  \cite{stylegan,DBLP:journals/corr/ChenDHSSA16,DBLP:conf/nips/ZhuZPDEWS17,huang2018munit}.
\item Our framework is general and can be applied to multiple use cases either with supervision on the transformation parameters or only with soft categorical labels. Other works have addressed only specific tasks with strong prior knowledge \cite{nguyenphuoc2019hologan,shi2019facetoparameter}.
\item Our generator not only learns to replicate the transformation observed at training time, but also to combine and smoothly interpolate multiple transformations.
\end{itemize}

\section{Related Works}
\label{sec:related}
\paragraph{Generative Adversarial Networks} 
The original GAN formulation \cite{NIPS2014_5423} creates samples that resemble those of a target distribution from a random noise vector. 
Since this vector is not interpretable, several approaches have been proposed to improve this aspect.
Chen et al. \cite{DBLP:journals/corr/ChenDHSSA16} map independent dimensions of the input space to different attributes in the output, while \cite{DBLP:journals/corr/MirzaO14} uses known labels as additional inputs to control the generation process.  
StyleGAN \cite{stylegan} proposes to disentangle content and style by using the input noise vector to generate a set of learned affine transformations that control different aspects of the image style at different stages of the network.
In all those works, however, the style depends on a highly dimensional latent space that cannot be easily interpreted. Thus, generating an image with a specific set of features can only be achieved by either brute-force sampling or tuning some preexisting configuration.

\paragraph{Image-to-image translation} 
Works in the \imtoim{} translation literature attempt to learn a mapping from an image domain to another (day to night, summer to winter\dots). 
GAN-based works like Pix2Pix \cite{pix2pix2016} or Pix2PixHD \cite{wang2018pix2pixHD} learn this transformation using paired images in the two domains. 
This setup allows for explicit supervision by directly comparing the generated and expected images at a pixel or feature level, achieving very realistic results. However, obtaining paired datasets is often difficult and for certain tasks outright impossible.
Cyclic-consistent architectures such as CycleGAN \cite{CycleGAN2017}, DiscoGAN \cite{kim2017learning} or CyCADA \cite{DBLP:journals/corr/abs-1711-03213} remove this constraint, enabling training with unpaired data in two different domains. Since these networks learn a uni-modal mapping by design, it is difficult to add variability or control the output. 
The common approach of feeding random noise does not work for these methods, as the networks tend to ignore it \cite{pix2pix2016}.
Multimodal \imtoim{} translation approaches, such as \cite{DBLP:conf/nips/ZhuZPDEWS17,huang2018multimodal,zhu2017toward,DBLP:journals/corr/abs-1808-00948}, have been recently proposed to obtain multiple outputs from a single input.
They implement variability by sampling from a latent space, a parameterization not easily controllable or interpretable without starting from some preexisting configuration.
StarGAN \cite{choi2018stargan} learns multiple image transformations using a single model by using attribute labels for guidance. Contrary to our approach, this work does not explore the interpolation between source and target domains. Additionally, it only focuses on discrete attribute transformations, whereas we also explore parametrized transformations over a continuous domain.

\paragraph{Interpretable Parameterizations} 
Recent approaches have attempted to shed light on the interpretability of the latent spaces used to condition generative models.
\cite{DBLP:journals/corr/abs-1903-07291} generates images disentangling content and style by conditioning the former on a semantic segmentation mask and the latter on a reference image. 
Approaches such as \cite{DBLP:journals/corr/AntipovBD17,DBLP:conf/cvpr/WangTLG18} learn a controllable transformation for the face-aging task.
However, they still rely on a categorical parameterization (a discrete set of transformations) while our model smoothly interpolates along a valid range. 
HoloGAN \cite{nguyenphuoc2019hologan} generates multiple viewpoints of the same image content produced by a single random latent vector. 
The network is specifically designed for the 3D rotation task, as it explicitly uses a rigid-body motion layer, thus it cannot be extended to non-geometric transformations.
Recently \cite{shi2019facetoparameter} proposes a network that, given an image of a face, learns to estimate the parameterization required to generate a similar 3D avatar model using a rendering engine. This task specific formulation does not generalize well either.
\cite{DBLP:journals/corr/LuTT17} modifies CycleGAN to increase the resolution of faces conditioned by a one-hot encoded attribute vector (skin color, hair, glasses\dots). This approach is also task-specific and discrete. A similar strategy is followed by \cite{Yu2019a}, where the conditioning parameters are intended as a ``guidance'' mechanism to improve the reconstruction, although realistic outcomes can be obtained by changing the parameters. This approach needs paired data, a requirement we wish to avoid.

Recent works focused on exploring the learned latent space and determine which dimensions alter a characteristic. \cite{DBLP:journals/corr/abs-1907-10786} attempts to find the decision boundary between binary labels, and  \cite{DBLP:journals/corr/abs-1907-07171} learns trajectories that conform to a smooth transformation of some feature (zoom, viewpoint\dots). \cite{Anokhin_2020_CVPR} produces images of high quality and learns a style interpolation, however, it still needs to extract the style code from a given image.
\section{Method}
\label{sec:method}

Given a source image domain $\domA{}$, a target image domain $\domB{}$ and a set of possible parametrizations for a transformation $\domPar{} \subset \mathbb{R}^n$, we aim to implement a \emph{parametrizable} mapping  $\genParAB{}: \domA{},\domPar{} \rightarrow \domB{}$. 
We learn this by means of a neural network that generates a valid sample $\sampB{} \in \domB{}$ by taking as input a source image $\sampA{} \in \domA{}$ and a parametrization for the transformation to apply $\sampP{} \in \domPar{}$.
We rely on two unpaired sets of training samples, $\{\sampA{_i}\}_{i=1}^N$ and $\{\sampB{_j},\sampP{_j}\}_{j=1}^M$, and on adversarial training by means of a parametrization-aware discriminator $\discParB{}: \domB{},\domPar{} \rightarrow [0,1]$ to provide the training signal for $\genParAB{}$.

\subsection{Preliminaries: CycleGAN}
\label{ssec:previous}
Our work builds on top of the cyclic-consistent formulation first introduced by \emph{CycleGAN} \cite{CycleGAN2017}, a GAN architecture designed to learn a mapping function $\genAB{}: \domA{} \rightarrow \domB{}$ between unpaired image domains $\domA{}$ and $\domB{}$. 
To achieve this objective the authors leverage an additional inverse mapping function $\genBA{}: \domB{} \rightarrow \domA{}$ and one discriminator for each image domain, \ie, $\discA{}$ and $\discB{}$. The optimization objective is expressed as:
expressed as:
\vspace{-1mm}
\begin{align}
\begin{split}\label{eq:loss_gan}
    \mathcal{L}_{gan}(\genAB{}, \discB{}, \domA{}, \domB{}) & = \mathbb{E}_{\sampB{} \sim \domB{}}[log \discB{(}\sampB{})] 
    + \mathbb{E}_{\sampA{} \sim \domA{}}[log(1-\discB{}(\genAB{(}\sampA{})))]
\end{split}\\[4pt]
\begin{split}\label{eq:loss_cycle}
    \mathcal{L}_{cyc}(\genAB{}, \genBA{}) & = \mathbb{E}_{\sampA{} \sim \domA{}}[||\genBA{}(\genAB{}(\sampA{}))-\sampA{}||_1]
    + \mathbb{E}_{\sampB{} \sim \domB{}}[||\genAB{}(\genBA{}(\sampB{}))-\sampB{}||_1]
\end{split}\\[4pt]
\begin{split}\label{eq:obj_cyclegan}
    \mathcal{L}(\genAB{},\genBA{},\discA{},\discB{}) & = \mathcal{L}_{gan}(\genAB{}, \discB{}, \domA{}, \domB{}) 
    + \mathcal{L}_{gan}(\genBA{}, \discA{}, \domB{}, \domA{}) + \lambda \mathcal{L}_{cyc}(\genAB{}, \genBA{})
\end{split}
\end{align}
where $\genAB{}$ and $\genBA{}$ are trained to minimize $\mathcal{L}$, while $\discA{}$ and $\discB{}$ to maximize it.
Our work starts from a similar formulation and extends it to produce controllable transformations by means of parametrization-aware generators and discriminators.  

\subsection{Learning parametrizable transformations}
\label{ssec:architecture}

\begin{figure*}
    \centering
    \includegraphics[width=0.8\linewidth]{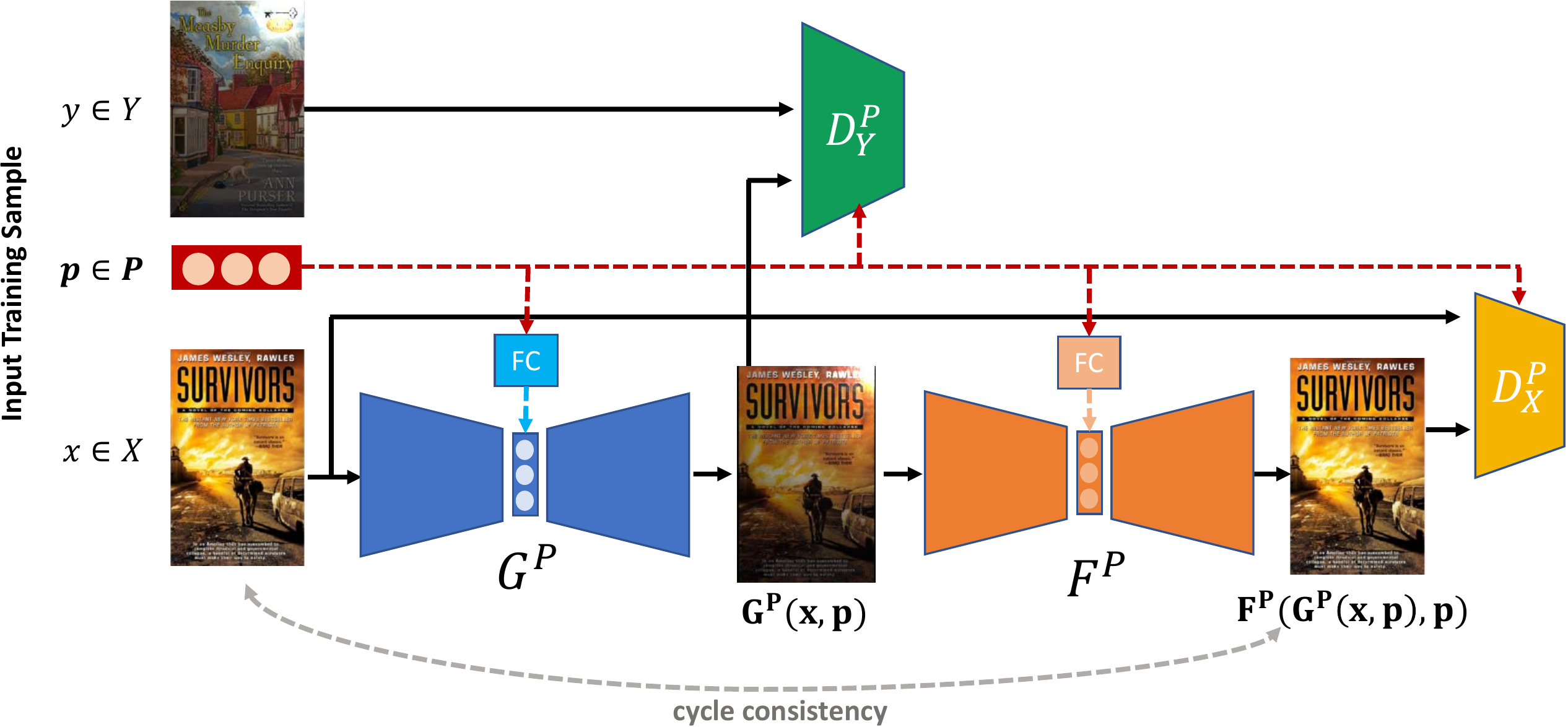}
    \caption{
    Training of the ParGAN architecture starting with a sample $\sampA{}\in\domA{}$. A complete train iteration includes a mirrored cycle starting with a sample $\sampB{} \in \domB{}$ being first fed to $\genParBA{}$, then to $\genParAB{}$.}
    \label{fig:architecture}
\end{figure*}

Since our objective is learning \emph{parametrizable} transformations, we extend the generators and discriminators of a cyclic architecture to have as additional input a conditioning vector denoted as $p$. 
We refer to those networks adding a P superscript: $\genParAB{}$, $\genParBA{}$, $\discParA{}$, $\discParB{}$. 
To maximize the generality of the proposed solution we encode the parameters of the transformation in a $n$-dimensional vector of floats: $\sampP{} \in \domPar{} \subset \mathbb{R}^n$.
In \autoref{fig:architecture} we report a schematic representation of half of a training iteration for the framework. 
A full train iteration will include a mirrored version of the architecture swapping $\sampA{}$ and $\sampB{}$ and inverting the order of the generators and discriminators.

\paragraph{Generators.} 
The generator architecture is based on a classic hourglass architecture, where a set of convolutional layers encode the image into a compressed representation that is fed to a series of residual blocks. 
In parallel, a stack of fully connected layers learn to transform the raw values of $\sampP$ to a higher dimensionality representation.
At the bottleneck layer of the generator the image representation and the transformed $\sampP$ are mixed by concatenation to every spatial location along the feature dimension. 
Then a series of convolutional layers decode the combined data into the final image. 
Instance normalization is used after each convolution with ReLU as activation function.

To compute $\mathcal{L}_{cyc}$, at training time, we provide both generators with the same conditioning vector $\sampP{}$.
Therefore, $\genParAB{}$ learns to apply the transformation parametrized by $\sampP{}$ while $\genParBA{}$ learns to undo it.

\paragraph{Discriminators.} 
Similarly to the generators, we extend the discriminator formulation of CycleGAN to take into account the transformation of the parameterization to learn. 
We denote these modified components as $\discParA{}$ and $\discParB{}$ respectively.
An initial convolutional layer operates solely on the image to extract a suitable deep representation. 
In parallel $\sampP{}$ is transformed through a series of fully connected layers in a similar fashion as in the generator. 
The learned representations of the image and of $\sampP{}$ are mixed by concatenation to every spatial location in the same way as the generators.
Afterwards, a series of convolutional layers operate on the mixed representation to produce a real/fake classification score for every image patch. 
The final prediction of the network is obtained with a mean average pooling over all patches. 
LeakyReLU is used as activation function, with instance normalization layers after every convolution.

\paragraph{Optimization objective.} The overall optimization goal of \pargan{} extends \autoref{eq:obj_cyclegan} to take into account the new parametric components and to rely on mean squared classification \cite{mao2017least}.
We denote with $(\sampA{},\sampB{},\sampP{}) \sim (\domA{},\domB{},\domPar{})$ the distribution of training examples.
Each training sample is composed of a random image from $\domA{}$ and an image from $\domB$ with its corresponding parametrization $p$.
We formalize our objective as: 
\vspace{-1mm}
\begin{equation}
\label{eq:loss_gan_p}
\begin{split}
    &\mathcal{L}_{gan}^\domPar{}(\genParAB{}, \discParB{}, \domA{}, \domB{}, \domPar{}) =  \mathbb{E}_{(\sampA{}, \sampB{}, \sampP{}) \sim (\domA{},\domB{},\domPar{})}
    \frac{1}{2} \left[ (1-\discParB{(}\sampB{}, \sampP{}))^2 + \discParB{}(\genAB{(}\sampA{}, \sampP{}), \sampP{})^2 \right]
\end{split}
\end{equation}
\vspace{-3mm}
\begin{equation}
\label{eq:loss_cycle_p}
\begin{split}
    \mathcal{L}_{cyc}^\domPar{}(\genParAB{}, \genParBA{}) & = \mathbb{E}_{(\sampA{}, \sampB{}, \sampP{}) \sim (\domA{}, \domB{}, \domPar{})}
    \frac{1}{2} \big[ ||\genBA{}^\domPar{}(\genAB{}^\domPar{}(\sampA{}, \sampP{}), \sampP{})-\sampA{}||_1 + ||\genAB{}^\domPar{}(\genBA{}^\domPar{}(\sampB{}, \sampP{}), \sampP{})-\sampB{}||_1 \big]
\end{split}
\end{equation}
\vspace{-2mm}
\begin{equation}
\label{eq:obj_ours}
\begin{split}
    \mathcal{L}(\genParAB{},\genParBA{},\discParA{},\discParB{}) & = \mathcal{L}_{gan}^\domPar{}(\genParAB{}, \discParB{}, \domA{}, \domB{}, \domPar{}) 
    + \mathcal{L}_{gan}^{\domPar{}}(\genParBA{}, \discParA{}, \domA{}, \domB{}, \domPar{})  + \lambda \mathcal{L}_{cyc}^\domPar{}(\genParAB{}, \genParBA{}) \\
\end{split}    
\end{equation}
\autoref{eq:obj_ours} defines the optimization objective composed of two GAN losses (defined in \autoref{eq:loss_gan_p}) and a cyclic consistency loss (defined \autoref{eq:loss_cycle_p}) weighted by a scalar $\lambda$.
The generators ($\genParAB{}$ and $\genParBA{}$) are optimized to minimize the objective, while the discriminators ($\discParA{}$ and $\discParB{}$) to maximize it.

\paragraph{Soft Parametrization:} 
As we will show in \autoref{ssec:soft} our formulation can be used even when the training set does not provide an explicit parametrization but only a source $(\domA{})$ and a target domain $(\domB)$.
In this setup, we can still use our framework to produce not only target images, but also to smoothly interpolate between the two domains.
First, we define a new target domain composed of $\domA{} \cup \domB{}$ and associate to each sample a one-dimensional parametrization vector $\sampP{}$.
We use as parametrization the value $0$ for samples from $\domA{}$ and $1$ for samples from $\domB{}$. 
Thus our target domain becomes $\{(x_1, 0)\dots(x_m, 0), (y_1, 1)\dots(y_n, 1)\}$.

Once we have created this softly parametrized target dataset, we train the GAN system according to \autoref{eq:obj_ours}.
Once $\genParAB{}$ is trained, we can use as parametrization any real value between $[0,1]$ to generate transformations that smoothly interpolate between the source and the target domains. 

When provided with many target domains $(\domB{_0},\dots, \domB{_n})$, we can apply the same technique using an $n$-dimensional vector for $\sampP{}$.
The parametrization corresponds to a vector of zeros for samples originated from $\domA{}$, while samples originated from the k\textsuperscript{th} target domain $\domB{_k}$ have the k\textsuperscript{th} entry set to one (one-hot encoding).
In \autoref{ssec:mixing} we show that the generator not only learns to smoothly interpolate between the source and any of the target domains independently, but also, to some extent, to mix the different transformations without having direct access to references for this kind of transformations during training.

\section{Experimental Results}
\label{sec:experimental}
\autoref{ssec:experimentalSetup} present our experimental setup, In \autoref{ssec:explicit} we show how we learn to replicate explicitly parametrized transformations.
\autoref{ssec:soft} shows the use of a soft parametrization approach
to learn a smooth interpolable transformation.
Finally \autoref{ssec:mixing} shows how to combine multiple transformations in an unsupervised fashion. 
Due to space constraint more experimental results and a comparison to other methods are reported in the supplementary.

\subsection{Experimental Setup}
\label{ssec:experimentalSetup}

We consider various datasets for experimental evaluation. Although some include paired samples (each image has a correspondence in the transformed domain), we always assume unpaired data.
All results reported in this paper have been generated using unseen images from the validation set.
\\
\textbf{Book covers.}
To experiment with the idea of learning complex data augmentations we have generated an ad-hoc dataset of book covers randomly illuminated by a single beam of light.
We start from clean and perfectly lit images from \cite{iwana2016judging}, which we  use as source domain ($\domA{}$).
Then we simulate the effect of a beam of light illuminating the cover with the Blender rendering engine\footnote{\href{https://www.blender.org/}{blender.org}} and use these images as target domain ($\domB{}$).
Each image is parametrized by a three dimensional float vector, with the first two dimensions encoding the $(x,y)$ coordinates of the center of the light in the image reference system, and the last the normalized intensity of the light.  
\\
\textbf{\synthia{}.} In order to experiment with a soft parameterization, we learn a transformation between different subsets of the \synthia{} \cite{7780721} dataset.
The dataset has 5 sequences, out of which we use \textit{01}, \textit{02}, \textit{04} and \textit{05} for training and \textit{06} for evaluation.
\\
\textbf{\init{}.} This dataset \cite{shen2019towards} consists of real-world dashcam images of driving scenes subdivided by weather conditions. 
We have used the \textit{sunny} and \textit{night} categories, with an 80:20 training/test split.

To measure quantitatively the impact of our transformation with respect to the source and target domains, we use the Fr\'echet Inception Distance (\FID{}) \cite{heusel2017gans}\footnote{We use the Inception v3 weights from \href{https://github.com/fchollet/deep-learning-models}{https://github.com/fchollet/deep-learning-models}.} and the Learned Perceptual Image Patch Similarity (\LPIPS{}) \cite{zhang2018unreasonable}\footnote{Implementation from \href{https://github.com/alexlee-gk/lpips-tensorflow}{github.com/alexlee-gk/lpips-tensorflow} using the AlexNet lin (v0.1) configuration.}.
\FID{} compares statistics of Inception-v3 activations to measure the distance between two image distributions, while \LPIPS{} explicitly compares pairs of images and evaluates the perceptual distance between image patches.

\subsection{Explicit Parametrization}
\label{ssec:explicit}

We first investigate if our model can learn an explicitly parametrized transformation when provided with a target dataset of annotated samples such as \bookCovers{}.

In \autoref{fig:explicit_parameterization} we show different images generated by \pargan{} starting from the same source image but with different parametrization vectors as input.
The generator correctly learns to generate realistic images augmented with a beam of light according to the requested parametrization.
For example in the last row the position of the light beam is fixed and only the beam intensity varies.
The picture also shows how the generator learns a \emph{smooth} and \emph{disentangled} transformation function even if trained with just sparse samples of the space of possible transformations parametrized by $\domPar{}$. 
Our formulation is the first to provide this level of \emph{explicit} and \emph{human interpretable} control over the transformation process.

In \autoref{tab:book_fid_lpips_y} we report the \FID{} and the \LPIPS{} metrics computed between different sets of real and generated samples parametrized by different $\sampP{}$.
We keep two axes of $p$ fixed and modify the remaining entry.

The tables show that when both distributions are parametrized by the same conditioning vector, the corresponding distances are minimal (bold value along the diagonal of the tables).
In case of a parameterization mismatch, instead, the distance between the two distributions tends to increase proportionally to the distance between the two parametrizations.
These results show that, for the \bookCovers{} dataset, our network can correctly learn the transformation parametrized by $\sampP{}$.

\newlength{\explicitParameterizationWidth}
\setlength{\explicitParameterizationWidth}{0.08\linewidth}
\begin{figure}
    \centering
    \setlength{\tabcolsep}{2pt}
    \begin{tabular}{c|ccccccc}
    {\scriptsize \textit{Input Image}}& & {\scriptsize \textit{Lowest}} & & & & & {\scriptsize \textit{Highest}}\\\midrule
    & \rotatebox{90}{\hspace{3mm} \scriptsize \textit{Change x}} &
    \includegraphics[width=\explicitParameterizationWidth]{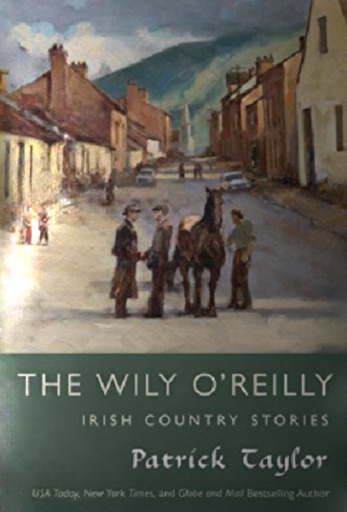} &
    \includegraphics[width=\explicitParameterizationWidth]{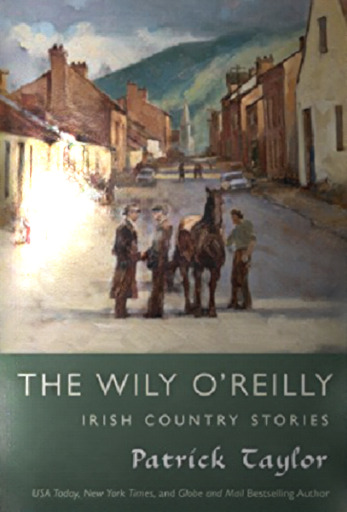} &
    \includegraphics[width=\explicitParameterizationWidth]{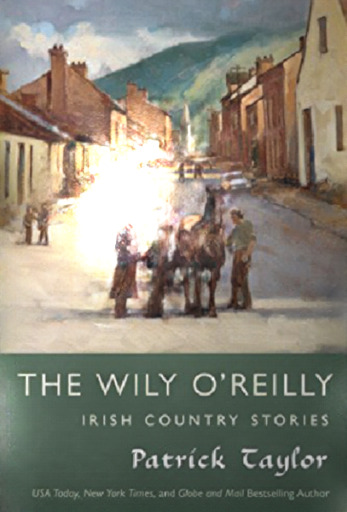} &
    \includegraphics[width=\explicitParameterizationWidth]{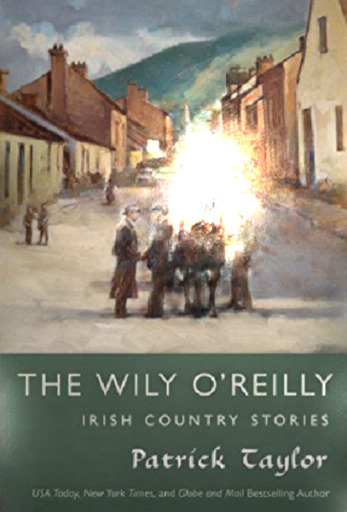} &
    \includegraphics[width=\explicitParameterizationWidth]{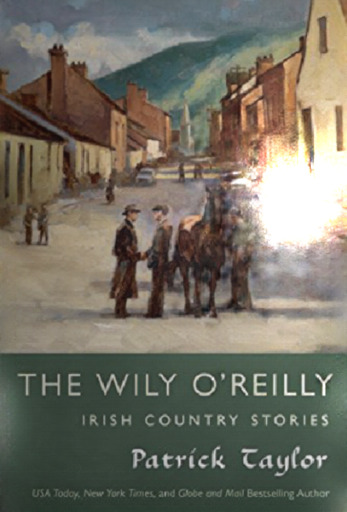} &
    \includegraphics[width=\explicitParameterizationWidth]{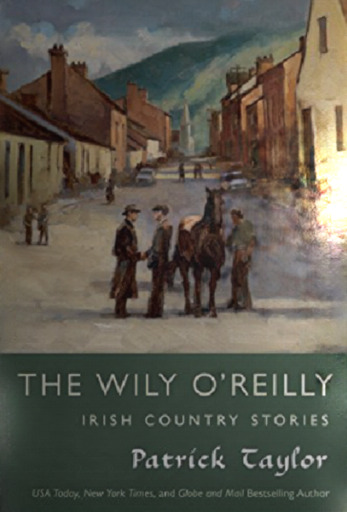}\\
    
    \includegraphics[width=\explicitParameterizationWidth]{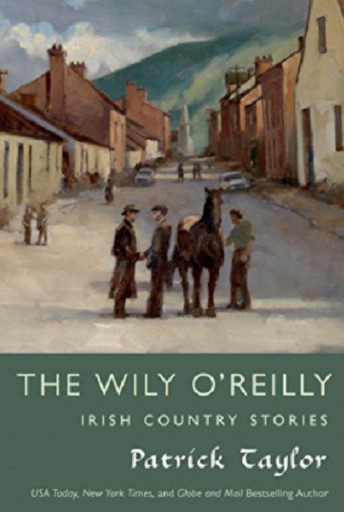} & \rotatebox{90}{\hspace{3mm} \scriptsize \textit{Change y}} &
    \includegraphics[width=\explicitParameterizationWidth]{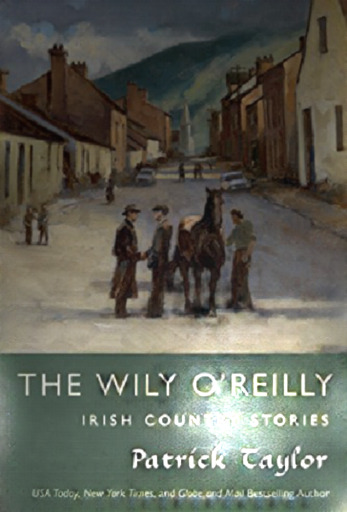} &
    \includegraphics[width=\explicitParameterizationWidth]{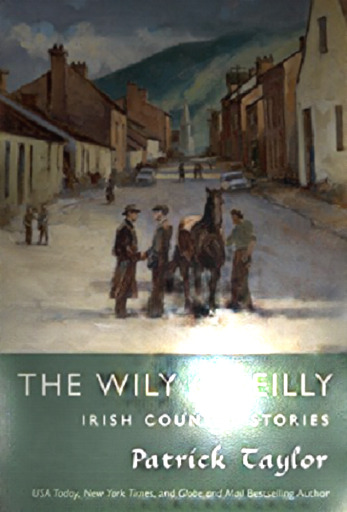} &
    \includegraphics[width=\explicitParameterizationWidth]{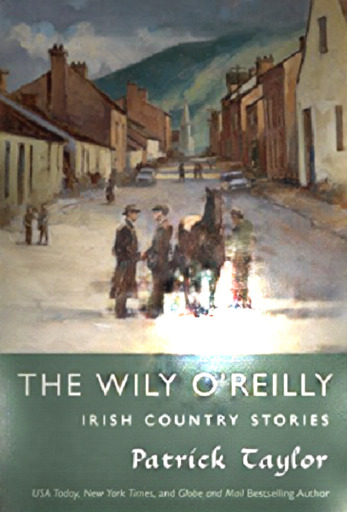} &
    \includegraphics[width=\explicitParameterizationWidth]{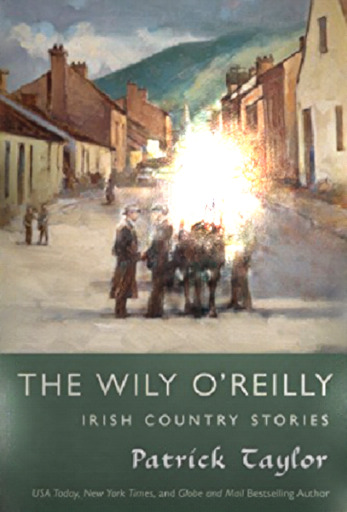} &
    \includegraphics[width=\explicitParameterizationWidth]{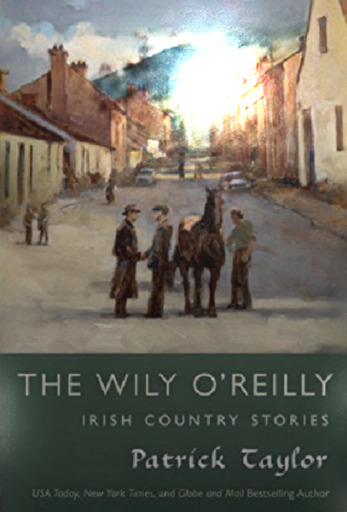} &
    \includegraphics[width=\explicitParameterizationWidth]{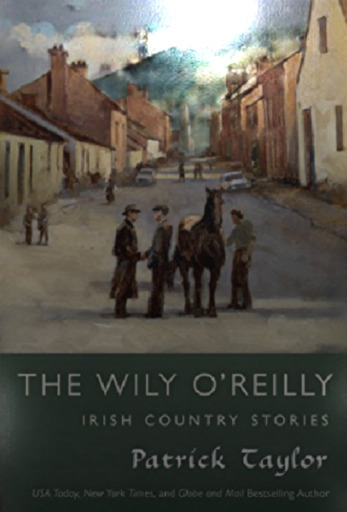}\\

    & \rotatebox{90}{\hspace{1mm} \scriptsize \textit{Illumination}} &
    \includegraphics[width=\explicitParameterizationWidth]{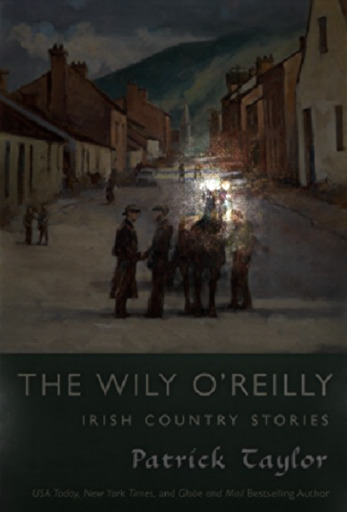} &
    \includegraphics[width=\explicitParameterizationWidth]{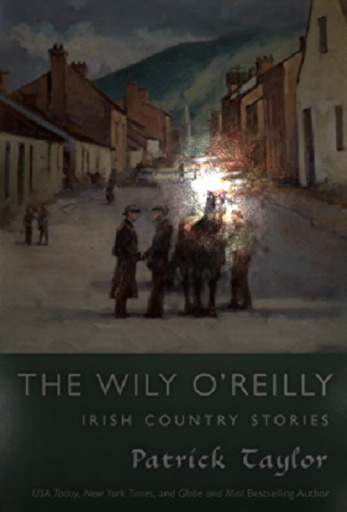} &
    \includegraphics[width=\explicitParameterizationWidth]{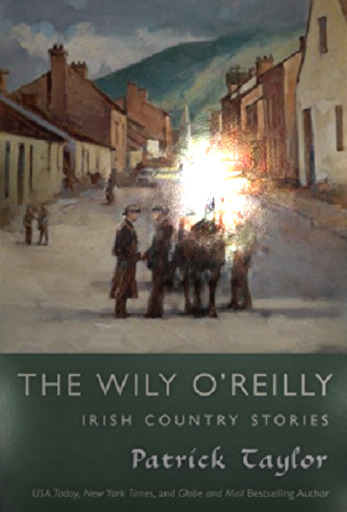} &
    \includegraphics[width=\explicitParameterizationWidth]{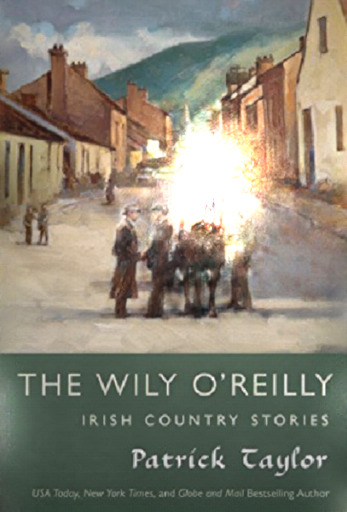} &
    \includegraphics[width=\explicitParameterizationWidth]{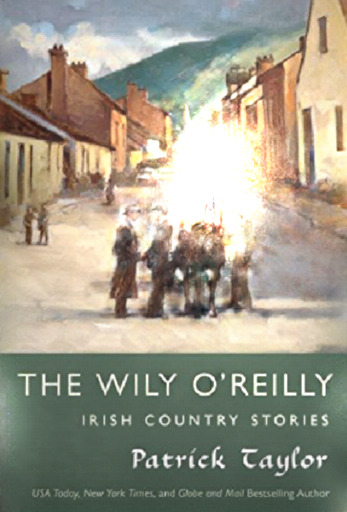} &
    \includegraphics[width=\explicitParameterizationWidth]{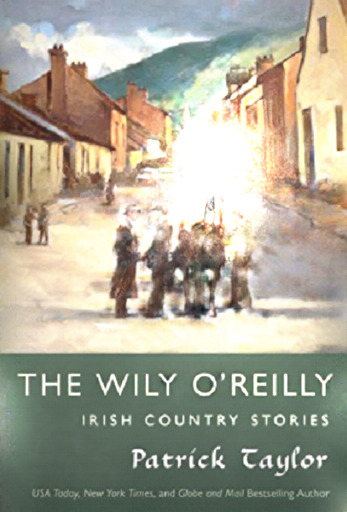}\\
    \end{tabular}
    \caption{Results on the \bookCovers{} dataset. Different components of the parameterization (x-axis, y-axis, illumination) are correctly applied independently by a single generator, as shown in each row.}
    \label{fig:explicit_parameterization}
\end{figure}

\newcommand{\bookFidLpipsScale}{0.8}
\begin{table}[h!]
\center
\scalebox{\bookFidLpipsScale}{
\begin{tabular}{ccc}
\begin{tabular}{l@{\hspace{.8\tabcolsep}}c|ccccc}
\toprule
& & \multicolumn{5}{c}{\textbf{Gen. images}}\\
 &\textbf{y}& 0.1 & 0.3 & 0.5 & 0.7 & 0.9 \\
 \hline
 \multirow{5}{*}{\rotatebox[]{90}{\textbf{Real images}}}
& 0.1 & \textbf{44.653} & 51.138 & 56.646 & 57.139 & 59.616 \\
& 0.3 & 45.999 & \textbf{39.51}  & 44.004 & 51.397 & 56.183 \\
& 0.5 & 58.377 & 49.87 & \textbf{36.199} & 43.358 & 51.292 \\
& 0.7 & 52.312 & 50.086 & 39.629 & 33.255 & 38.819 \\
& 0.9 & 59.292 & 56.471 & 46.54  & \textbf{32.878} & \textbf{27.206} \\
\bottomrule
\end{tabular} 
& &
\begin{tabular}{l@{\hspace{.8\tabcolsep}}c|ccccc}
\toprule
& & \multicolumn{5}{c}{\textbf{Gen. images}}\\
 &\textbf{y}& 0.1 & 0.3 & 0.5 & 0.7 & 0.9 \\
 \hline
 \multirow{5}{*}{\rotatebox[]{90}{\textbf{Real images}}}
& 0.1 & \textbf{0.064} & 0.084 & 0.115 & 0.134 & 0.147 \\
& 0.3 & 0.077 & \textbf{0.056} & 0.079 & 0.109 & 0.132 \\
& 0.5 & 0.109 & 0.082 & \textbf{0.055} & 0.083 & 0.108 \\
& 0.7 & 0.125 & 0.106 & 0.071 & \textbf{0.051} & 0.073 \\
& 0.9 & 0.134 & 0.120 & 0.095 & 0.059 & \textbf{0.041} \\
\bottomrule
\end{tabular} \\
(a) \FID{} &  & (b) \LPIPS{} \\
\end{tabular}
}
\caption{Distances between real and generated images for \bookCovers{}. We fix the x and illumination components of $p$ and sample different values for y. The generated images are closest (\textbf{bold}) to the real images with the same \textit{y} value and the distance increases with it. 
The scores average the values for 100 samples. For the results on the \textit{x} and illumination axes, we refer to the supplement.
}
\label{tab:book_fid_lpips_y}
\end{table}

\subsection{Soft Parametrization}
\label{ssec:soft}

We now test \pargan{} for the case when no explicit parametrization is provided at training time using the soft parametrization technique described in \autoref{ssec:architecture}. 
We train a mapping between two domains and evaluate the impact of the parametrization by sampling different values for $p$ in $[0,1]$.
In this scenario the input parametrization acts as a knob to control the intensity of the transformation being applied, with $0$ not modifying anything (\textit{source} $\rightarrow$ \textit{source} mapping) and $1$ applying the complete transformation (\textit{source} $\rightarrow$ \textit{target}).
Interestingly, the generator learns to smoothly interpolate between the two states even if during training no explicit supervision for this task was provided. In \autoref{fig:experimental_soft_parameterization} and \autoref{fig:teaser} we show images generated using this configuration.

\newlength{\softParameterizationLength}
\setlength{\softParameterizationLength}{0.13\linewidth}
\begin{figure}[h!]
\setlength{\tabcolsep}{1pt}
\centering
\begin{tabular}{c|ccccccc}
    \toprule
    {\footnotesize Source} & & {\footnotesize 0.0} & {\footnotesize 0.25} & {\footnotesize 0.5} & {\footnotesize 0.75} & {\footnotesize 1.0} \\ 
    \midrule
    \includegraphics[width=\softParameterizationLength]{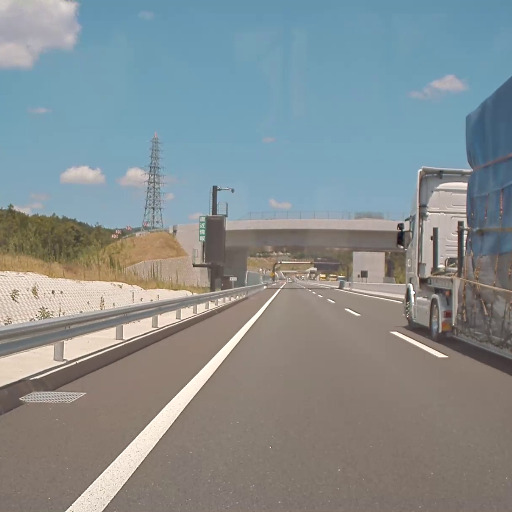}&
    \rotatebox{90}{\parbox{1.7cm}{\centering\scriptsize\init{}\\\daytonight{}}} & 
    \includegraphics[width=\softParameterizationLength]{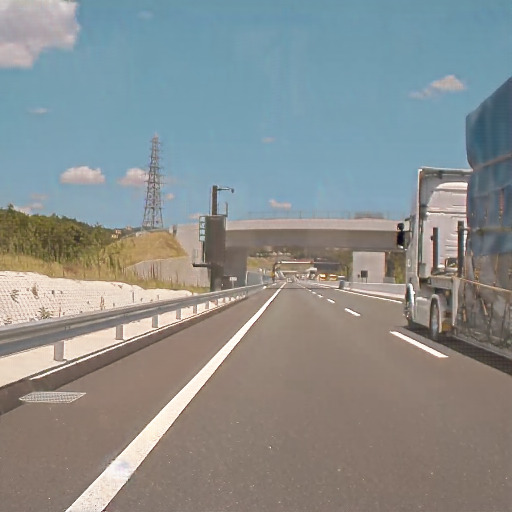}& \includegraphics[width=\softParameterizationLength]{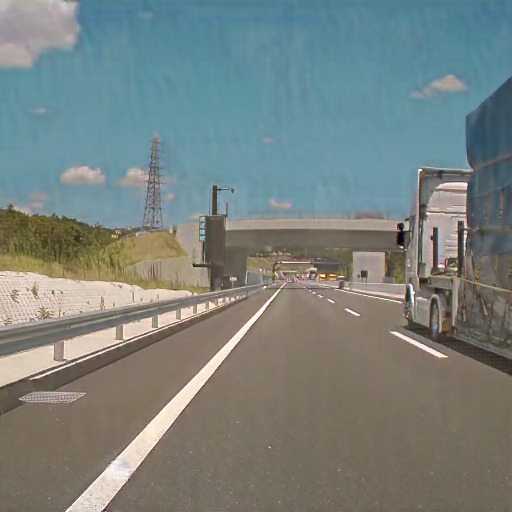}&
    \includegraphics[width=\softParameterizationLength]{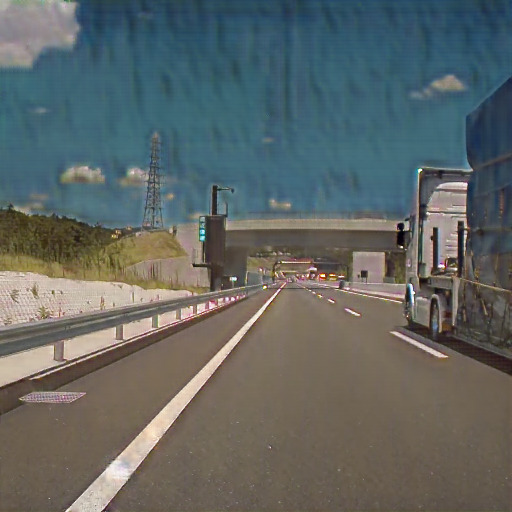}&
    \includegraphics[width=\softParameterizationLength]{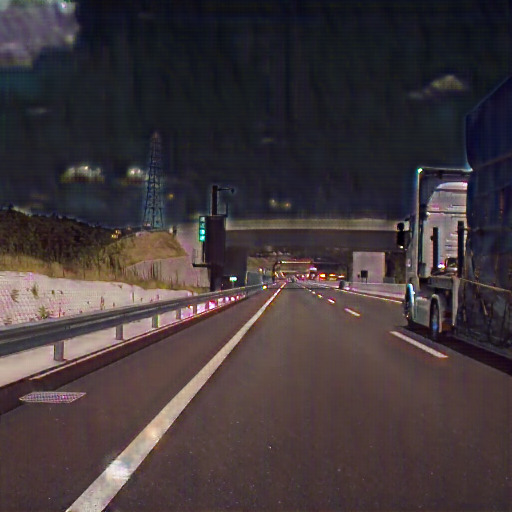}&
    \includegraphics[width=\softParameterizationLength]{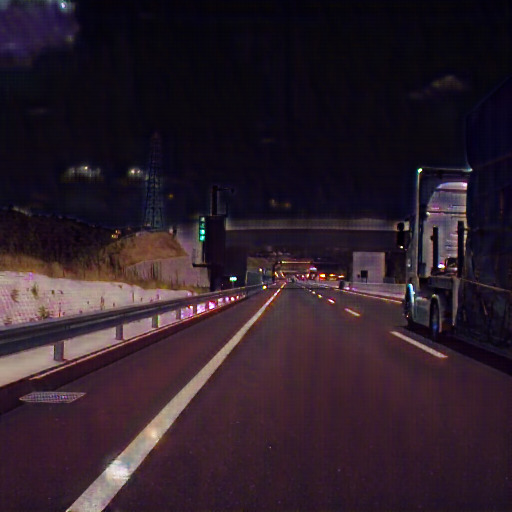}\\
    
\end{tabular}
\caption{Evaluation using a soft parametrization. We refer to the supplement for more samples.}
\label{fig:experimental_soft_parameterization}
\end{figure}

Ideally, the distance between the source and transformed images should increase with $p$, as the GAN progressively introduces more changes. 
Likewise, the distance to the target domain should decrease. 
We verify this experimentally and in \autoref{tab:fid_lpips_summertowinter_daytonight} we report the \FID{} and \LPIPS{} scores for the \synthia{} \summertowinter{} and \init{} \daytonight{} tasks comparing the result images with randomly picked sets of source and target samples. 

\begin{table}[h!]
\center
\scalebox{0.85}{
\begin{tabular}{c@{\hskip 5mm}c}
\begin{tabular}{c|cc|cc}
\toprule
&\multicolumn{2}{c}{\textbf{Vs. source images}} & \multicolumn{2}{c}{\textbf{Vs. target images}}\\
\hline
\textbf{p} & \FID{} & \LPIPS{} & \FID{} & \LPIPS{}\\
\hline
0    & \cellcolor{metric_step0}76.361 & \cellcolor{metric_step0}0.586 & \cellcolor{metric_step4}150.662 & \cellcolor{metric_step4}0.625 \\
0.25 & \cellcolor{metric_step1}79.533 & \cellcolor{metric_step1}0.591   & \cellcolor{metric_step3}147.997 & \cellcolor{metric_step3}0.616 \\
0.5  & \cellcolor{metric_step2}92.919  & \cellcolor{metric_step2}0.611  & \cellcolor{metric_step2}140.181 & \cellcolor{metric_step2}0.605  \\
0.75 & \cellcolor{metric_step4}101.385 & \cellcolor{metric_step4}0.618  & \cellcolor{metric_step1}137.599 & \cellcolor{metric_step1}0.601  \\
1    & \cellcolor{metric_step3}100.634 & \cellcolor{metric_step3}0.613  & \cellcolor{metric_step0}132.208 & \cellcolor{metric_step0}0.590  \\
\bottomrule
\end{tabular} &
\begin{tabular}{c|cc|cc}
\toprule
&\multicolumn{2}{c}{\textbf{Vs. source images}} & \multicolumn{2}{c}{\textbf{Vs. target images}}\\
\hline
\textbf{p} & \FID{} & \LPIPS{} & \FID{} & \LPIPS{}\\
\hline
0. & \cellcolor{metric_step0} 95.625 & \cellcolor{metric_step0} 0.650 & \cellcolor{metric_step4}188.851 & \cellcolor{metric_step4} 0.734\\
0.25 & \cellcolor{metric_step1} 96.735 & \cellcolor{metric_step1} 0.654 & \cellcolor{metric_step3}183.410 & \cellcolor{metric_step3} 0.731\\
0.5 & \cellcolor{metric_step2} 101.213 & \cellcolor{metric_step2} 0.659 & \cellcolor{metric_step2}177.468 & \cellcolor{metric_step2} 0.721\\
0.75 & \cellcolor{metric_step3} 108.562 & \cellcolor{metric_step3} 0.668 & \cellcolor{metric_step1}166.436 & \cellcolor{metric_step1} 0.700\\
1 & \cellcolor{metric_step4} 115.793 & \cellcolor{metric_step4} 0.683 & \cellcolor{metric_step0}149.868 & \cellcolor{metric_step0} 0.679\\
\bottomrule
\end{tabular}\\
a) \synthia{} \summertowinter{} & b) \init{} \daytonight{}\\
\end{tabular}
}
\caption{Average \FID{} and \LPIPS{} scores. Dark color in the denotes lower values (more similar images). The distance w.r.t. the source domain increases with $p$ and decreases for the target.}
\label{tab:fid_lpips_summertowinter_daytonight}
\end{table}

The distance to the target images is often much larger than for the source domain, since in one case we are applying a complex image transformation and in the other a simple identity mapping. 

\subsection{Simultaneous learning of transformations}

We test the behavior of our model when using a single source domain and multiple targets. Following the strategy outlined in \autoref{ssec:architecture} we use a soft-parameterization associating to each target domain a one-hot encoded categorical vector $p$. 

Thus, for a two-target transformation, the source domain has label $p=[0, 0]$, and the two target domains have $p=[1,0]$ and $p=[0,1]$ respectively.

We train in the same fashion as the previous experiments for the \summertowinternight{} task on the \synthia{} dataset and report the result in \autoref{fig:experimental_synthia_summer2winternight}.
Each transformation is learned independently and does not conflict with the others (the generator learns a \textit{disentangled} mapping). 
We refer to the supplementary material for a qualitative and quantitative comparison between the transformations learned jointly and independently as well as for experimental results.

\newlength{\synthiaSummerWinterNightLen}
\setlength{\synthiaSummerWinterNightLen}{0.13\linewidth}
\begin{figure}[t!]
    \centering
    \setlength{\tabcolsep}{2pt}
    \begin{tabular}{c|cccc}
    
    \multirow{2}{*}[6mm]{\begin{overpic}[width=\synthiaSummerWinterNightLen]{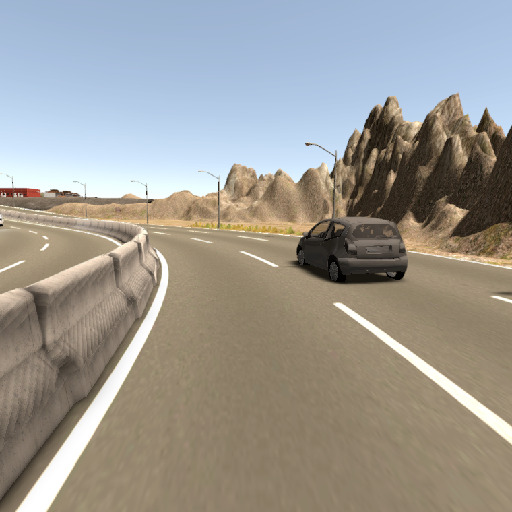}
        \put (17,6) {\parbox{15mm}{\begin{flushright}\textbf{\textcolor{white}{{\scriptsize $\text{Source}$}}}\end{flushright}}}
    \end{overpic}} &
    
    \begin{overpic}[width=\synthiaSummerWinterNightLen]{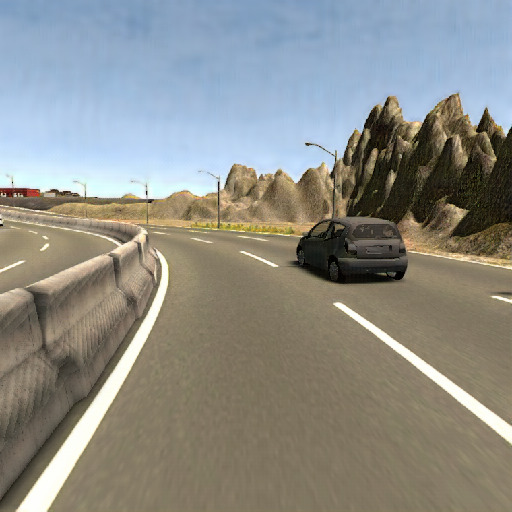}
        \put (17,6) {\parbox{15mm}{\begin{flushright}\textbf{\textcolor{white}{{\scriptsize $p=[0, 0.25]$}}}\end{flushright}}}
    \end{overpic} &
    \begin{overpic}[width=\synthiaSummerWinterNightLen]{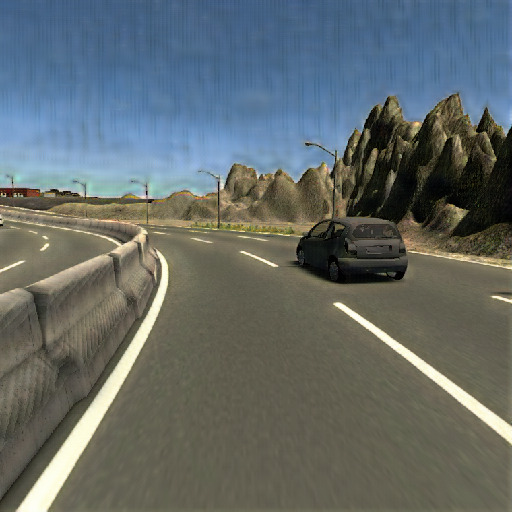}
        \put (17,6) {\parbox{15mm}{\begin{flushright}\textbf{\textcolor{white}{{\scriptsize $p=[0, 0.50]$}}}\end{flushright}}}
    \end{overpic} &
    \begin{overpic}[width=\synthiaSummerWinterNightLen]{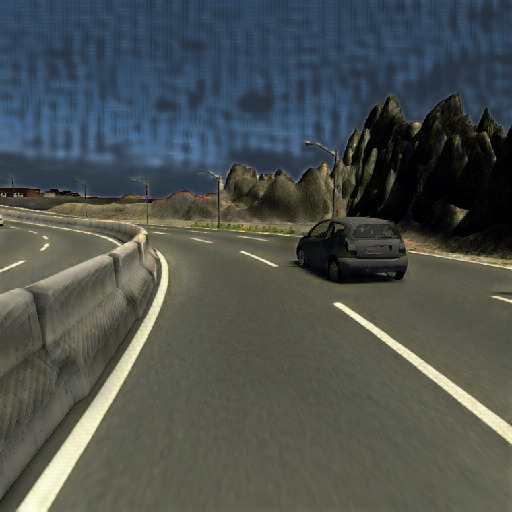}
        \put (17,6) {\parbox{15mm}{\begin{flushright}\textbf{\textcolor{white}{{\scriptsize $p=[0, 0.75]$}}}\end{flushright}}}
    \end{overpic} &
    \begin{overpic}[width=\synthiaSummerWinterNightLen]{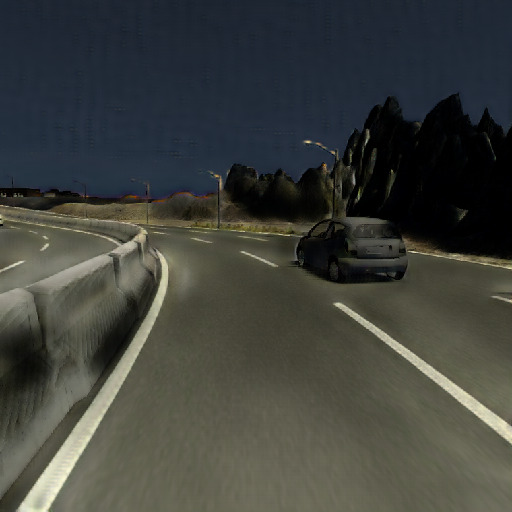}
        \put (17,6) {\parbox{15mm}{\begin{flushright}\textbf{\textcolor{white}{{\scriptsize $p=[0, 1]$}}}\end{flushright}}}
    \end{overpic} \\
    &
    \begin{overpic}[width=\synthiaSummerWinterNightLen]{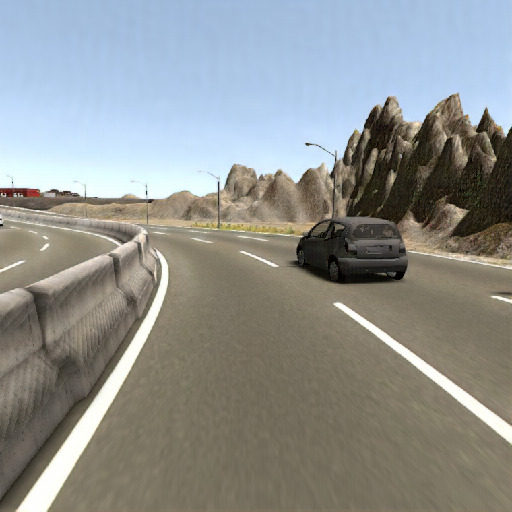}
        \put (17,6) {\parbox{15mm}{\begin{flushright}\textbf{\textcolor{white}{{\scriptsize $p=[0.25, 0]$}}}\end{flushright}}}
    \end{overpic} &
    \begin{overpic}[width=\synthiaSummerWinterNightLen]{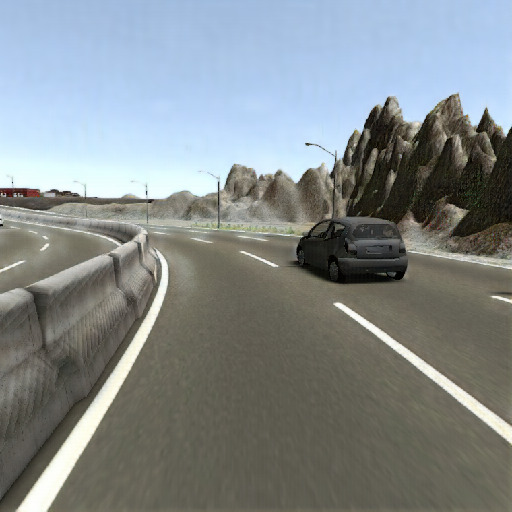}
        \put (17,6) {\parbox{15mm}{\begin{flushright}\textbf{\textcolor{white}{{\scriptsize $p=[0.50, 0]$}}}\end{flushright}}}
    \end{overpic} &
    \begin{overpic}[width=\synthiaSummerWinterNightLen]{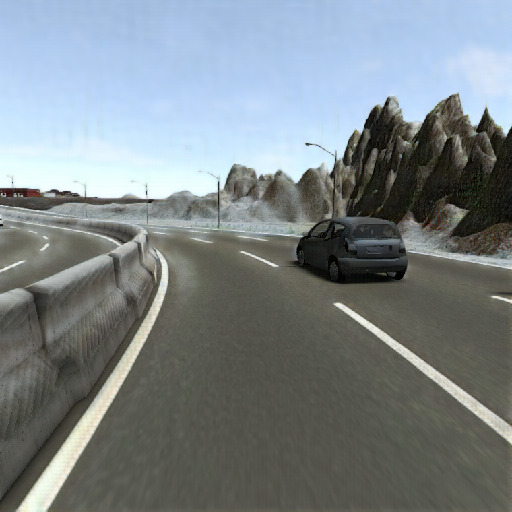}
        \put (17,6) {\parbox{15mm}{\begin{flushright}\textbf{\textcolor{white}{{\scriptsize $p=[0.75, 0]$}}}\end{flushright}}}
    \end{overpic} &
    \begin{overpic}[width=\synthiaSummerWinterNightLen]{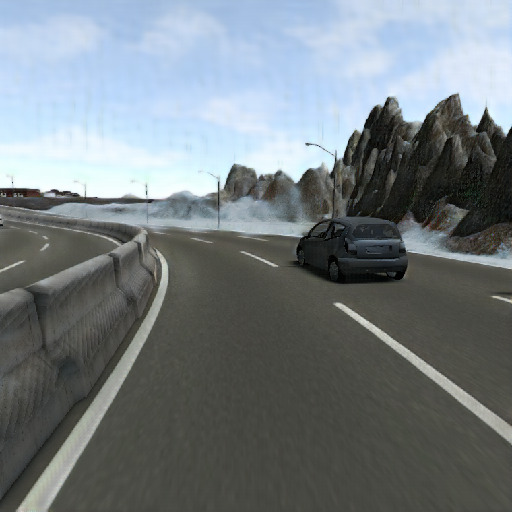}
        \put (17,6) {\parbox{15mm}{\begin{flushright}\textbf{\textcolor{white}{{\scriptsize $p=[1, 0]$}}}\end{flushright}}}
    \end{overpic} \\
    \end{tabular}
    \caption{\summertowinternight{} transformations (\textbf{Top}: Night, \textbf{Bottom}: Winter). The network learns disentangled transformations that can be applied without mixing.}
    \label{fig:experimental_synthia_summer2winternight}
\end{figure}

\subsubsection*{Mixing Transformations}
\label{ssec:mixing}

\pargan{} can learn to \emph{mix} the two transformations to some extent, producing images that display features from different domains (e.g. dark sky and snow). We show some results in \autoref{fig:mixings_winter_night_softrain}.

\newlength{\arbitraryMixingsLen}
\setlength{\arbitraryMixingsLen}{0.17\linewidth}
\begin{figure}[h!]
    \centering
    \setlength{\tabcolsep}{1pt}
    \begin{tabular}{ccccc}
    \begin{overpic}[width=\arbitraryMixingsLen]{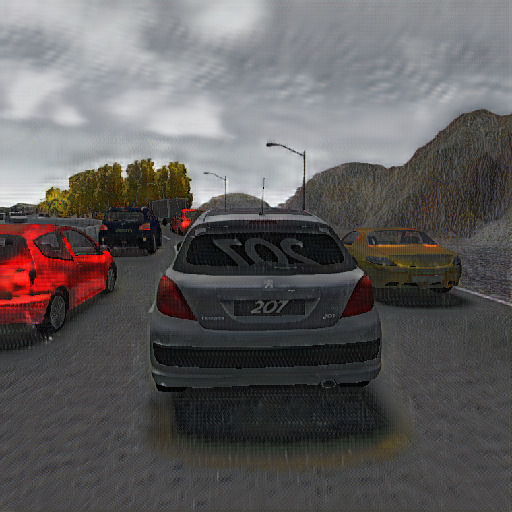}
        \put (34,20) {\parbox{15mm}{\begin{flushright}\textbf{\textcolor{white}{{\scriptsize 0\% \textit{winter}} {\scriptsize 25\% \textit{night}} {\scriptsize 75\% \textit{rain}}}}\end{flushright}}}
    \end{overpic} &
    \begin{overpic}[width=\arbitraryMixingsLen]{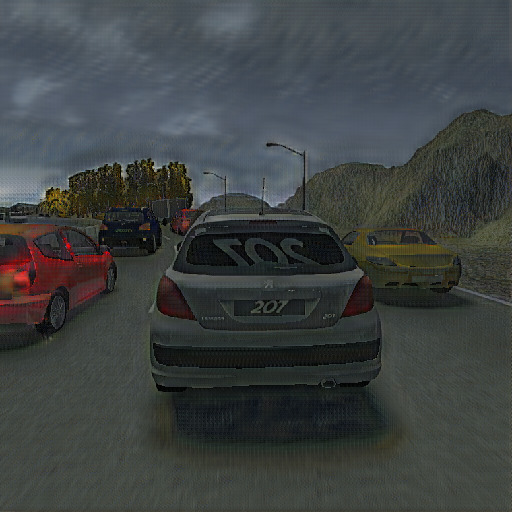}
        \put (34,20) {\parbox{15mm}{\begin{flushright}\textbf{\textcolor{white}{{\scriptsize 0\% \textit{winter}} {\scriptsize 100\% \textit{night}} {\scriptsize 100\% \textit{rain}}}}\end{flushright}}}
    \end{overpic} &
    \begin{overpic}[width=\arbitraryMixingsLen]{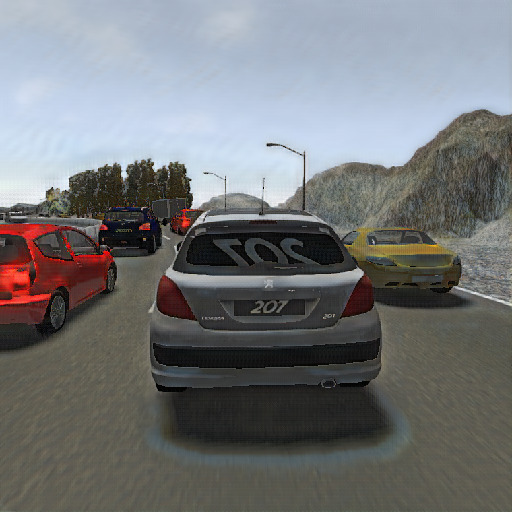}
        \put (34,20) {\parbox{15mm}{\begin{flushright}\textbf{\textcolor{white}{{\scriptsize 25\% \textit{winter}} {\scriptsize 0\% \textit{night}} {\scriptsize 50\% \textit{rain}}}}\end{flushright}}}
    \end{overpic} &
    
    \begin{overpic}[width=\arbitraryMixingsLen]{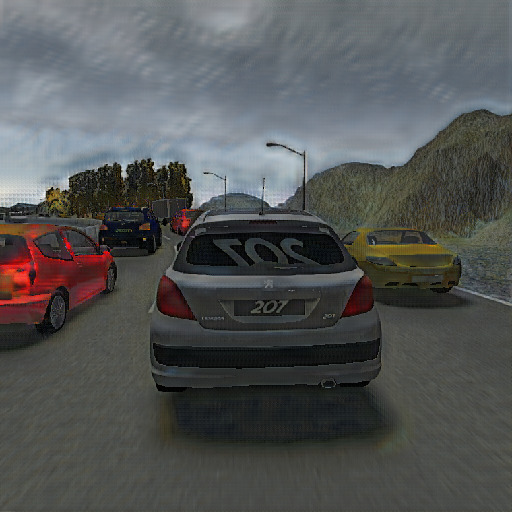}
        \put (34,20) {\parbox{15mm}{\begin{flushright}\textbf{\textcolor{white}{{\scriptsize 50\% \textit{winter}} {\scriptsize 75\% \textit{night}} {\scriptsize 50\% \textit{rain}}}}\end{flushright}}}
    \end{overpic} &
    
    \begin{overpic}[width=\arbitraryMixingsLen]{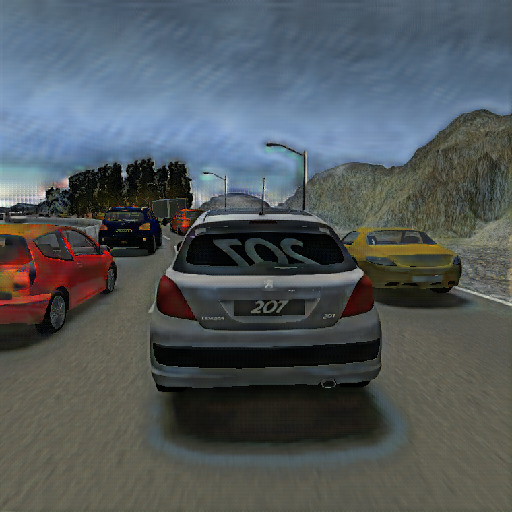}
        \put (34,20) {\parbox{15mm}{\begin{flushright}\textbf{\textcolor{white}{{\scriptsize 50\% \textit{winter}} {\scriptsize 75\% \textit{night}} {\scriptsize 0\% \textit{rain}}}}\end{flushright}}}
    \end{overpic} \\
    
    \end{tabular}
    \caption{Arbitrary mixings on the \summertowinternightsoftrain{} task.}
    \label{fig:mixings_winter_night_softrain}
\end{figure}
\section{Conclusions}
\label{sec:conclusions}

We have proposed \pargan{}, a new model for \imtoim{} translation using intuitive parameterizations to control the outcome.
Our formulation is able to learn not only how to replicate complex parametric transformations, but also how to smoothly interpolate between disjoint domains using only a soft categorical parameterization. 
Additionally, our method does not rely on prior knowledge or ad-hoc architectures. While those approaches would have likely lead to faster convergence and higher reconstruction quality, it would constrain the versatility of the method to a small range of tasks, requiring fundamental changes in order to learn different transformations. 
Since human interpretability is at the core of our proposal, our network always takes a parameterization of the transformation encoded in real-world magnitudes. 
Finally, we have shown some promising results on learning how to mix different transformations in a completely unsupervised fashion.

In the future, we wish to test our method on more parametric transformations to further verify the generality of the proposed solutions.
Moreover, our method still generates some artifacts while trying to mix different transformations. 
We believe that this issue might be mitigated by explicitly adding an adversarial component in the loss function to target these use cases.
We will explore this direction in future works.
\clearpage
\section*{Broader Impact}

Despite their recent development, learning-based approaches for image transformation have already found applications in various fields, such as entertainment or content generation. We believe that tasks such as object detection or semantic segmentation can benefit from our method by increasing the amount of available training data. Specially, our method can be of particular interest to reduce bias for unbalanced categories, such as day and night in the context of autonomous driving. This can help to increase accuracy on underrepresented modalities. Parametrizable methods like ours can also determine the kind of data to generate more precisely based on particular requirements. Examples are constraining the location of an augmentation to a particular area, or limiting the percentage of a global transformation to apply (such as 50\% snow). In addition to data augmentation techniques, the network can also be deployed as part of image editing tools to allow end users to edit images directly.

While the quality of GAN-generated imagery keeps improving, some generated images may still present some artifacts. This is not only aesthetically unpleasant but it can also impact the success of downstream tasks inducing unwanted biases into the generated datasets. A recent work \cite{DBLP:journals/corr/abs-1901-03892}  also points out that images generated by cyclic architectures might contain invisible artifacts not present in the original image distribution, again a source of bias in the data set. The main way to fight these unwanted artifacts is to increase the quality of the generated images (\ie, better architectures and better adversarial loss functions) as well as to add explicit regularization to limit this type of misbehavior.  

\par\vfill\par

\clearpage

\bibliographystyle{splncs}
\bibliography{egbib}

\begin{thebibliography}{10}

\bibitem{CycleGAN2017}
Zhu, J.Y., Park, T., Isola, P., Efros, A.A.:
\newblock Unpaired image-to-image translation using cycle-consistent
  adversarial networks.
\newblock In: Computer Vision (ICCV), 2017 IEEE International Conference on.
  (2017)

\bibitem{pix2pix2016}
Isola, P., Zhu, J.Y., Zhou, T., Efros, A.A.:
\newblock Image-to-image translation with conditional adversarial networks.
\newblock arxiv (2016)

\bibitem{hoffman2014continuous}
Hoffman, J., Darrell, T., Saenko, K.:
\newblock Continuous manifold based adaptation for evolving visual domains.
\newblock In: Proceedings of the IEEE Conference on Computer Vision and Pattern
  Recognition. (2014)  867--874

\bibitem{zhu2017toward}
Zhu, J.Y., Zhang, R., Pathak, D., Darrell, T., Efros, A.A., Wang, O.,
  Shechtman, E.:
\newblock Toward multimodal image-to-image translation.
\newblock In: Advances in Neural Information Processing Systems. (2017)
  465--476

\bibitem{huang2018multimodal}
Huang, X., Liu, M.Y., Belongie, S., Kautz, J.:
\newblock Multimodal unsupervised image-to-image translation.
\newblock In: Proceedings of the European Conference on Computer Vision (ECCV).
  (2018)  172--189

\bibitem{DBLP:journals/corr/abs-1808-00948}
Lee, H., Tseng, H., Huang, J., Singh, M.K., Yang, M.:
\newblock Diverse image-to-image translation via disentangled representations.
\newblock CoRR \textbf{abs/1808.00948} (2018)

\bibitem{wang2018pix2pixHD}
Wang, T.C., Liu, M.Y., Zhu, J.Y., Tao, A., Kautz, J., Catanzaro, B.:
\newblock High-resolution image synthesis and semantic manipulation with
  conditional gans.
\newblock In: Proceedings of the IEEE Conference on Computer Vision and Pattern
  Recognition. (2018)

\bibitem{stylegan}
Karras, T., Laine, S., Aila, T.:
\newblock A style-based generator architecture for generative adversarial
  networks.
\newblock CoRR \textbf{abs/1812.04948} (2018)

\bibitem{DBLP:journals/corr/ChenDHSSA16}
Chen, X., Duan, Y., Houthooft, R., Schulman, J., Sutskever, I., Abbeel, P.:
\newblock Infogan: Interpretable representation learning by information
  maximizing generative adversarial nets.
\newblock CoRR \textbf{abs/1606.03657} (2016)

\bibitem{DBLP:conf/nips/ZhuZPDEWS17}
Zhu, J., Zhang, R., Pathak, D., Darrell, T., Efros, A.A., Wang, O., Shechtman,
  E.:
\newblock Toward multimodal image-to-image translation.
\newblock In Guyon, I., von Luxburg, U., Bengio, S., Wallach, H.M., Fergus, R.,
  Vishwanathan, S.V.N., Garnett, R., eds.: Advances in Neural Information
  Processing Systems 30: Annual Conference on Neural Information Processing
  Systems 2017, 4-9 December 2017, Long Beach, CA, {USA}. (2017)  465--476

\bibitem{huang2018munit}
Huang, X., Liu, M.Y., Belongie, S., Kautz, J.:
\newblock Multimodal unsupervised image-to-image translation.
\newblock In: ECCV. (2018)

\bibitem{nguyenphuoc2019hologan}
Nguyen-Phuoc, T., Li, C., Theis, L., Richardt, C., Yang, Y.L.:
\newblock Hologan: Unsupervised learning of 3d representations from natural
  images (2019)

\bibitem{shi2019facetoparameter}
Shi, T., Yuan, Y., Fan, C., Zou, Z., Shi, Z., Liu, Y.:
\newblock Face-to-parameter translation for game character auto-creation (2019)

\bibitem{NIPS2014_5423}
Goodfellow, I., Pouget-Abadie, J., Mirza, M., Xu, B., Warde-Farley, D., Ozair,
  S., Courville, A., Bengio, Y.:
\newblock Generative adversarial nets.
\newblock In Ghahramani, Z., Welling, M., Cortes, C., Lawrence, N.D.,
  Weinberger, K.Q., eds.: Advances in Neural Information Processing Systems 27.
\newblock Curran Associates, Inc. (2014)  2672--2680

\bibitem{DBLP:journals/corr/MirzaO14}
Mirza, M., Osindero, S.:
\newblock Conditional generative adversarial nets.
\newblock CoRR \textbf{abs/1411.1784} (2014)

\bibitem{kim2017learning}
Kim, T., Cha, M., Kim, H., Lee, J.K., Kim, J.:
\newblock Learning to discover cross-domain relations with generative
  adversarial networks.
\newblock In: Proceedings of the 34th International Conference on Machine
  Learning-Volume 70, JMLR. org (2017)  1857--1865

\bibitem{DBLP:journals/corr/abs-1711-03213}
Hoffman, J., Tzeng, E., Park, T., Zhu, J., Isola, P., Saenko, K., Efros, A.A.,
  Darrell, T.:
\newblock Cycada: Cycle-consistent adversarial domain adaptation.
\newblock CoRR \textbf{abs/1711.03213} (2017)

\bibitem{choi2018stargan}
Choi, Y., Choi, M., Kim, M., Ha, J.W., Kim, S., Choo, J.:
\newblock Stargan: Unified generative adversarial networks for multi-domain
  image-to-image translation.
\newblock In: Proceedings of the IEEE Conference on Computer Vision and Pattern
  Recognition. (2018)

\bibitem{DBLP:journals/corr/abs-1903-07291}
Park, T., Liu, M., Wang, T., Zhu, J.:
\newblock Semantic image synthesis with spatially-adaptive normalization.
\newblock CoRR \textbf{abs/1903.07291} (2019)

\bibitem{DBLP:journals/corr/AntipovBD17}
Antipov, G., Baccouche, M., Dugelay, J.:
\newblock Face aging with conditional generative adversarial networks.
\newblock CoRR \textbf{abs/1702.01983} (2017)

\bibitem{DBLP:conf/cvpr/WangTLG18}
Wang, Z., Tang, X., Luo, W., Gao, S.:
\newblock Face aging with identity-preserved conditional generative adversarial
  networks.
\newblock In: 2018 {IEEE} Conference on Computer Vision and Pattern
  Recognition, {CVPR} 2018, Salt Lake City, UT, USA, June 18-22, 2018, {IEEE}
  Computer Society (2018)  7939--7947

\bibitem{DBLP:journals/corr/LuTT17}
Lu, Y., Tai, Y., Tang, C.:
\newblock Conditional cyclegan for attribute guided face image generation.
\newblock CoRR \textbf{abs/1705.09966} (2017)

\bibitem{Yu2019a}
Yu, X., Porikli, F., Fernando, B., Hartley, R.:
\newblock Hallucinating unaligned face images by multiscale transformative
  discriminative networks.
\newblock International Journal of Computer Vision (2019)

\bibitem{DBLP:journals/corr/abs-1907-10786}
Shen, Y., Gu, J., Tang, X., Zhou, B.:
\newblock Interpreting the latent space of gans for semantic face editing.
\newblock CoRR \textbf{abs/1907.10786} (2019)

\bibitem{DBLP:journals/corr/abs-1907-07171}
Jahanian, A., Chai, L., Isola, P.:
\newblock On the "steerability" of generative adversarial networks.
\newblock CoRR \textbf{abs/1907.07171} (2019)

\bibitem{Anokhin_2020_CVPR}
Anokhin, I., Solovev, P., Korzhenkov, D., Kharlamov, A., Khakhulin, T.,
  Silvestrov, A., Nikolenko, S., Lempitsky, V., Sterkin, G.:
\newblock High-resolution daytime translation without domain labels.
\newblock In: The IEEE Conference on Computer Vision and Pattern Recognition
  (CVPR). (June 2020)

\bibitem{mao2017least}
Mao, X., Li, Q., Xie, H., Lau, R.Y., Wang, Z., Paul~Smolley, S.:
\newblock Least squares generative adversarial networks.
\newblock In: Proceedings of the IEEE International Conference on Computer
  Vision. (2017)  2794--2802

\bibitem{iwana2016judging}
Iwana, B.K., Raza~Rizvi, S.T., Ahmed, S., Dengel, A., Uchida, S.:
\newblock Judging a book by its cover.
\newblock arXiv preprint arXiv:1610.09204 (2016)

\bibitem{7780721}
{Ros}, G., {Sellart}, L., {Materzynska}, J., {Vazquez}, D., {Lopez}, A.M.:
\newblock The synthia dataset: A large collection of synthetic images for
  semantic segmentation of urban scenes.
\newblock In: 2016 IEEE Conference on Computer Vision and Pattern Recognition
  (CVPR). (June 2016)  3234--3243

\bibitem{shen2019towards}
Shen, Z., Huang, M., Shi, J., Xue, X., Huang, T.:
\newblock Towards instance-level image-to-image translation.
\newblock In: Proceedings of IEEE Conference on Computer Vision and Pattern
  Recognition. (2019)

\bibitem{heusel2017gans}
Heusel, M., Ramsauer, H., Unterthiner, T., Nessler, B., Hochreiter, S.:
\newblock Gans trained by a two time-scale update rule converge to a local nash
  equilibrium.
\newblock In: Advances in Neural Information Processing Systems. (2017)
  6626--6637

\bibitem{zhang2018unreasonable}
Zhang, R., Isola, P., Efros, A.A., Shechtman, E., Wang, O.:
\newblock The unreasonable effectiveness of deep features as a perceptual
  metric.
\newblock In: Proceedings of the IEEE Conference on Computer Vision and Pattern
  Recognition. (2018)  586--595

\bibitem{DBLP:journals/corr/abs-1901-03892}
Zhang, K.A., Cuesta{-}Infante, A., Xu, L., Veeramachaneni, K.:
\newblock Steganogan: High capacity image steganography with gans.
\newblock CoRR \textbf{abs/1901.03892} (2019)

\bibitem{tensorflow2015-whitepaper}
Abadi, M., Agarwal, A., Barham, P., Brevdo, E., Chen, Z., Citro, C., Corrado,
  G.S., Davis, A., Dean, J., Devin, M., Ghemawat, S., Goodfellow, I., Harp, A.,
  Irving, G., Isard, M., Jia, Y., Jozefowicz, R., Kaiser, L., Kudlur, M.,
  Levenberg, J., Man\'{e}, D., Monga, R., Moore, S., Murray, D., Olah, C.,
  Schuster, M., Shlens, J., Steiner, B., Sutskever, I., Talwar, K., Tucker, P.,
  Vanhoucke, V., Vasudevan, V., Vi\'{e}gas, F., Vinyals, O., Warden, P.,
  Wattenberg, M., Wicke, M., Yu, Y., Zheng, X.:
\newblock {TensorFlow}: Large-scale machine learning on heterogeneous systems
  (2015) Software available from tensorflow.org.

\bibitem{DBLP:journals/corr/KingmaB14}
Kingma, D.P., Ba, J.:
\newblock Adam: {A} method for stochastic optimization.
\newblock In Bengio, Y., LeCun, Y., eds.: 3rd International Conference on
  Learning Representations, {ICLR} 2015, San Diego, CA, USA, May 7-9, 2015,
  Conference Track Proceedings. (2015)

\end{thebibliography}

\clearpage
\setcounter{page}{1}
\setcounter{figure}{0}
\setcounter{table}{0}
\setcounter{section}{1}
\clearpage
\section*{Supplementary material for \@title}
\label{sec:supplement}

In the main body of this paper we have proposed our method for controllable \imtoim{} translation and showed its capabilities on different datasets. 
In this supplement, we show additional experiments and qualitative results in order to further analyze the properties of our method and how it compares to existing approaches.

In \autoref{supp:datasets} we describe the datasets used to evaluate our method. Sections \ref{supp:results_explicit},\ref{supp:disentanglement} and \ref{supp:results_soft} report additional quantitative and qualitative results. 
In \autoref{supp:transformation_quality} we compare our model to a standard CycleGAN \cite{CycleGAN2017} to show that the addition of the parametrization does not affect the generated image quality. 
In \autoref{supp:mixing} we show how our models learns to some extent to interpolate between different domain even without having ever received explicit supervision for it
In \autoref{supp:results_feature_wise} and \autoref{supp:latent_space_analysis} we provide some intuition on how the network learns parametric transformations.
In \autoref{supp:domain_generalization} we show some interesting domain generalization result on image acquired by a phone. 
Finally we conclude with some implementation details, \autoref{supp:implementation_details}, and ablation studies on the better way of injecting $p$ in the generator and discriminator networks \autoref{supp:concatenation}.

\subsection{Datasets} \label{supp:datasets}

\newlength{\datasetExamplesLength}
\setlength{\datasetExamplesLength}{0.16\linewidth}
\begin{figure}[th!]
    \setlength{\tabcolsep}{1pt}
    \centering
      \begin{tabular}{c|c}
        \includegraphics[height=2.5cm]{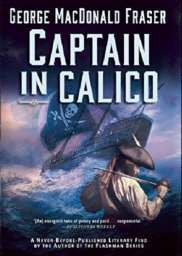}
        \includegraphics[height=2.5cm]{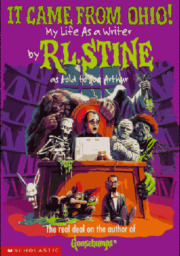}
        \includegraphics[height=2.5cm]{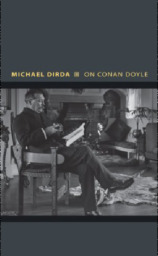}
        &  \includegraphics[height=2.5cm]{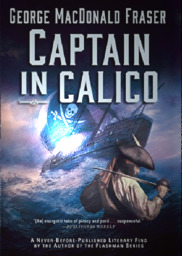}
        \includegraphics[height=2.5cm]{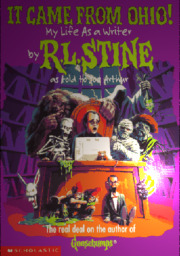}
        \includegraphics[height=2.5cm]{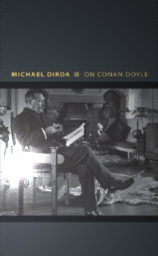}\\
    \end{tabular}
    
    \begin{tabular}{cccccc}
    \begin{overpic}[width=\datasetExamplesLength]{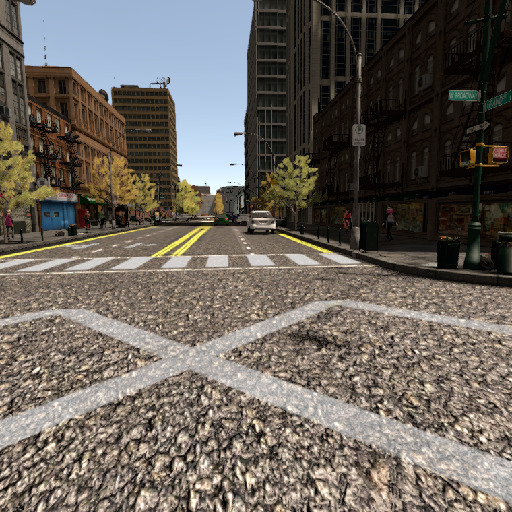}
        \put (2,2) {\textbf{\textcolor{white}{Summer}}}
    \end{overpic} &
    \begin{overpic}[width=\datasetExamplesLength]{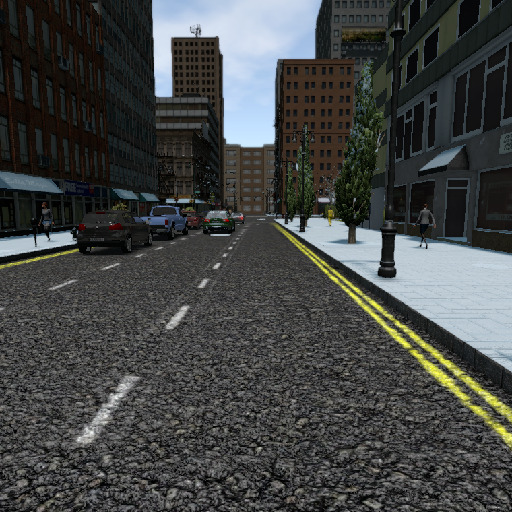}
        \put (2,2) {\textbf{\textcolor{white}{Winter}}}
    \end{overpic} &
    \begin{overpic}[width=\datasetExamplesLength]{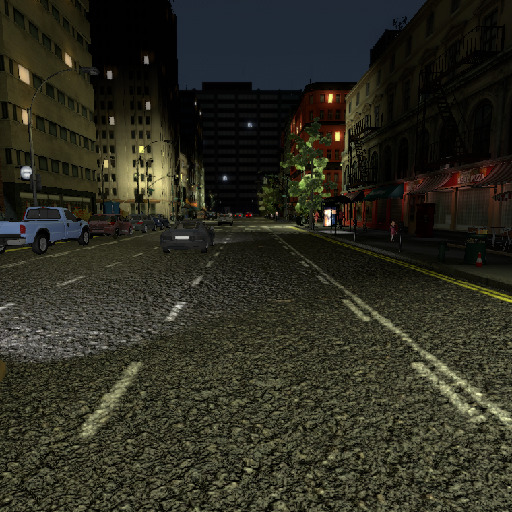}
        \put (2,2) {\textbf{\textcolor{white}{Night}}}
    \end{overpic} &
    \begin{overpic}[width=\datasetExamplesLength]{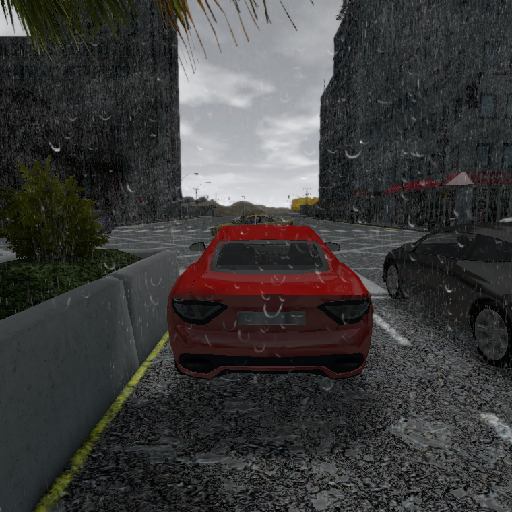}
        \put (2,2) {\textbf{\textcolor{white}{Softrain}}}
    \end{overpic} &
    \begin{overpic}[width=\datasetExamplesLength]{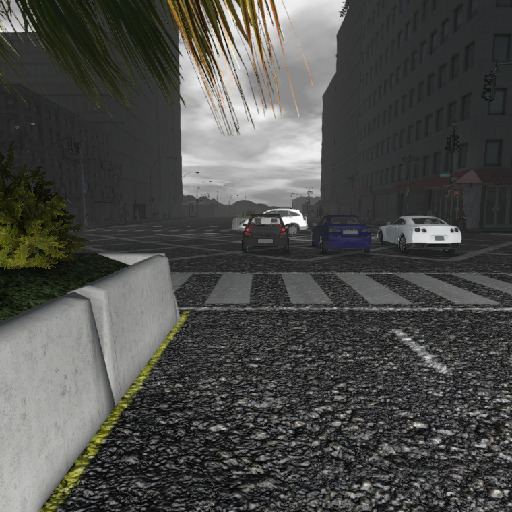}
        \put (2,2) {\textbf{\textcolor{white}{Fog}}}
    \end{overpic} &
    \begin{overpic}[width=\datasetExamplesLength]{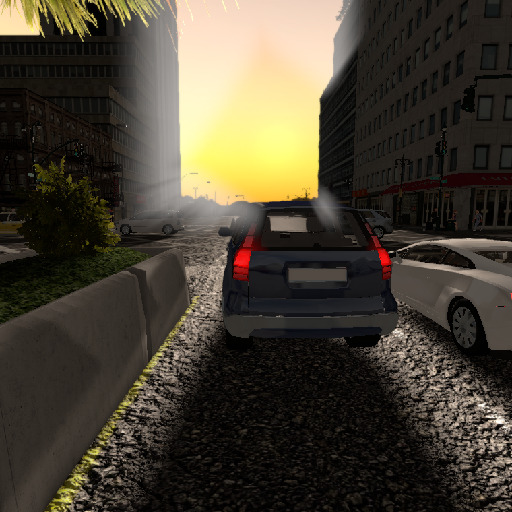}
        \put (2,2) {\textbf{\textcolor{white}{Sunset}}}
    \end{overpic}
    \end{tabular}
    
    \begin{tabular}{ccc|ccc}
    \includegraphics[width=\datasetExamplesLength]{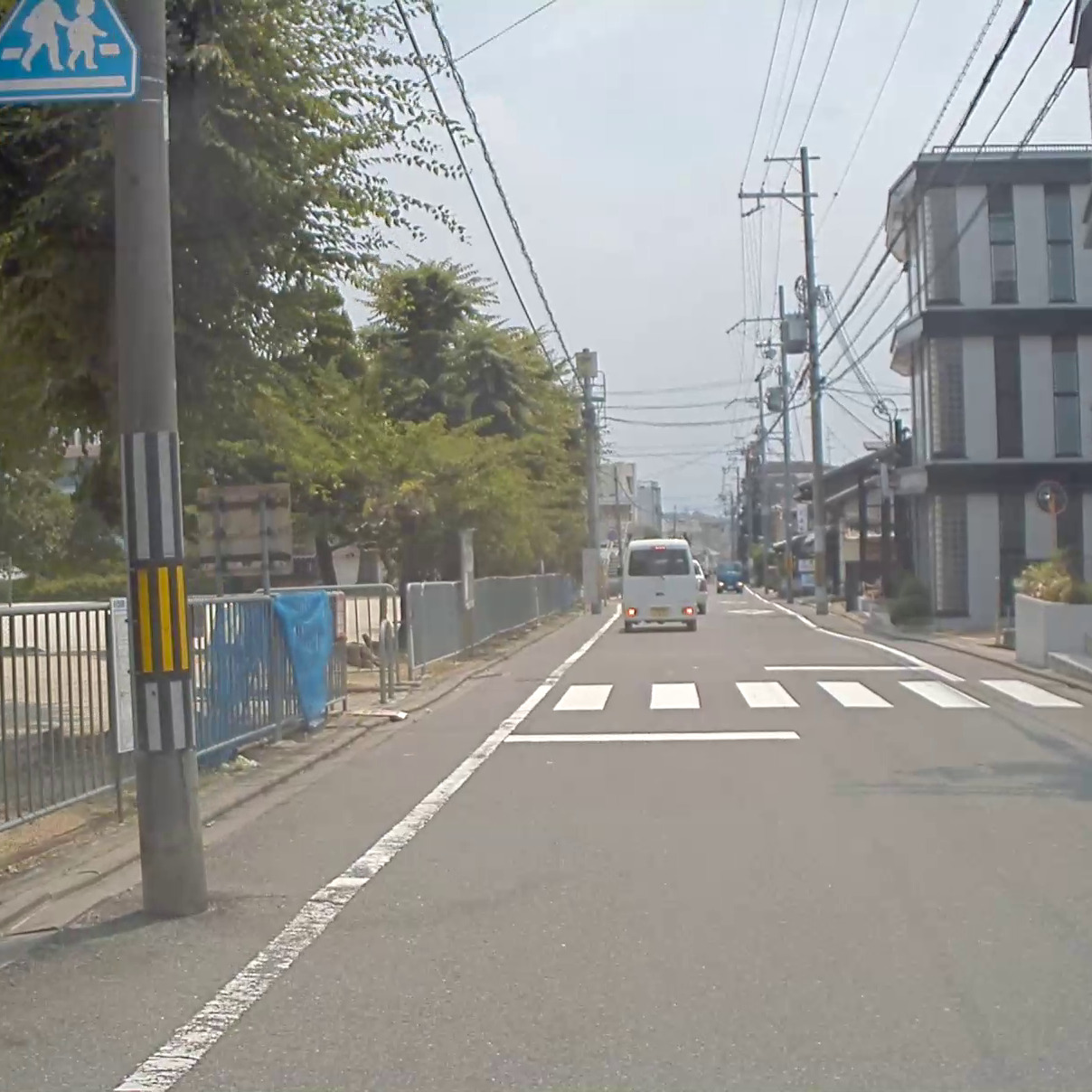} &
    \includegraphics[width=\datasetExamplesLength]{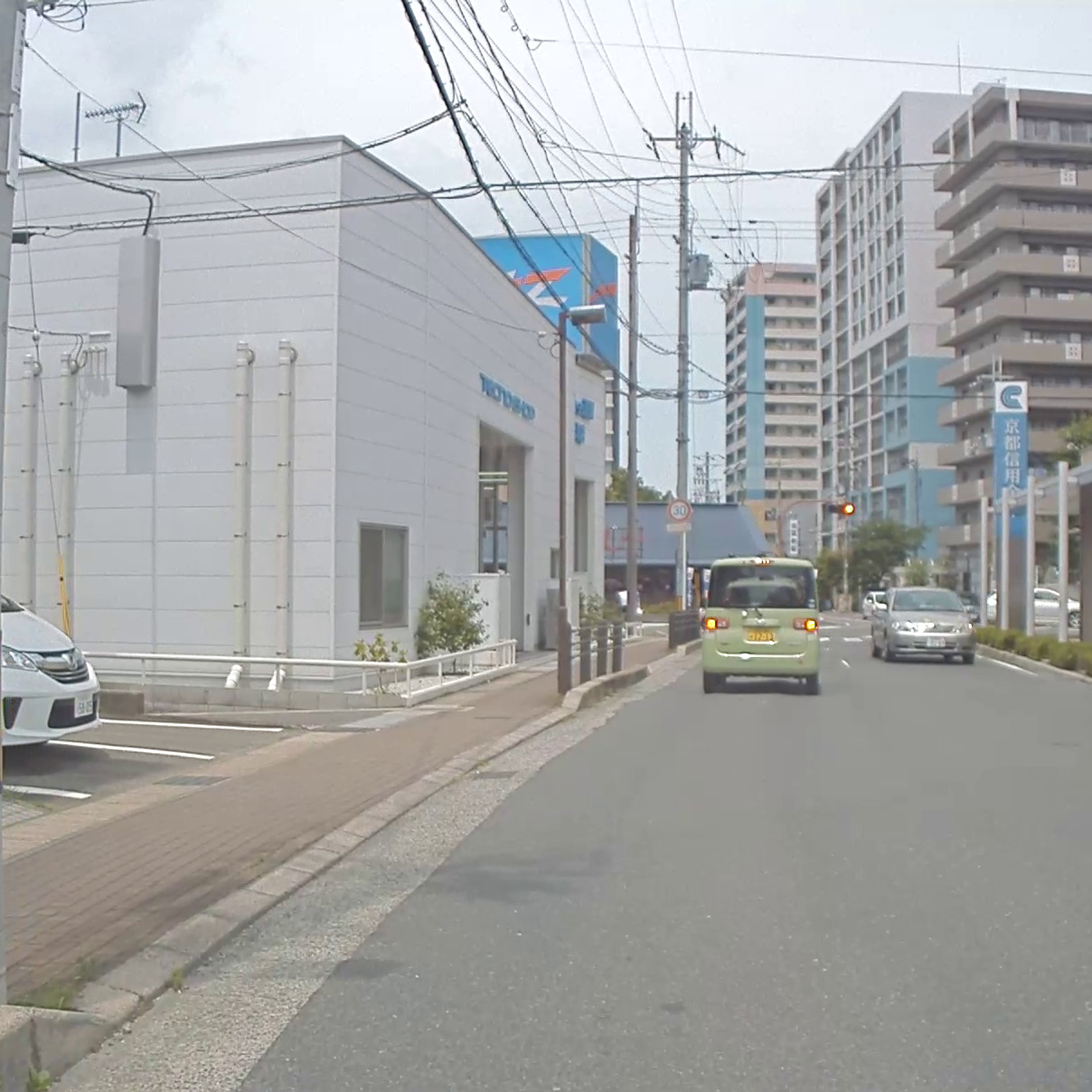} &
    \includegraphics[width=\datasetExamplesLength]{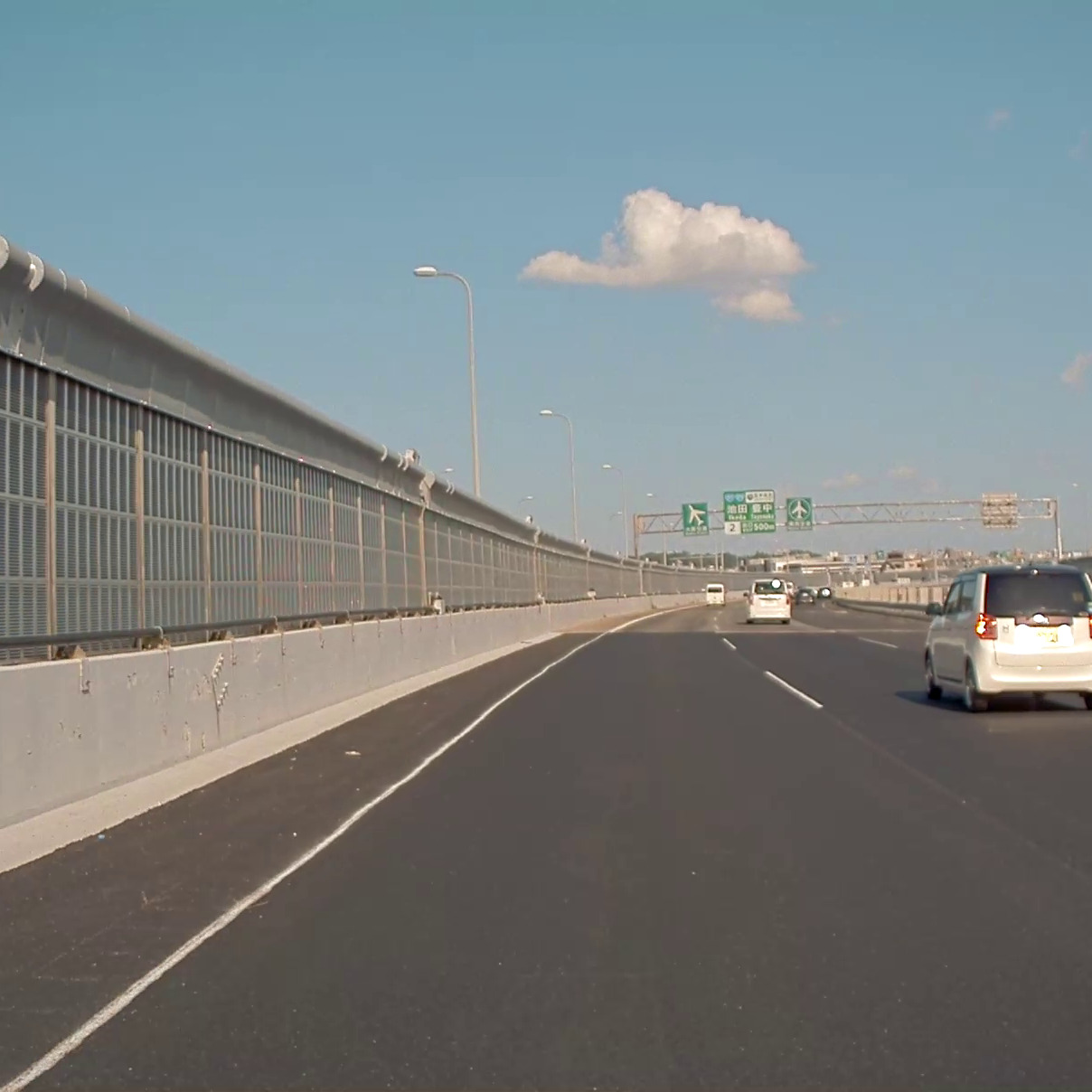} &
        
    \includegraphics[width=\datasetExamplesLength]{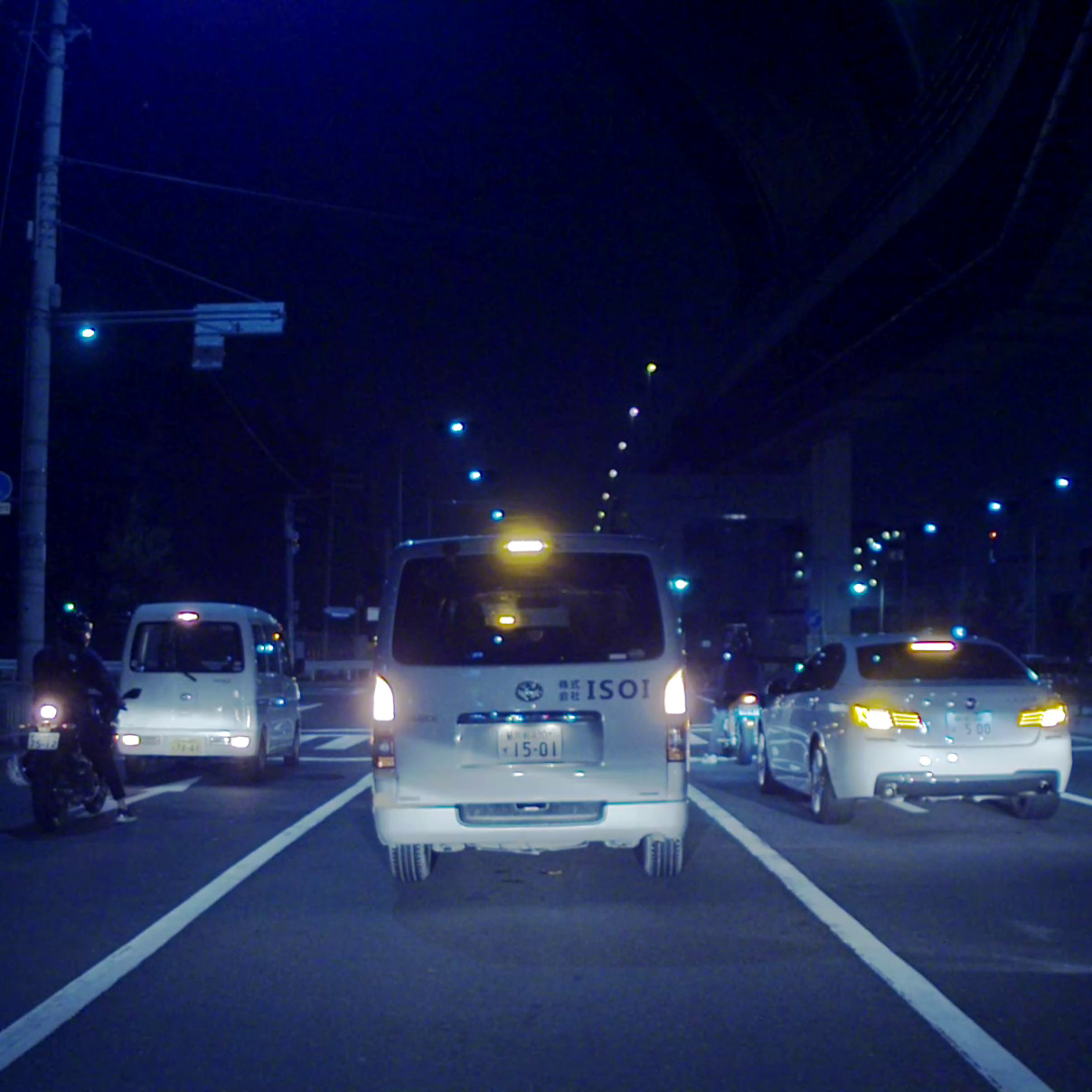} &
    \includegraphics[width=\datasetExamplesLength]{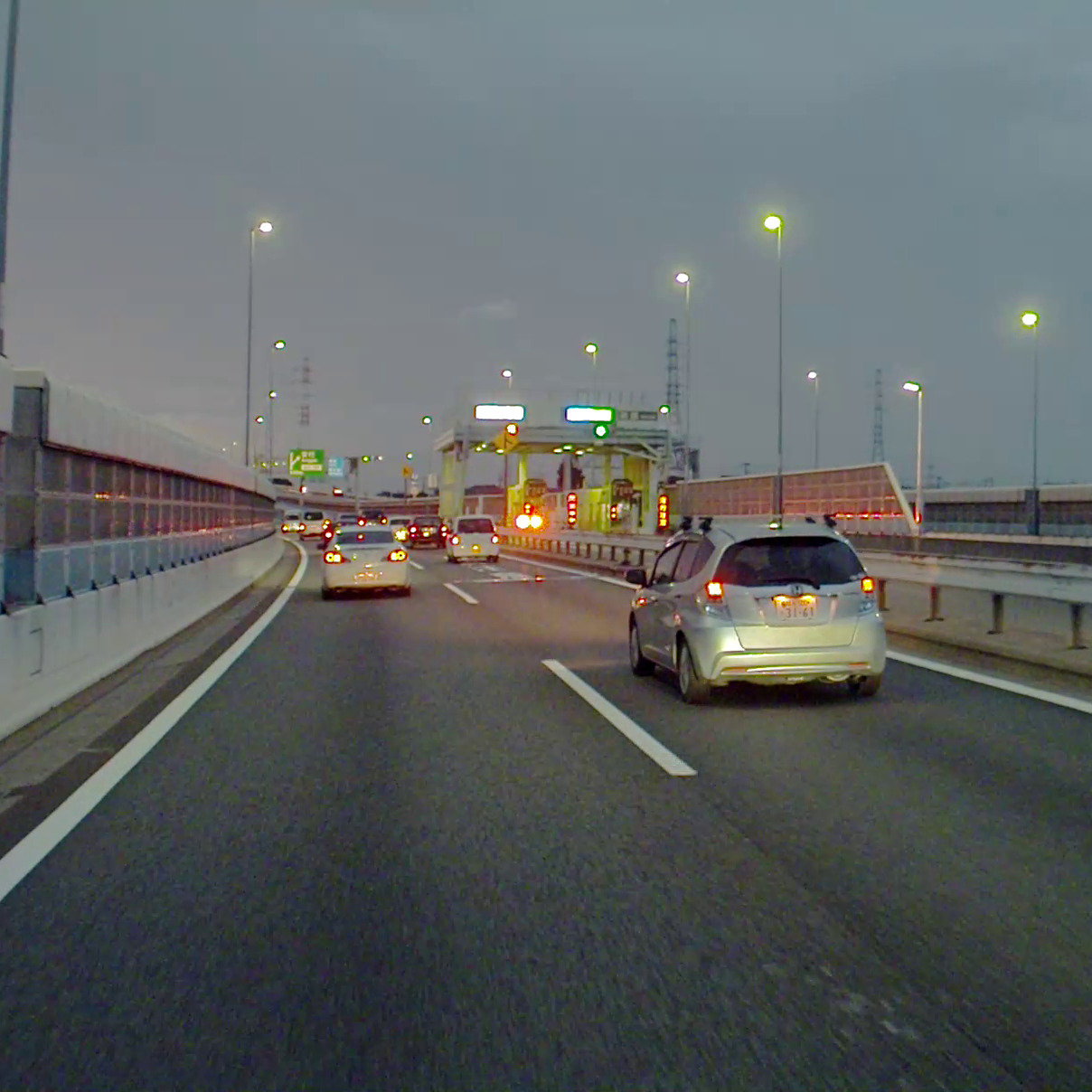} &
    \includegraphics[width=\datasetExamplesLength]{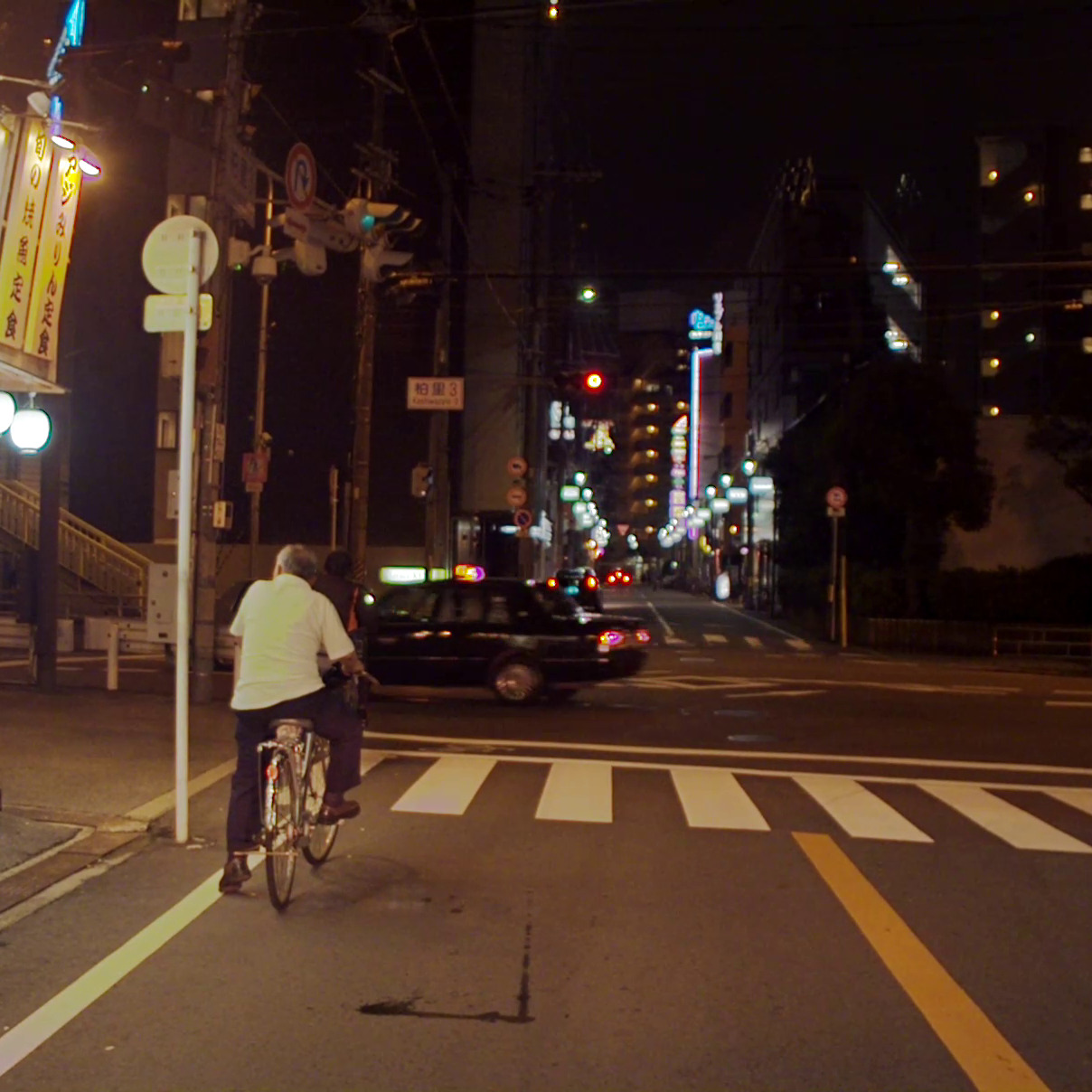}
    \end{tabular}
    \caption{Examples from different datasets, one per row: (1) \bookCovers{} \cite{iwana2016judging} (left for original images and right for image rendered with our light transformations), (2) \synthia{} \cite{7780721}, and (3) \init{} \cite{shen2019towards} (left for day and right for night).}
    \label{fig:dataset_examples}
\end{figure}

We explore the capabilities of our model in two different tasks: learning a transformation guided by an explicit parametrization and domain translation using \textit{soft labels}. 
We use both real and synthetic data. 
In \autoref{fig:dataset_examples} we show several samples from the datasets that we used for training.

\clearpage
\subsection{Quantitative analysis of explicit parameterization} \label{supp:results_explicit}

In order to measure the quality of our learned transformations, we measure the similarity (in terms of \FID{} and \LPIPS{} between a ground truth image with known transformation properties and the same book cover with varying values in $p$. 
In Tables \ref{tab:book_fid_lpips_x} and \ref{tab:book_fid_lpips_illumination} we report the results when changing the values of the x and illumination axes for \bookCovers{}.

\renewcommand{\bookFidLpipsScale}{0.85}

\begin{table}[h!]
\center
\scalebox{\bookFidLpipsScale}{
\begin{tabular}{ccc}
\begin{tabular}{l@{\hspace{.8\tabcolsep}}c|ccccc}
\toprule
& & \multicolumn{5}{c}{\textbf{Gen. images}}\\
 &\textbf{x}& 0.1 & 0.3 & 0.5 & 0.7 & 0.9 \\
 \hline
 \multirow{5}{*}{\rotatebox[]{90}{\textbf{Real images}}}
& 0.1 & \textbf{41.739} & \textbf{47.339} & 52.822 & 48.888 & 46.95  \\
& 0.3 & 50.613 & 48.411 & 48.506 & 52.25  & 53.345 \\
& 0.5 & 52.887 & 51.408 & \textbf{46.358} & 50.455 & 54.313 \\
& 0.7 & 52.763 & 54.878 & 49.369 & \textbf{41.152} & 42.421 \\
& 0.9 & 51.103 & 55.345 & 53.384 & 45.096 & \textbf{38.200} \\
\bottomrule
\end{tabular} 
& &
\begin{tabular}{l@{\hspace{.8\tabcolsep}}c|ccccc}
\toprule
& & \multicolumn{5}{c}{\textbf{Gen. images}}\\
 &\textbf{x}& 0.1 & 0.3 & 0.5 & 0.7 & 0.9 \\
 \hline
 \multirow{5}{*}{\rotatebox[]{90}{\textbf{Real images}}}
& 0.1 & \textbf{0.062} & 0.076 & 0.102 & 0.120  & 0.125 \\
& 0.3 & 0.080  & \textbf{0.061} & 0.080  & 0.103 & 0.115 \\
& 0.5 & 0.106 & 0.085 & \textbf{0.065} & 0.087 & 0.104 \\
& 0.7 & 0.114 & 0.101 & 0.076 & \textbf{0.060}  & 0.074 \\
& 0.9 & 0.117 & 0.107 & 0.090  & 0.066 & \textbf{0.054}\\
\bottomrule
\end{tabular} \\
(a) \FID{} &  & (b) \LPIPS{} \\
\end{tabular}
}
\caption{Comparison between real and generated images for the \bookCovers{} dataset, keeping y and illumination fixed and varying the x component of $p$. The more dissimilar the values of x, the higher the distance. In \textbf{bold} the smallest value column-wise.}
\label{tab:book_fid_lpips_x}
\end{table}

\begin{table}[h!]
\center
\scalebox{\bookFidLpipsScale}{
\begin{tabular}{ccc}
\begin{tabular}{l@{\hspace{.8\tabcolsep}}c|ccccc}
\toprule
& & \multicolumn{5}{c}{\textbf{Gen. images}}\\
 &\textbf{ill.}& 0.1 & 0.3 & 0.5 & 0.7 & 0.9 \\
 \hline
 \multirow{5}{*}{\rotatebox[]{90}{\textbf{Real images}}}
& 0.1 & \textbf{38.434} & 43.924 & 54.309 & 60.664 & 64.754 \\
& 0.3 & 53.696 & 42.526 & 45.665 & 49.593 & 53.514 \\
& 0.5 & 50.360  & \textbf{35.291} & \textbf{34.303} & \textbf{36.321} & \textbf{38.541} \\
& 0.7 & 58.330  & 42.100   & 38.004 & 39.777 & 42.197 \\
& 0.9 & 69.816 & 51.095 & 46.816 & 45.315 & 46.511 \\
\bottomrule
\end{tabular} 
& &
\begin{tabular}{l@{\hspace{.8\tabcolsep}}c|ccccc}
\toprule
& & \multicolumn{5}{c}{\textbf{Gen. images}}\\
 &\textbf{ill.}& 0.1 & 0.3 & 0.5 & 0.7 & 0.9 \\
 \hline
 \multirow{5}{*}{\rotatebox[]{90}{\textbf{Real images}}}
& 0.1 & \textbf{0.063} & 0.08  & 0.121 & 0.15  & 0.168 \\
& 0.3 & 0.107 & \textbf{0.065} & 0.073 & 0.09  & 0.105 \\
& 0.5 & 0.139 & 0.069 & \textbf{0.053} & 0.058 & 0.067 \\
& 0.7 & 0.171 & 0.089 & 0.06  & \textbf{0.055} & \textbf{0.059} \\
& 0.9 & 0.196 & 0.11  & 0.074 & 0.063 & 0.061 \\
\bottomrule
\end{tabular} \\
(a) \FID{} &  & (b) \LPIPS{} \\
\end{tabular}
}
\caption{Comparison between real and generated images for the \bookCovers{} dataset on the illumination component of $p$.}
\label{tab:book_fid_lpips_illumination}
\end{table}

These results complement Tab. 1 in the main paper and show how our generator can correctly apply the desired transformation in most cases.
This is testified by the distances between generated and real images being usually lower when the parametrization between the two sets matches (\ie, values along the diagonal).

\subsection{Disentanglement}
\label{supp:disentanglement}
In \autoref{fig:supplement_book_covers} we show the disentangling capabilities of our model applied to the \bookCovers{} dataset. 
Each axis can be modified independently with no effect over the others, and arbitrary combinations are also feasible.

\newlength{\supplementBookCovers}
\setlength{\supplementBookCovers}{0.16\textwidth}
\begin{figure}[th!]
    \setlength{\tabcolsep}{2pt}
    \centering
    \begin{tabular}{c|cccccc}
        \rotatebox{90}{{\scriptsize \textit{Change x}}} &
        \includegraphics[width=\supplementBookCovers]{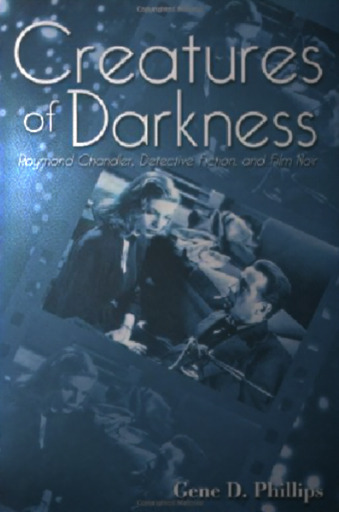} &
        \includegraphics[width=\supplementBookCovers]{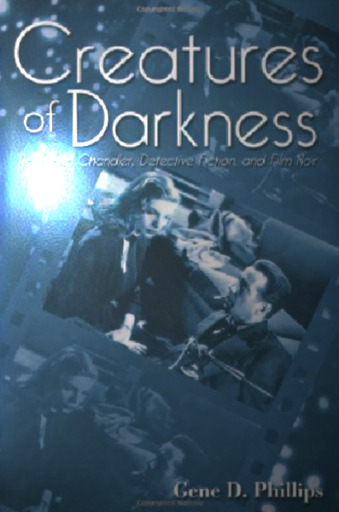} &
        \includegraphics[width=\supplementBookCovers]{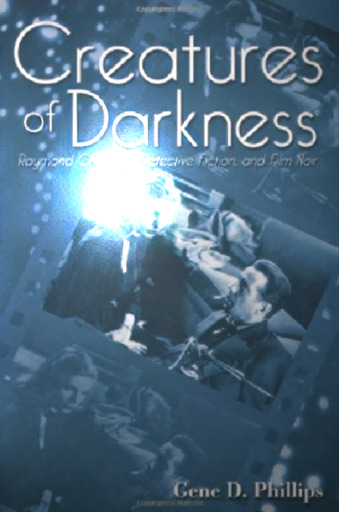} &
        \includegraphics[width=\supplementBookCovers]{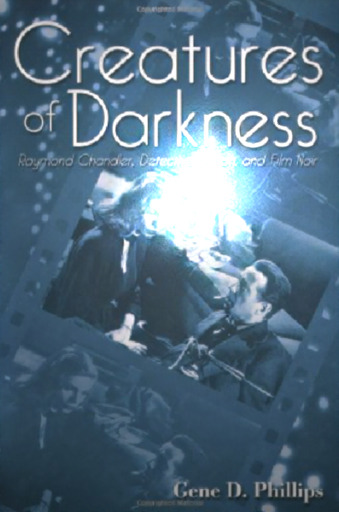} &
        \includegraphics[width=\supplementBookCovers]{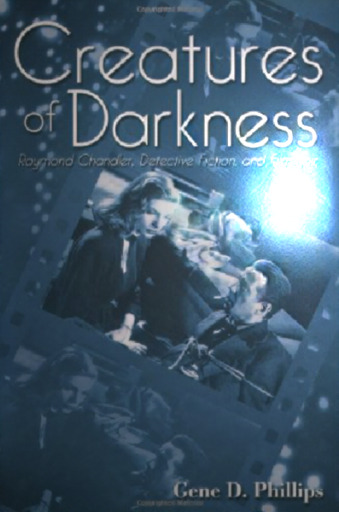} &
        \includegraphics[width=\supplementBookCovers]{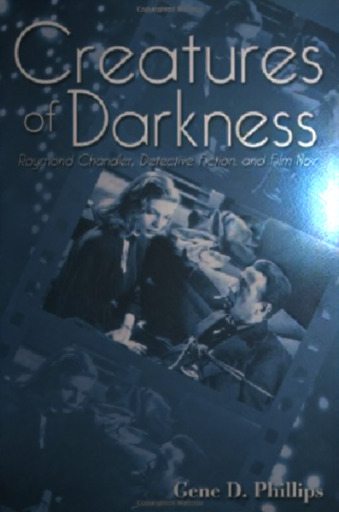}\\
        
        \rotatebox{90}{{\scriptsize \textit{Change y}}} &
        \includegraphics[width=\supplementBookCovers]{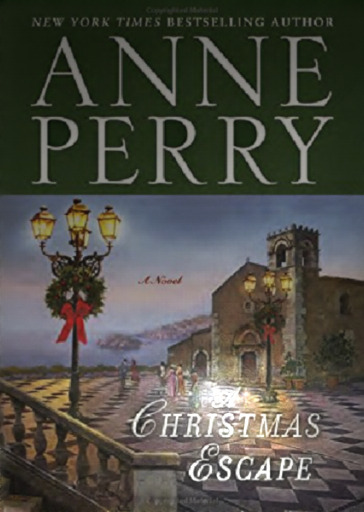} &
        \includegraphics[width=\supplementBookCovers]{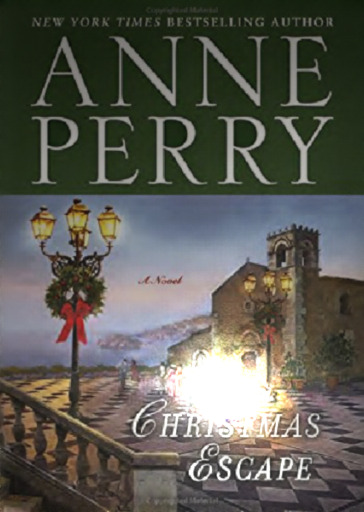} &
        \includegraphics[width=\supplementBookCovers]{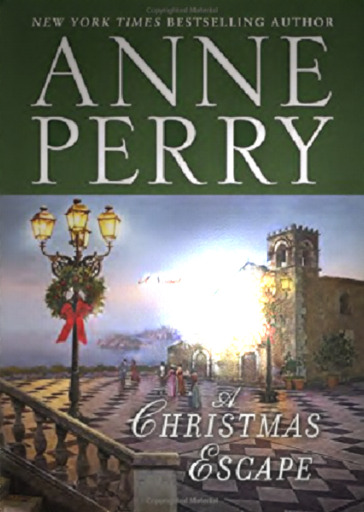} &
        \includegraphics[width=\supplementBookCovers]{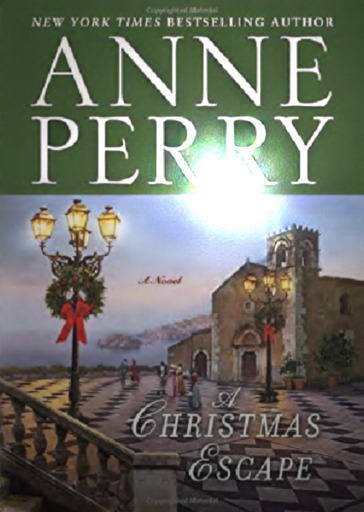} &
        \includegraphics[width=\supplementBookCovers]{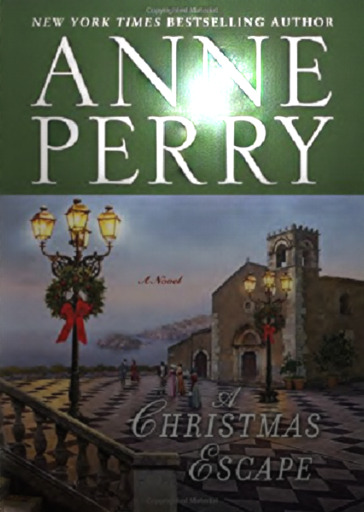} &
        \includegraphics[width=\supplementBookCovers]{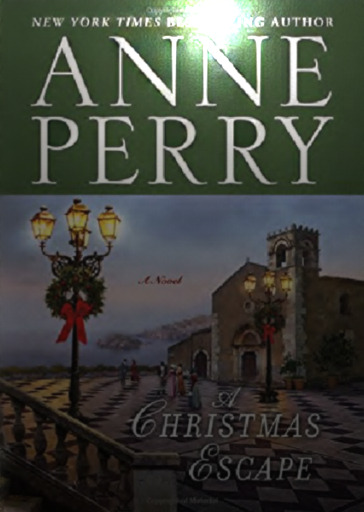}\\
        
        \rotatebox{90}{{\scriptsize \textit{Change illumination}}} &
        \includegraphics[width=\supplementBookCovers]{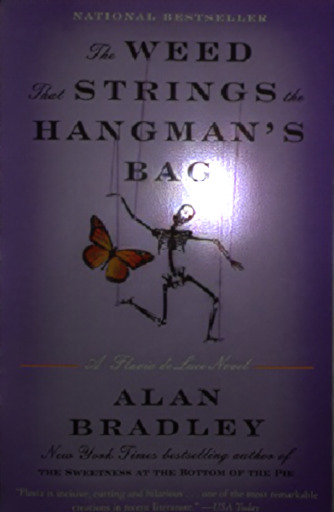} &
        \includegraphics[width=\supplementBookCovers]{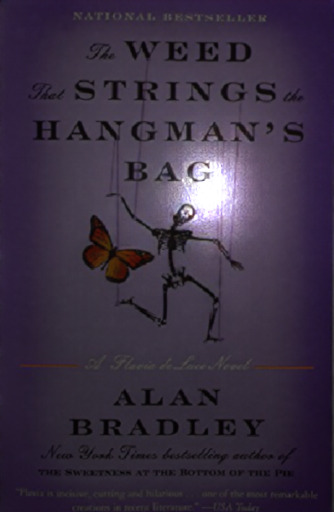} &
        \includegraphics[width=\supplementBookCovers]{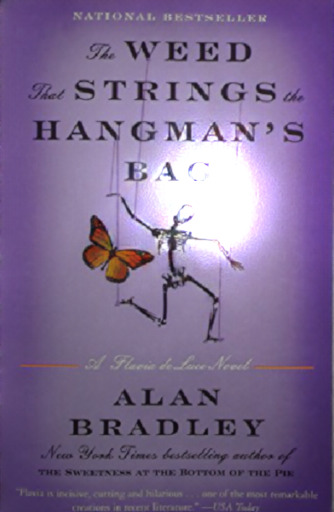} &
        \includegraphics[width=\supplementBookCovers]{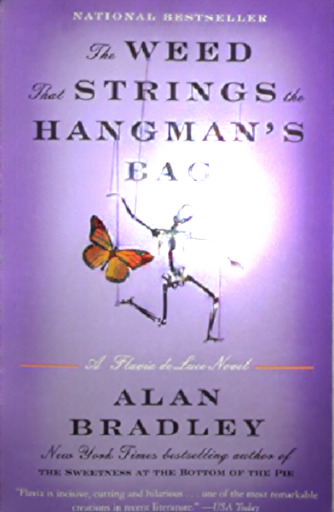} &
        \includegraphics[width=\supplementBookCovers]{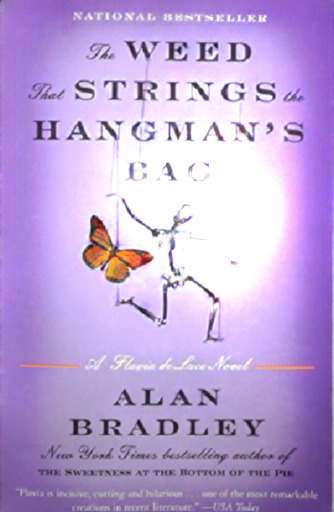} &
        \includegraphics[width=\supplementBookCovers]{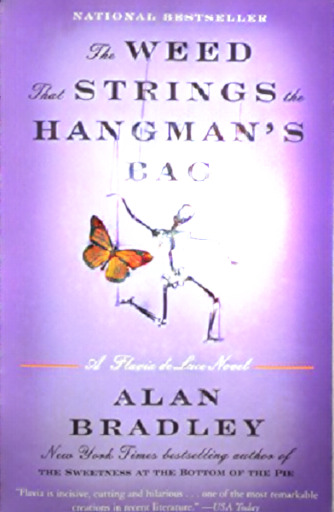}\\

        \rotatebox{90}{{\scriptsize \textit{Change all}}} &
        \includegraphics[width=\supplementBookCovers]{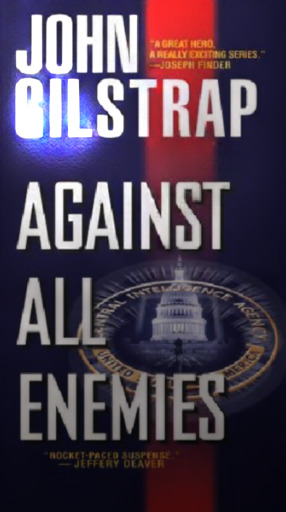} &
        \includegraphics[width=\supplementBookCovers]{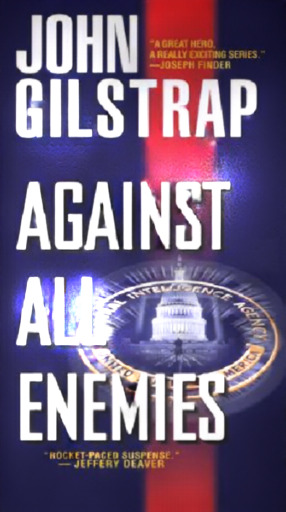} &
        \includegraphics[width=\supplementBookCovers]{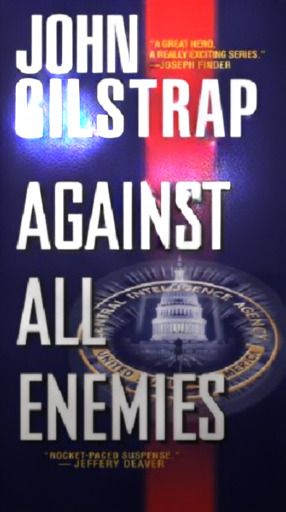} &
        \includegraphics[width=\supplementBookCovers]{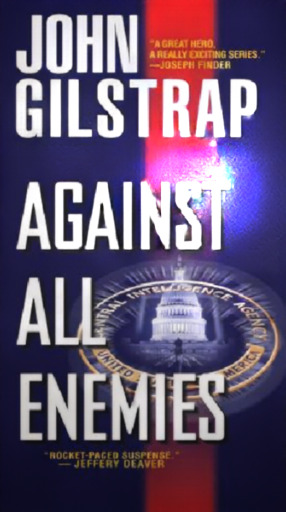} &
        \includegraphics[width=\supplementBookCovers]{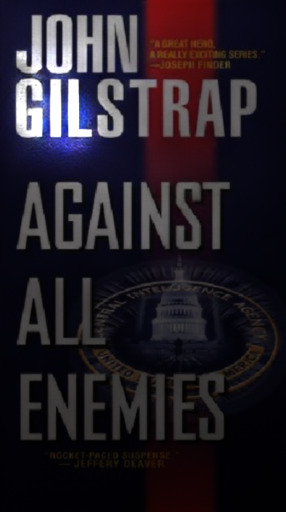} &
        \includegraphics[width=\supplementBookCovers]{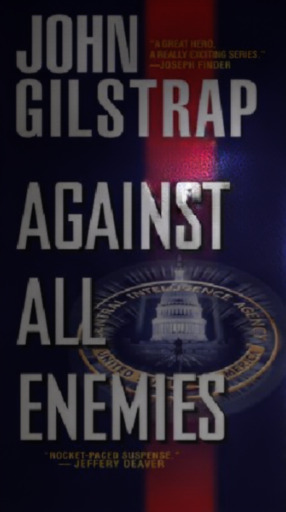}\\
    \end{tabular}
    \caption{
    Additional results on the \bookCovers{} dataset. The first three rows show disentangled transformations, where two out of three axes are kept fixed. The last row contains entangled transformations, where all axes are modified simultaneously.
    }
    \label{fig:supplement_book_covers}
\end{figure}

\clearpage
\subsection{Results for the soft-parametrization} \label{supp:results_soft}

\begin{figure}[h!]
\setlength{\tabcolsep}{3pt}
\centering
\begin{tabular}{c|ccccccc}
    \toprule
    Source & & 0.0 & 0.25 & 0.5 & 0.75 & 1.0 \\ 
    \midrule
    \includegraphics[width=0.15\linewidth]{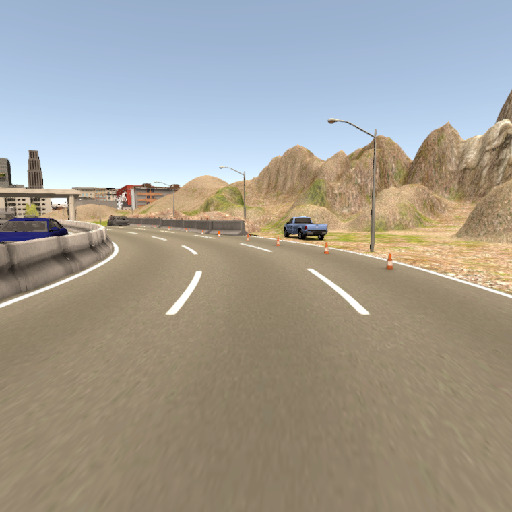}& \rotatebox{90}{\scriptsize\summertowinter} & 
    \includegraphics[width=0.15\linewidth]{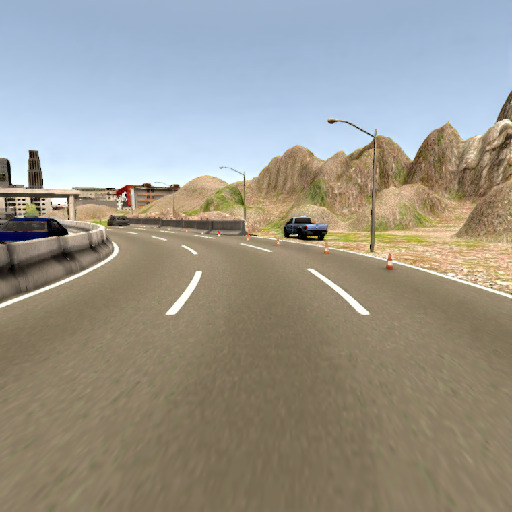}&
    \includegraphics[width=0.15\linewidth]{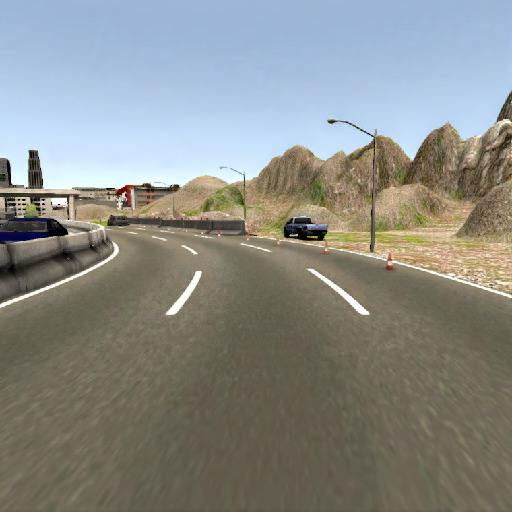}&
    \includegraphics[width=0.15\linewidth]{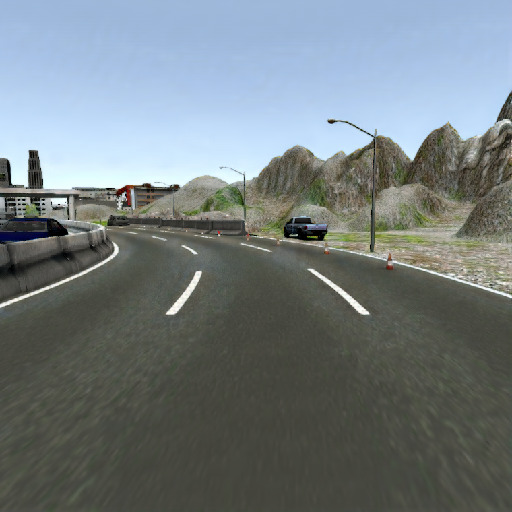}&
    \includegraphics[width=0.15\linewidth]{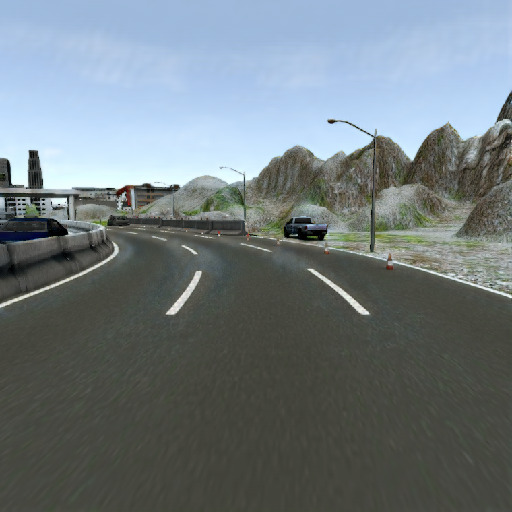}&
    \includegraphics[width=0.15\linewidth]{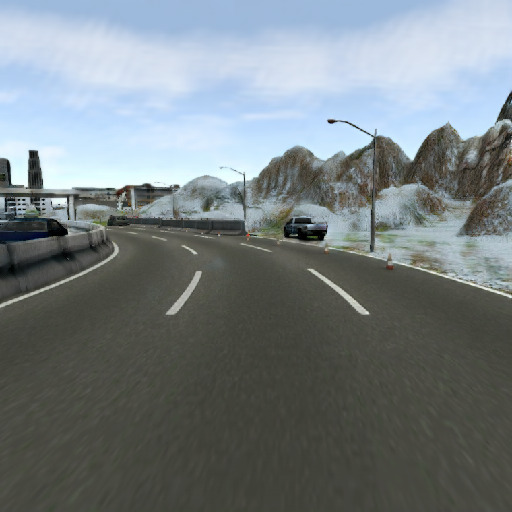}\\
    
    \includegraphics[width=0.15\linewidth]{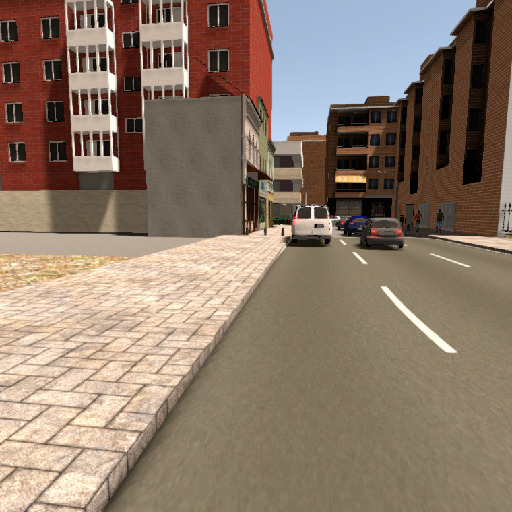}& \raisebox{2mm}{\rotatebox{90}{\scriptsize\summertofog}} &
    \includegraphics[width=0.15\linewidth]{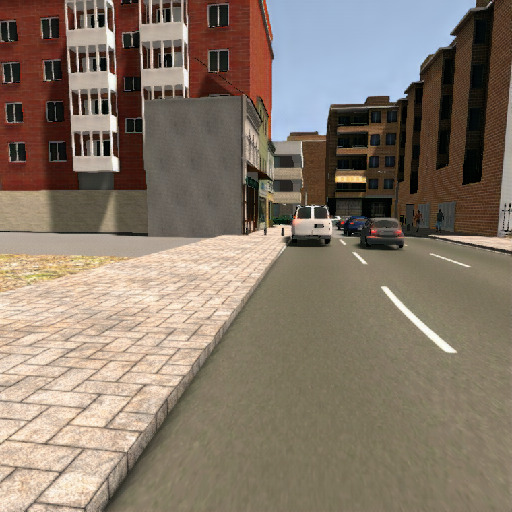}&
    \includegraphics[width=0.15\linewidth]{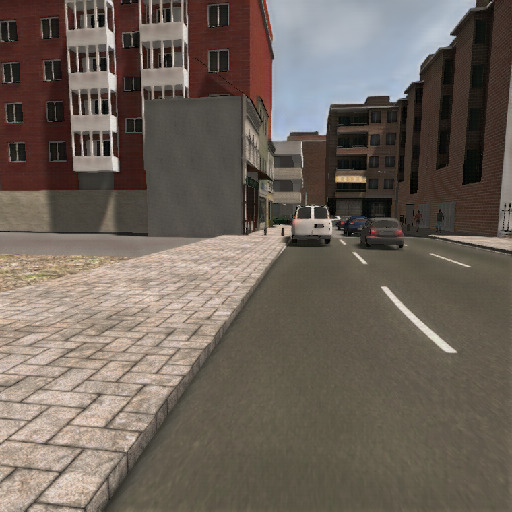}&
    \includegraphics[width=0.15\linewidth]{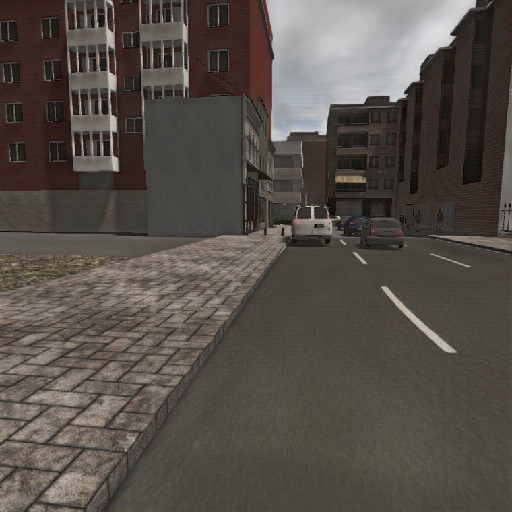}&
    \includegraphics[width=0.15\linewidth]{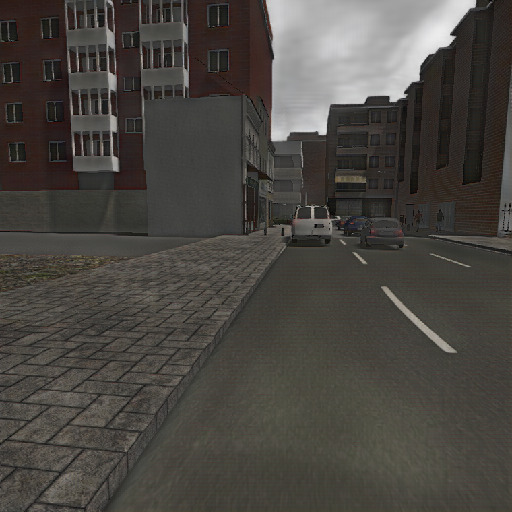}&
    \includegraphics[width=0.15\linewidth]{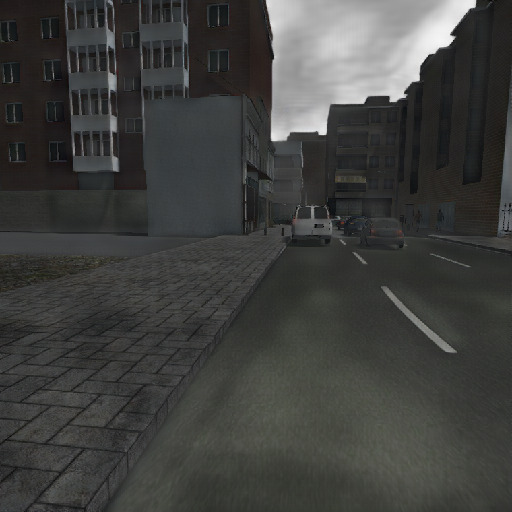}\\
    
    \includegraphics[width=0.15\linewidth]{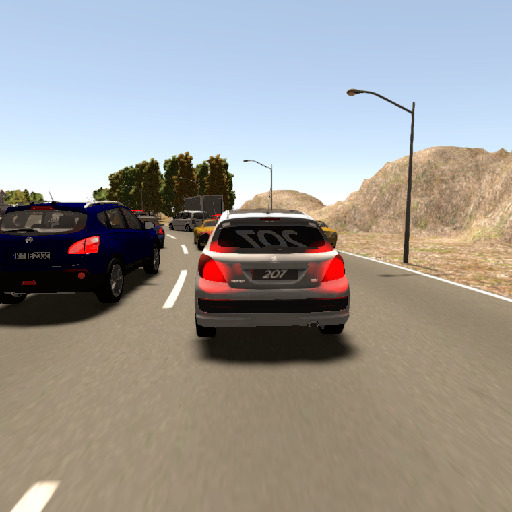}& \raisebox{1mm}{\rotatebox{90}{\scriptsize\summertosunset}} &
    \includegraphics[width=0.15\linewidth]{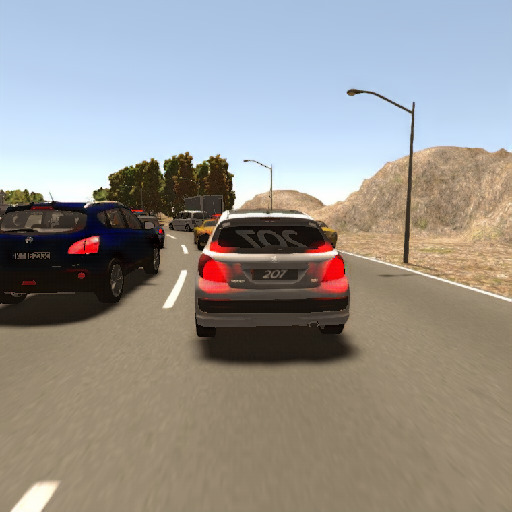}&
    \includegraphics[width=0.15\linewidth]{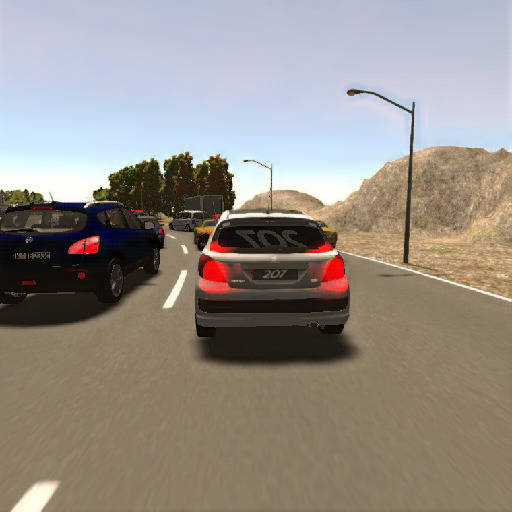}&
    \includegraphics[width=0.15\linewidth]{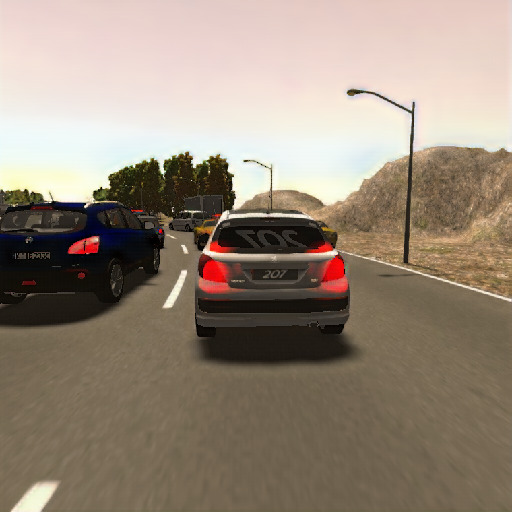}&
    \includegraphics[width=0.15\linewidth]{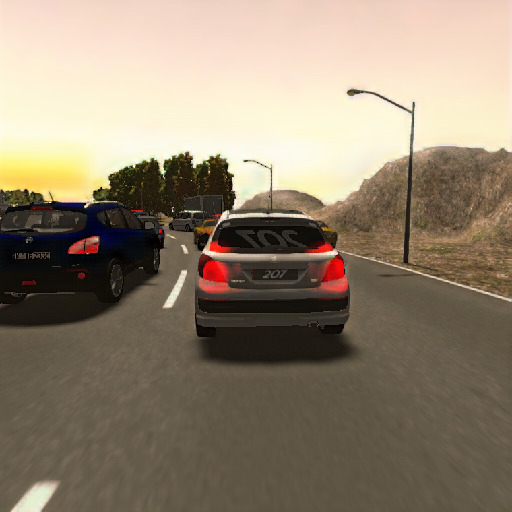}&
    \includegraphics[width=0.15\linewidth]{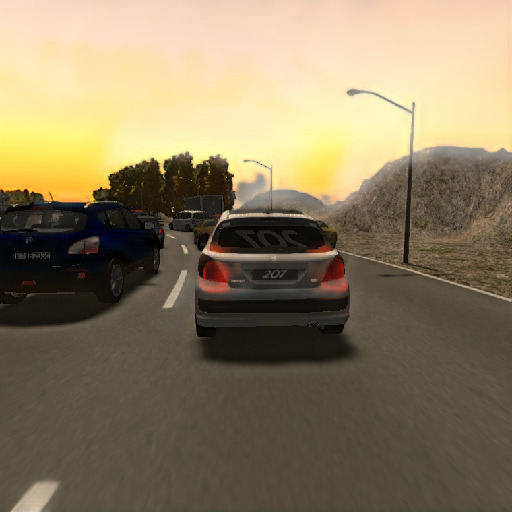}\\
    
    \includegraphics[width=0.15\linewidth]{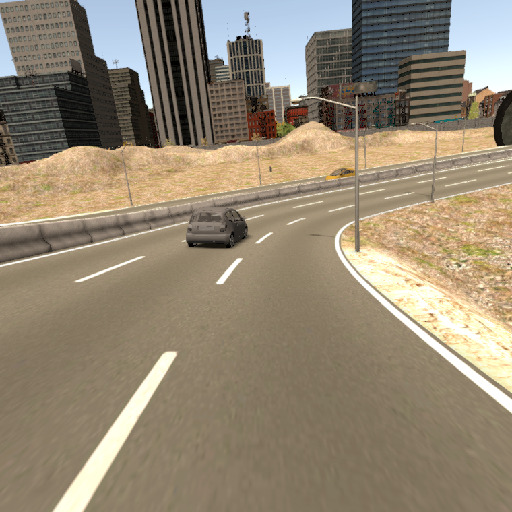}& \rotatebox{90}{\scriptsize\summertosoftrain} &
    \includegraphics[width=0.15\linewidth]{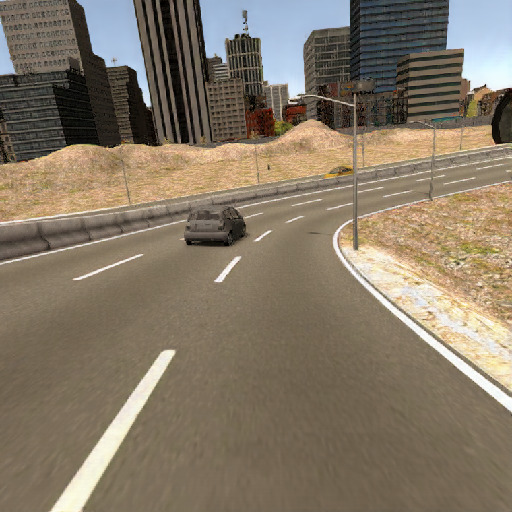}&
    \includegraphics[width=0.15\linewidth]{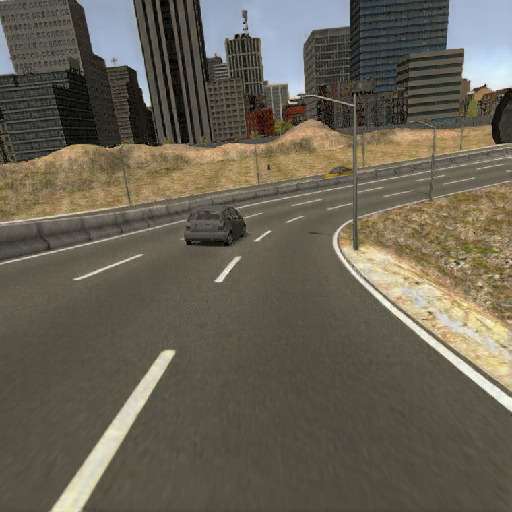}&
    \includegraphics[width=0.15\linewidth]{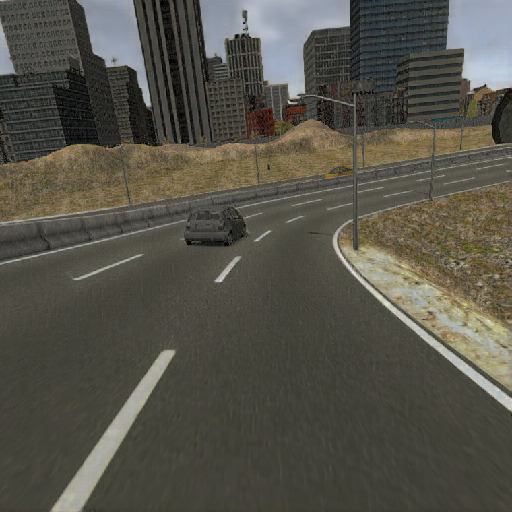}&
    \includegraphics[width=0.15\linewidth]{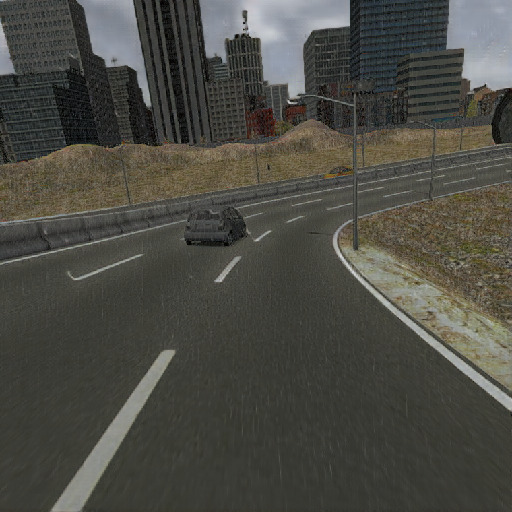}&
    \includegraphics[width=0.15\linewidth]{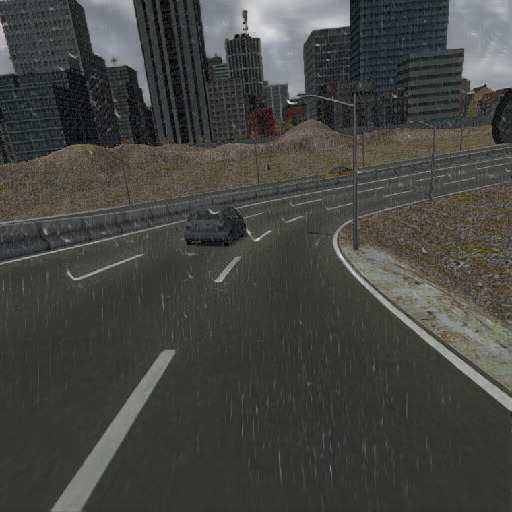}\\
\end{tabular}
\caption{Evaluation on different tasks for \synthia{}.}
\label{fig:experimental_synthia}
\end{figure}

We show in \autoref{fig:experimental_synthia} generated images for the \synthia{} dataset. 
We use the \textit{summer} subset as source domain and learn multiple transformations (\summertowinter, \summertofog, \summertosunset, \summertosoftrain).

\begin{figure}[h!]
    \centering
    \newlength{\imagesize}
    \setlength{\imagesize}{0.16\linewidth}
    \setlength{\tabcolsep}{2pt}
    \begin{tabular}{c|ccccc}
    \toprule
    Source & 0.0 & 0.25 & 0.5 & 0.75 & 1.0 \\ 
    \midrule
    
    \includegraphics[width=\imagesize]{images/datasets/init-4-test/32_00160-original}&
    \includegraphics[width=\imagesize]{images/datasets/init-4-test/32_00160-0.00}& \includegraphics[width=\imagesize]{images/datasets/init-4-test/32_00160-0.25}&
    \includegraphics[width=\imagesize]{images/datasets/init-4-test/32_00160-0.50}&
    \includegraphics[width=\imagesize]{images/datasets/init-4-test/32_00160-0.75}&
    \includegraphics[width=\imagesize]{images/datasets/init-4-test/32_00160-1.00}\\

    \includegraphics[width=\imagesize]{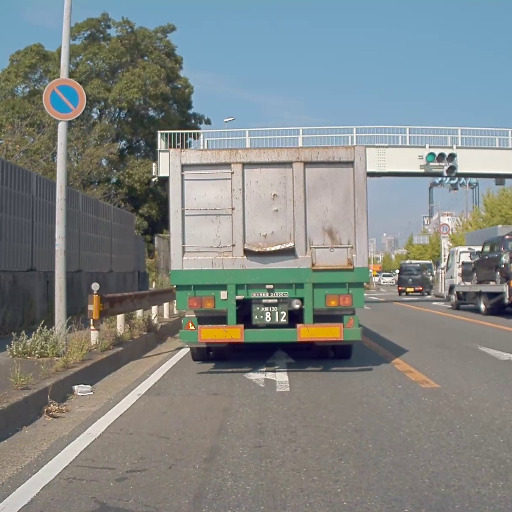}&
    \includegraphics[width=\imagesize]{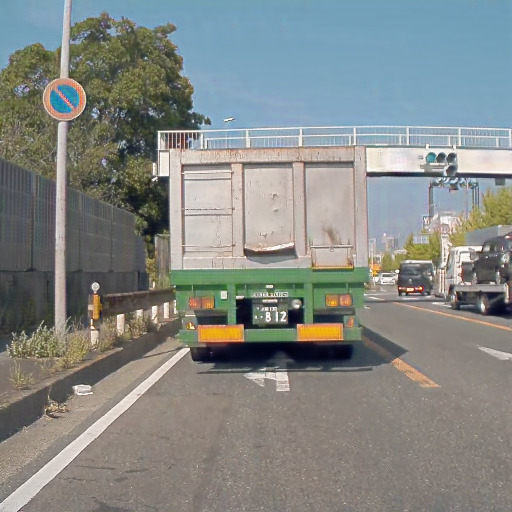}& \includegraphics[width=\imagesize]{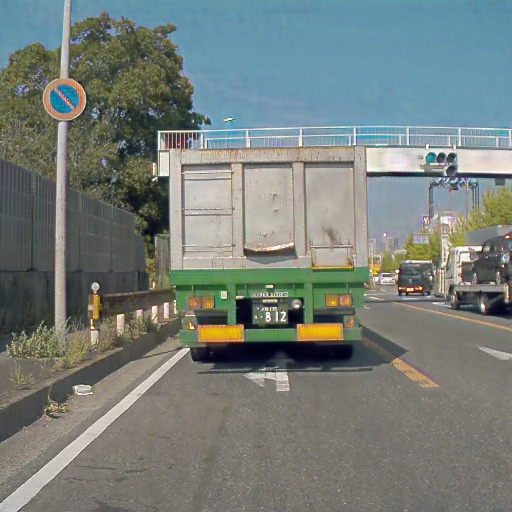}&
    \includegraphics[width=\imagesize]{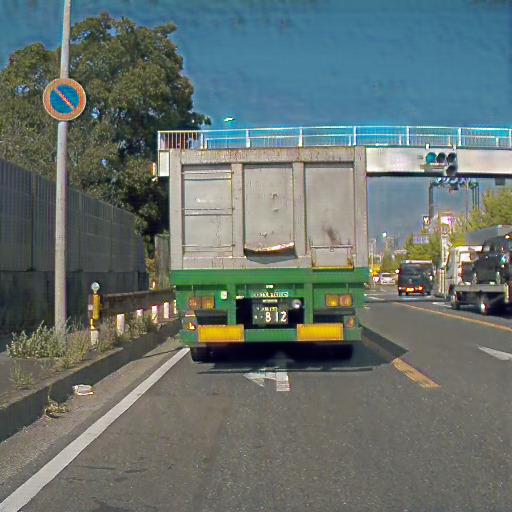}&
    \includegraphics[width=\imagesize]{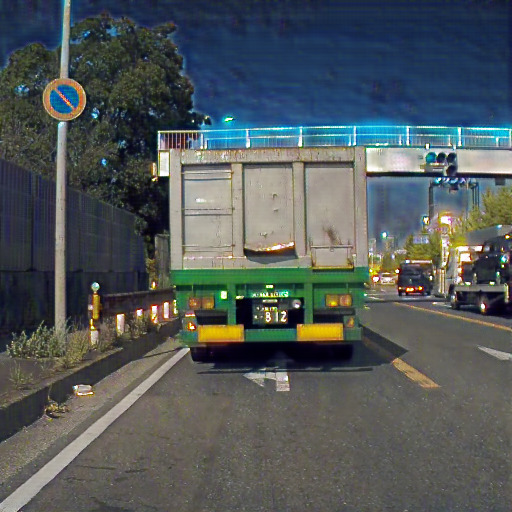}&
    \includegraphics[width=\imagesize]{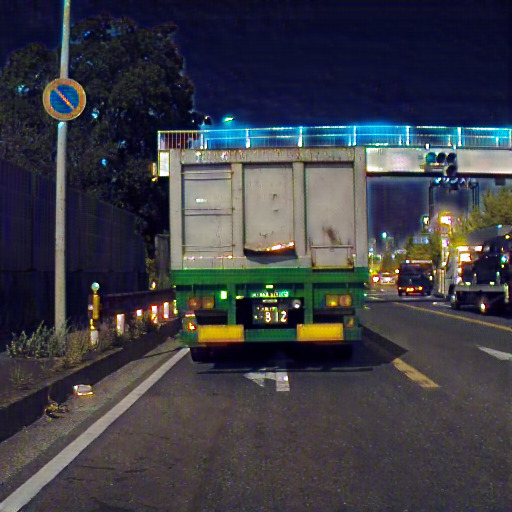}\\

    \includegraphics[width=\imagesize]{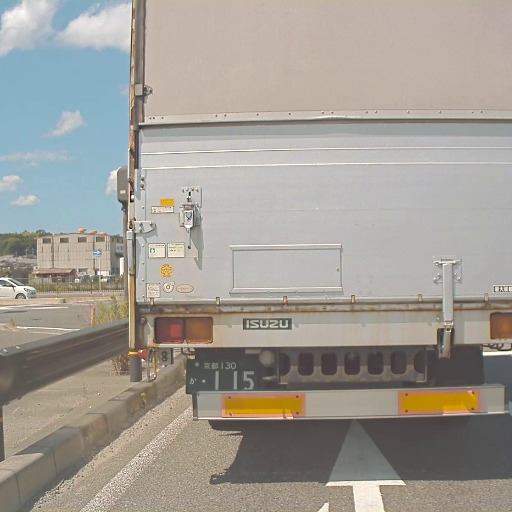} &
    \includegraphics[width=\imagesize]{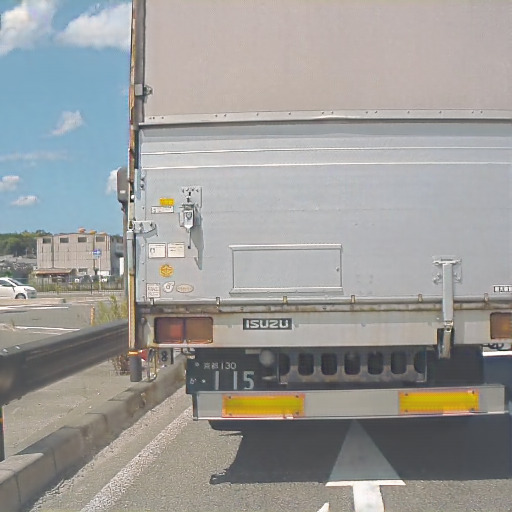} &
    \includegraphics[width=\imagesize]{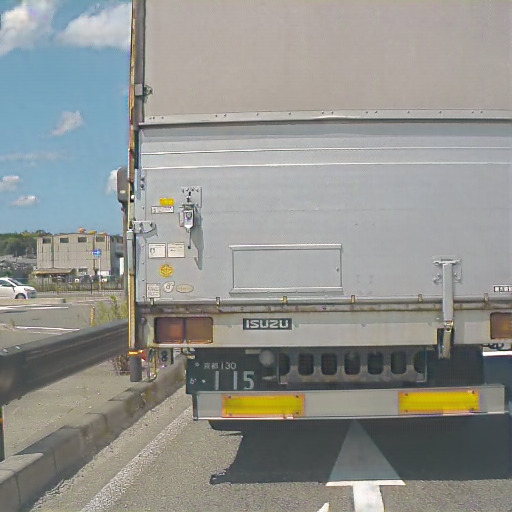} &
    \includegraphics[width=\imagesize]{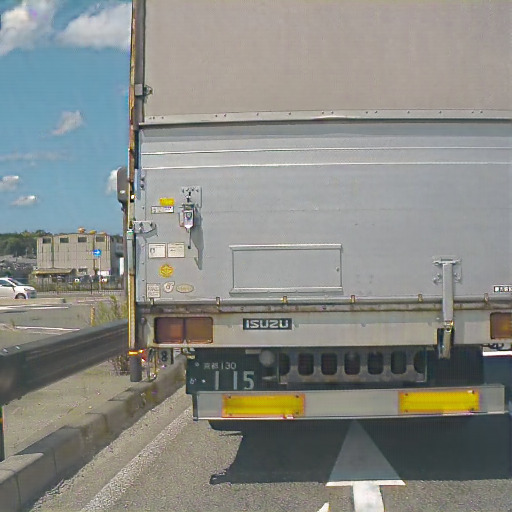} &
    \includegraphics[width=\imagesize]{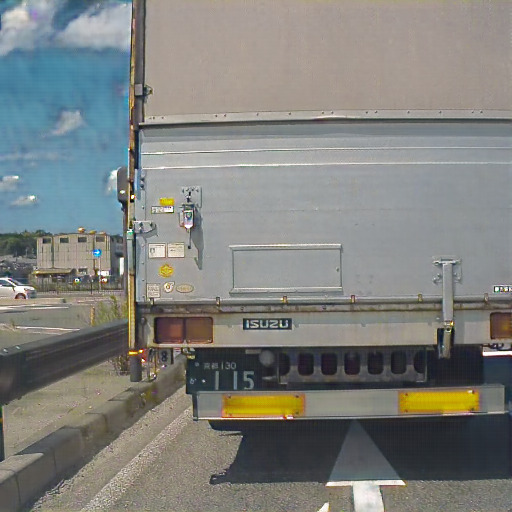} &
    \includegraphics[width=\imagesize]{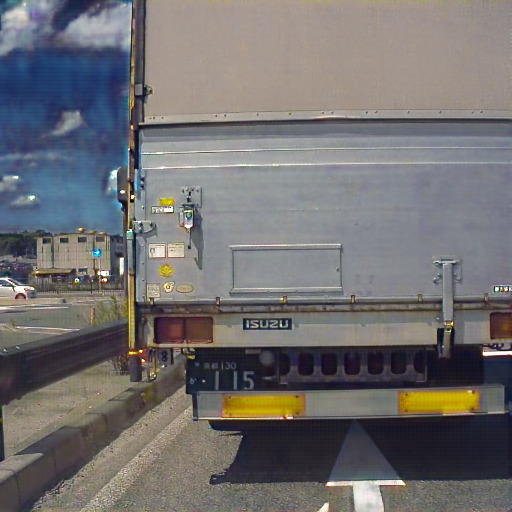}\\

    \includegraphics[width=\imagesize]{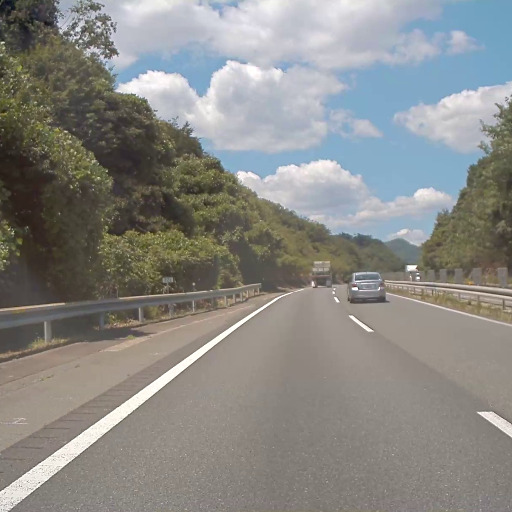} &
    \includegraphics[width=\imagesize]{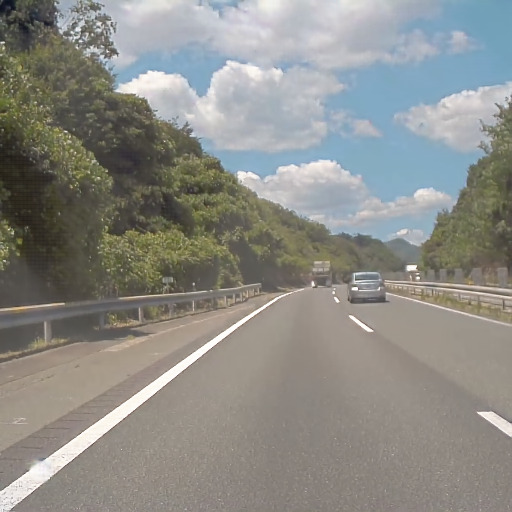} &
    \includegraphics[width=\imagesize]{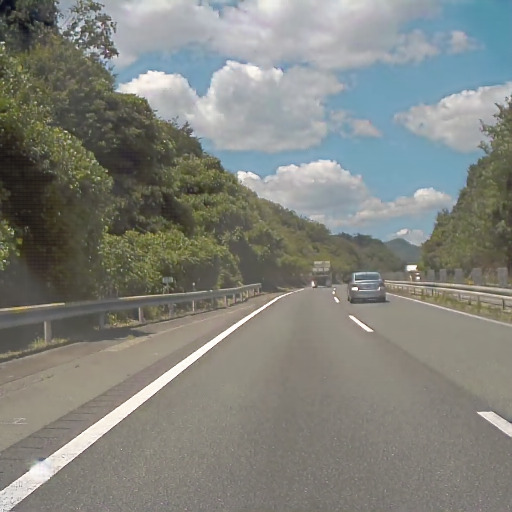} &
    \includegraphics[width=\imagesize]{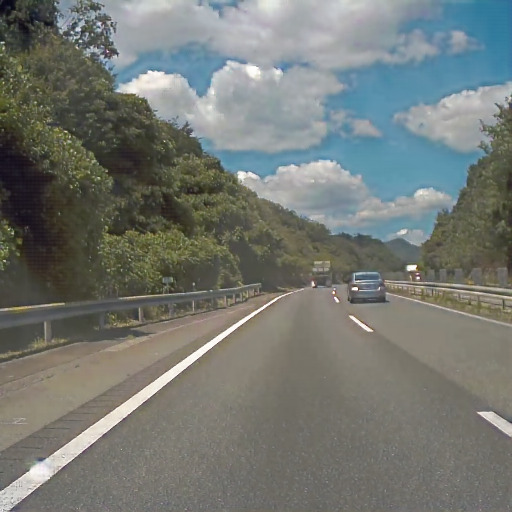} &
    \includegraphics[width=\imagesize]{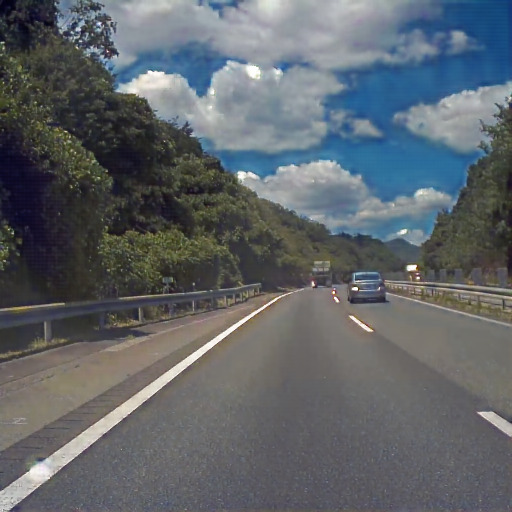} &
    \includegraphics[width=\imagesize]{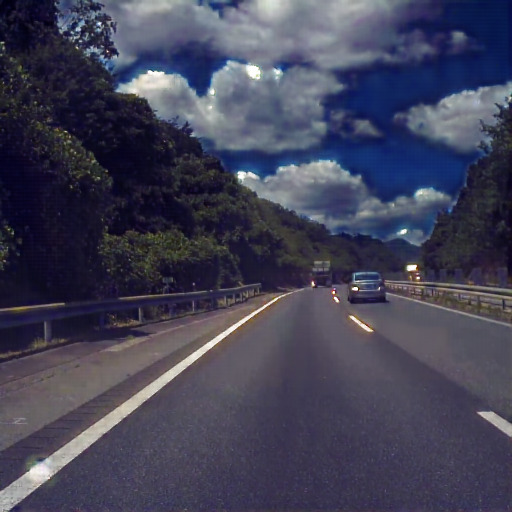}\\

    \includegraphics[width=\imagesize]{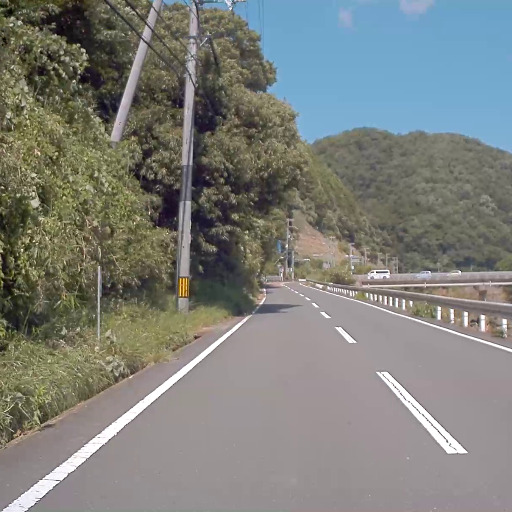} &
    \includegraphics[width=\imagesize]{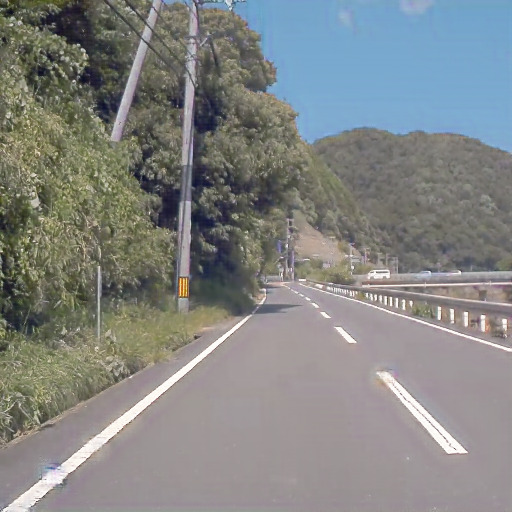} &
    \includegraphics[width=\imagesize]{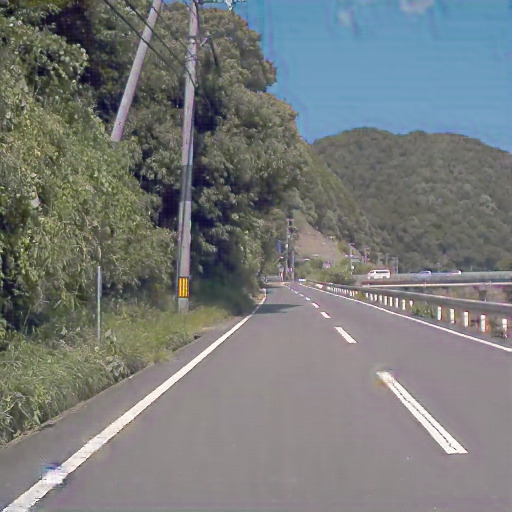} &
    \includegraphics[width=\imagesize]{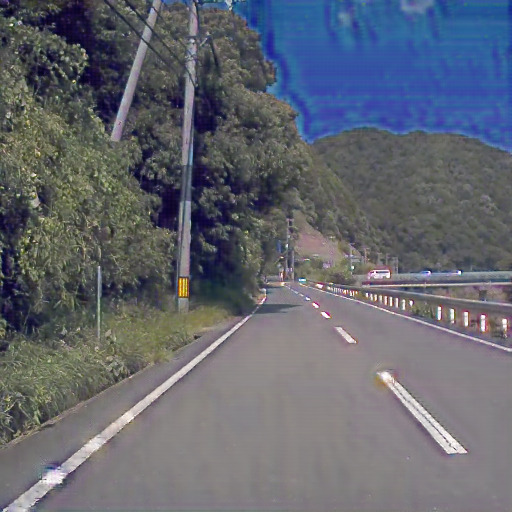} &
    \includegraphics[width=\imagesize]{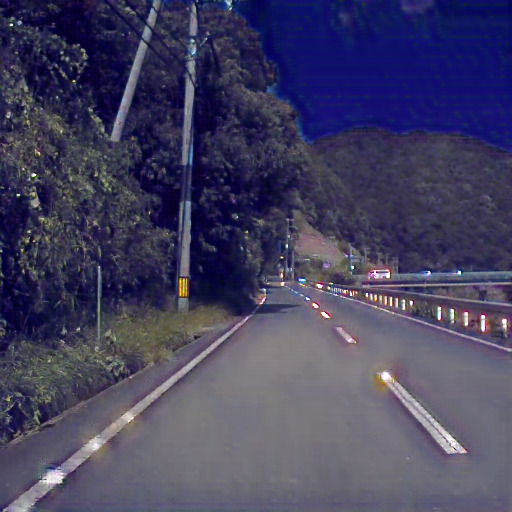} &
    \includegraphics[width=\imagesize]{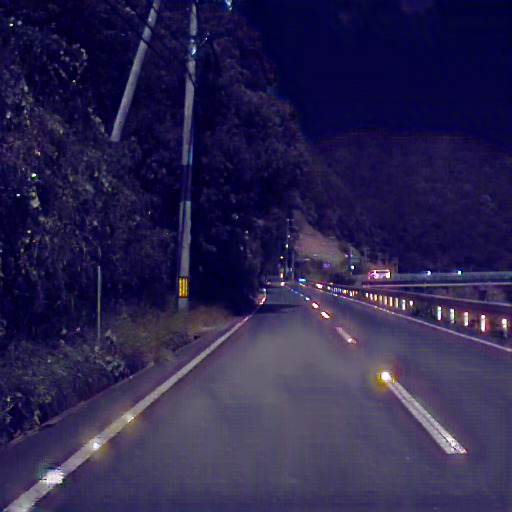}\\
    
    \includegraphics[width=\imagesize]{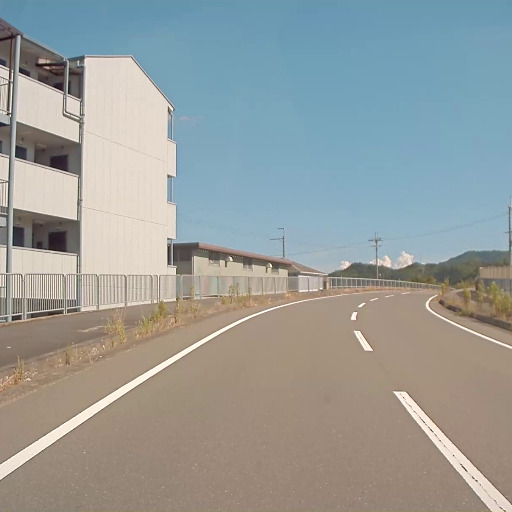} &
    \includegraphics[width=\imagesize]{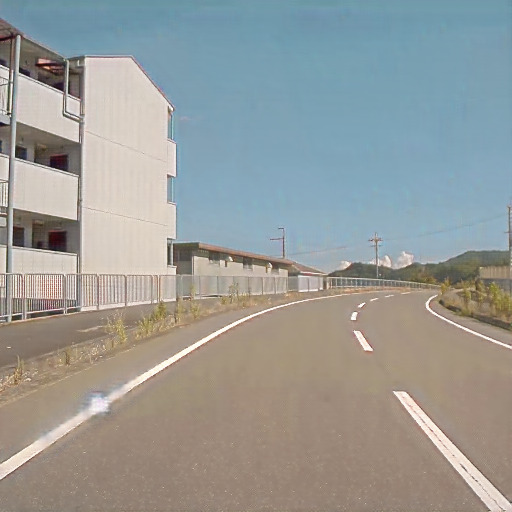} &
    \includegraphics[width=\imagesize]{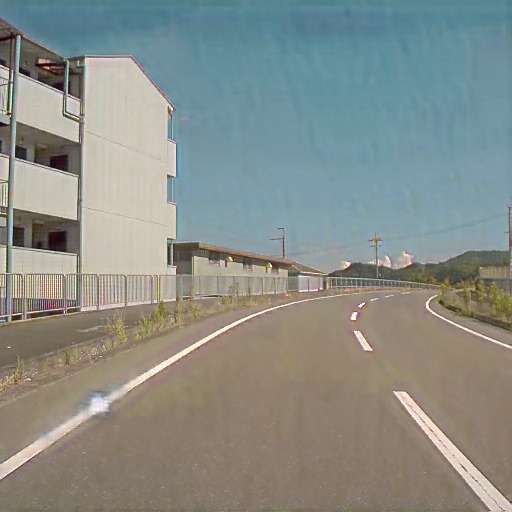} &
    \includegraphics[width=\imagesize]{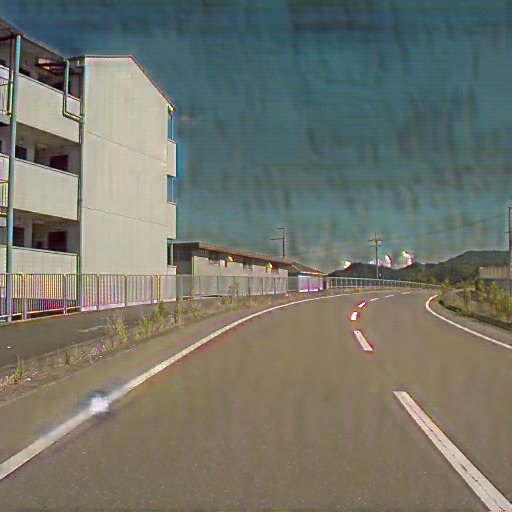} &
    \includegraphics[width=\imagesize]{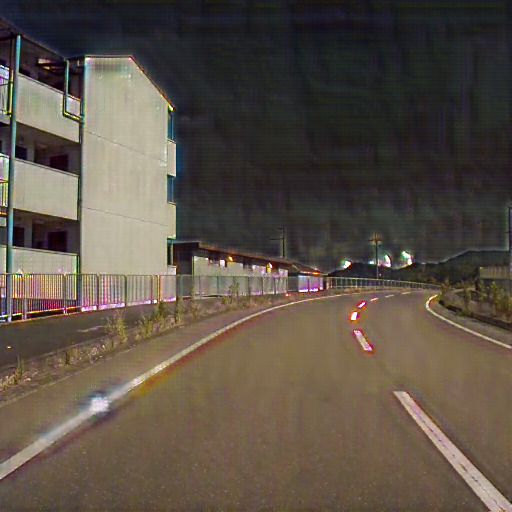} &
    \includegraphics[width=\imagesize]{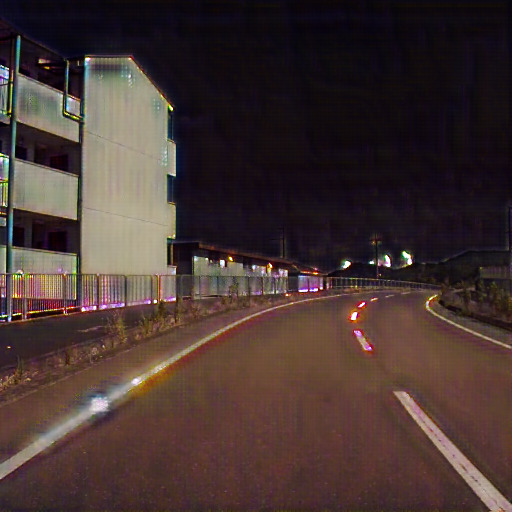}\\

    \includegraphics[width=\imagesize]{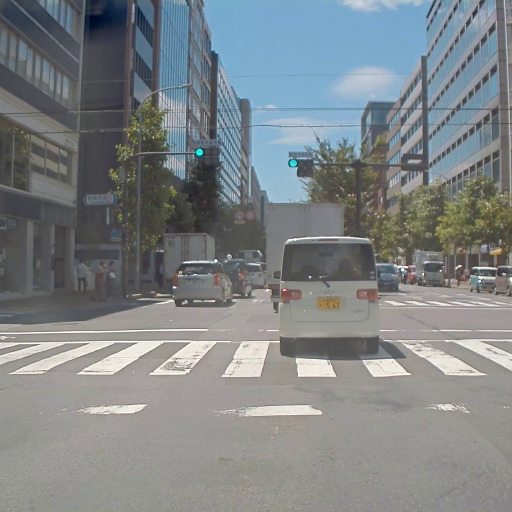} &
    \includegraphics[width=\imagesize]{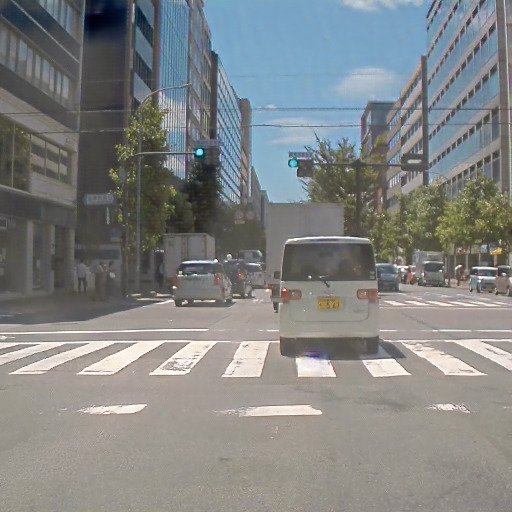} &
    \includegraphics[width=\imagesize]{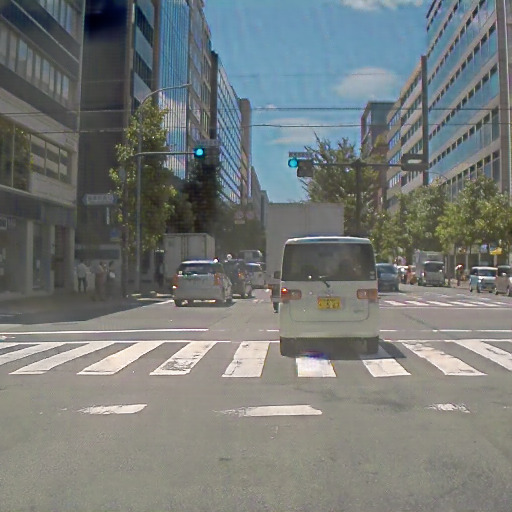} &
    \includegraphics[width=\imagesize]{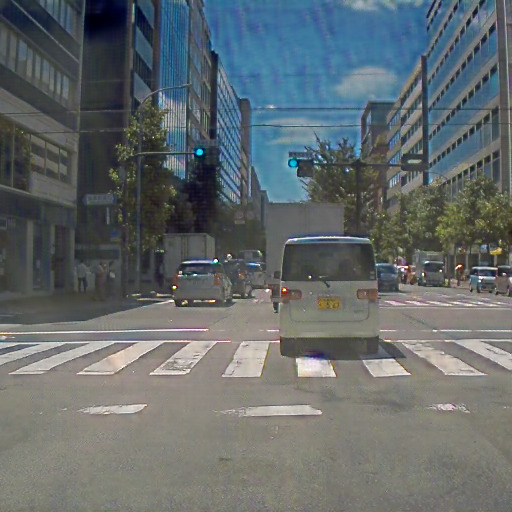} &
    \includegraphics[width=\imagesize]{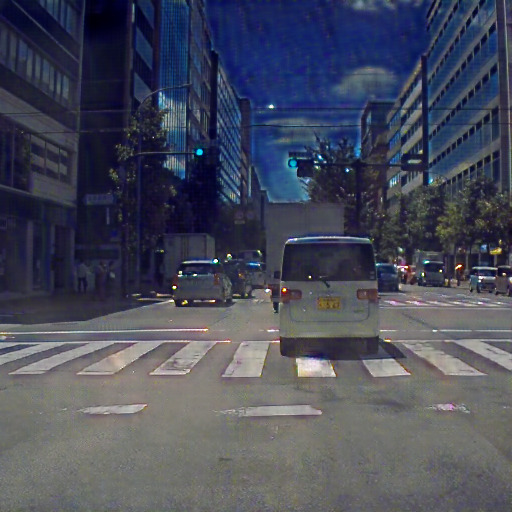} &
    \includegraphics[width=\imagesize]{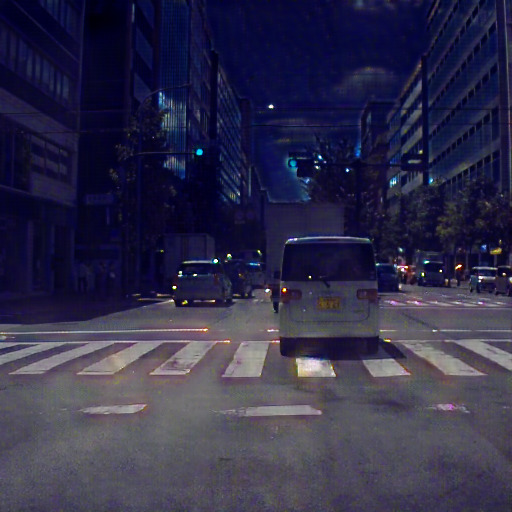}\\
    
    \includegraphics[width=\imagesize]{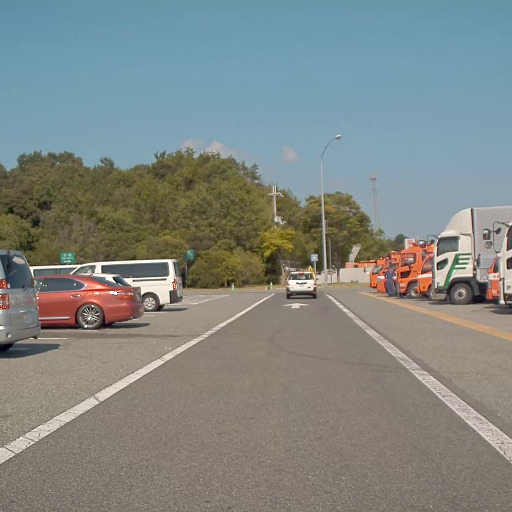} &
    \includegraphics[width=\imagesize]{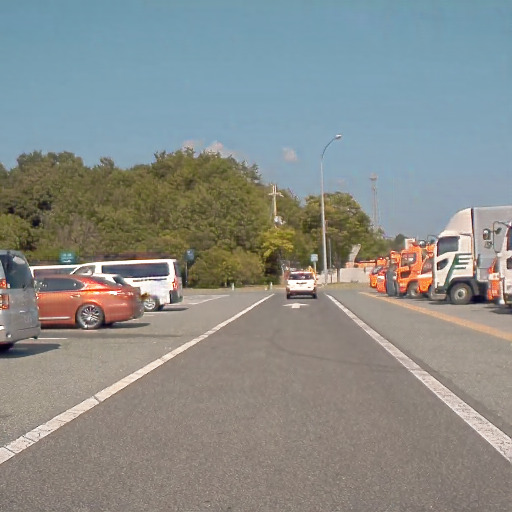} &
    \includegraphics[width=\imagesize]{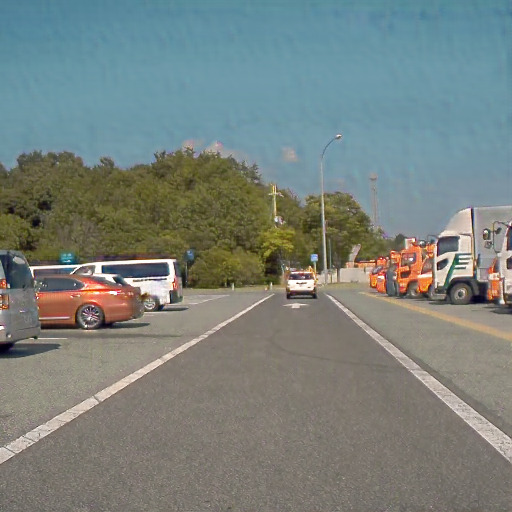} &
    \includegraphics[width=\imagesize]{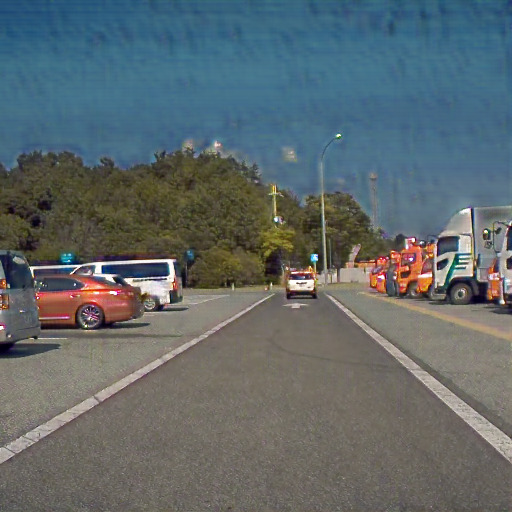} &
    \includegraphics[width=\imagesize]{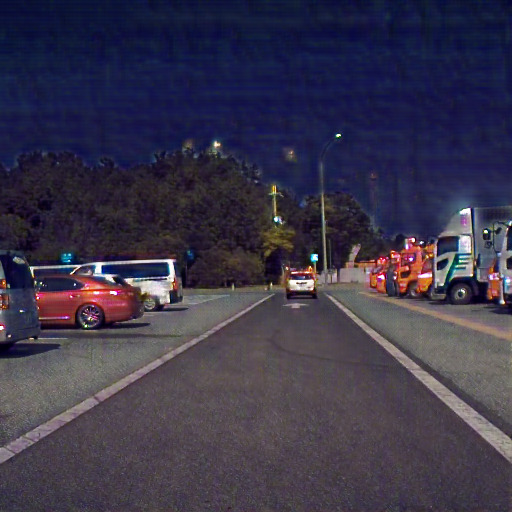} &
    \includegraphics[width=\imagesize]{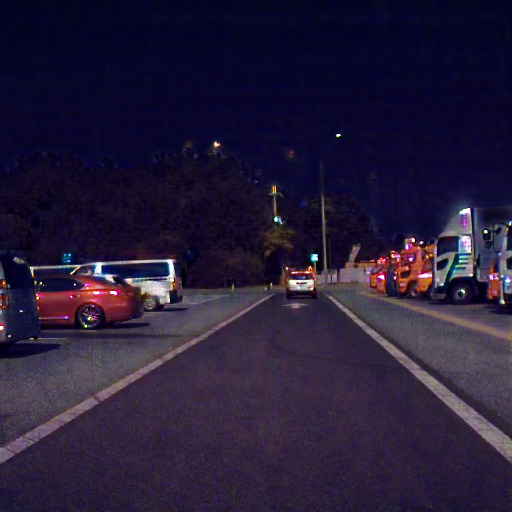}\\

    \includegraphics[width=\imagesize]{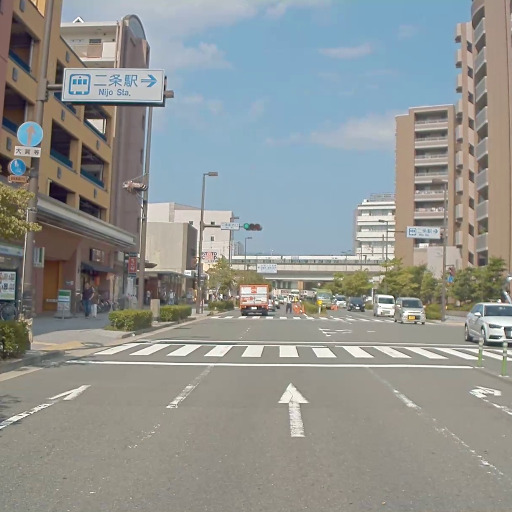} &
    \includegraphics[width=\imagesize]{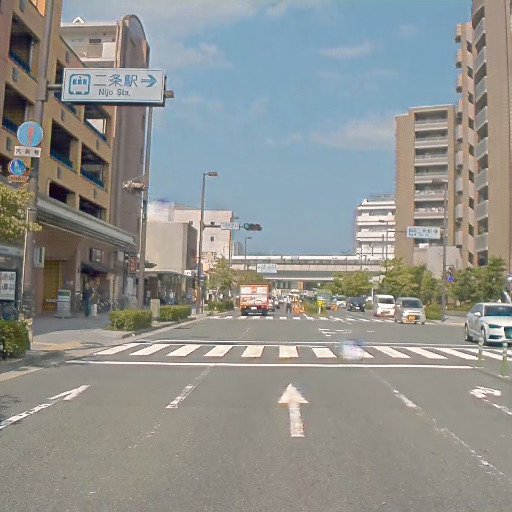} &
    \includegraphics[width=\imagesize]{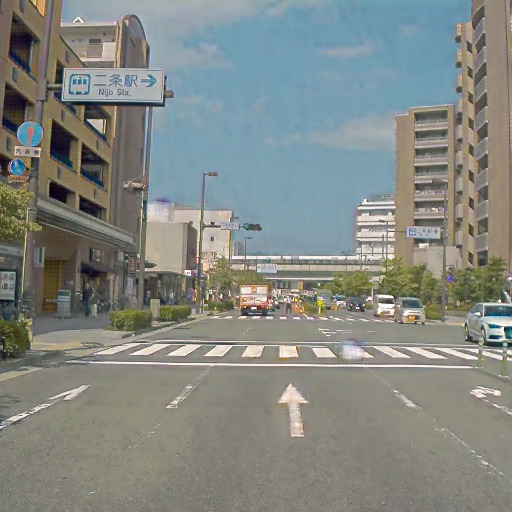} &
    \includegraphics[width=\imagesize]{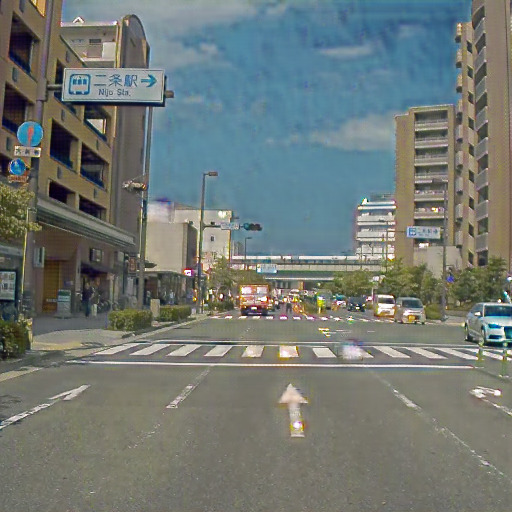} &
    \includegraphics[width=\imagesize]{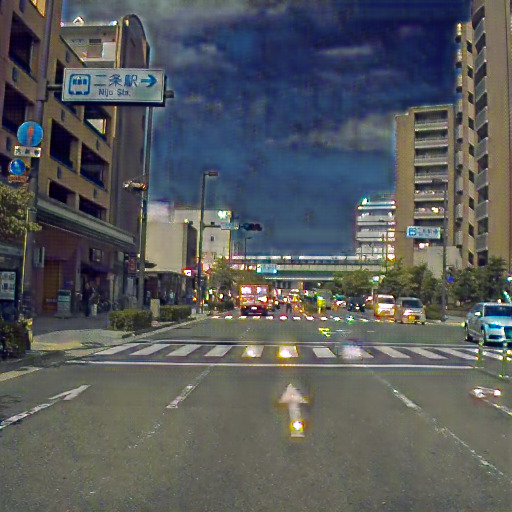} &
    \includegraphics[width=\imagesize]{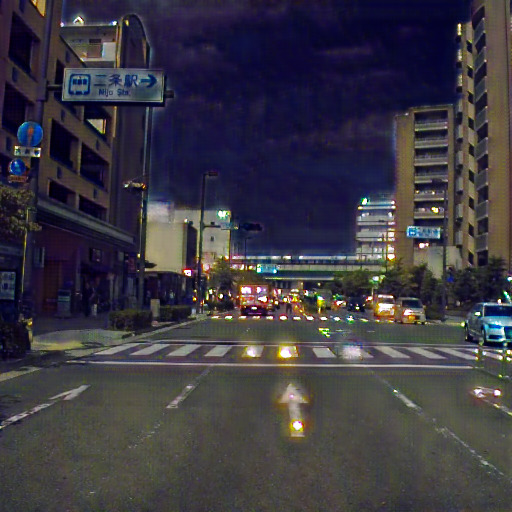}\\
    \end{tabular}
    \caption{Results on the \init{} dataset for the \daytonight{} case}
    \label{fig:supplement_init_night2day}
\end{figure}

In \autoref{fig:supplement_init_night2day} we show qualitative results using the \init{} dataset for the \daytonight{} task. The network learns to interpolate between the two domains, progressively darkening the image in the necessary areas (sky, road) and adding elements proper of the target domain (car lights, street lights).

\begin{figure}[bh!]
\setlength{\tabcolsep}{2pt}
\centering
\begin{tabular}{c|ccccccc}
    \toprule
    Source &  & 0.0 & 0.25 & 0.5 & 0.75 & 1.0 \\ 
    \midrule
    & \rotatebox{90}{\summertodawn} &
    \includegraphics[width=0.15\linewidth]{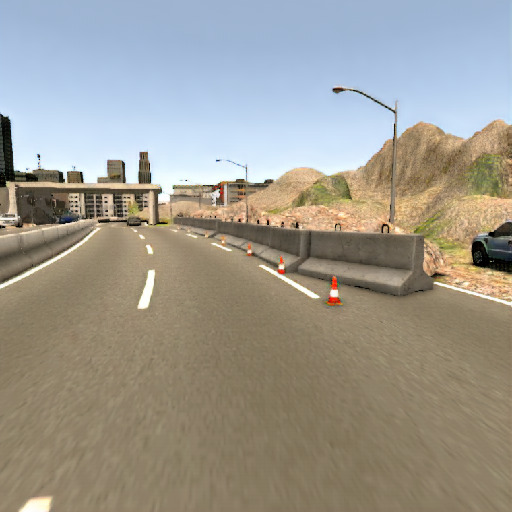} &
    \includegraphics[width=0.15\linewidth]{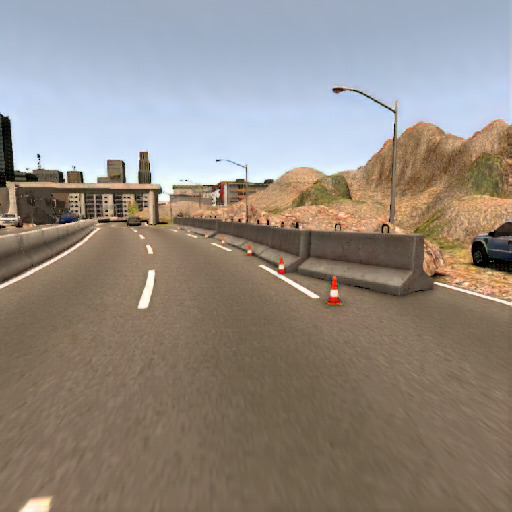} &
    \includegraphics[width=0.15\linewidth]{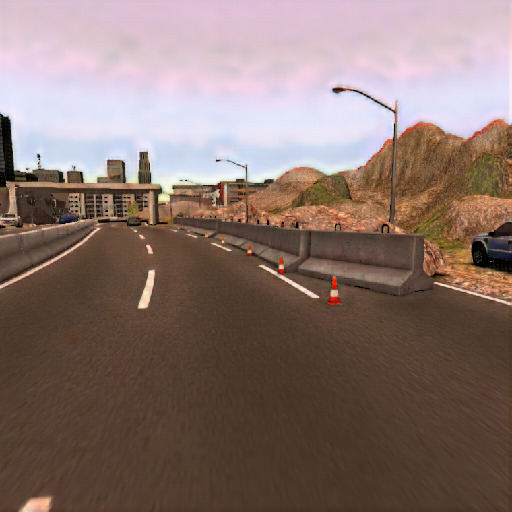} &
    \includegraphics[width=0.15\linewidth]{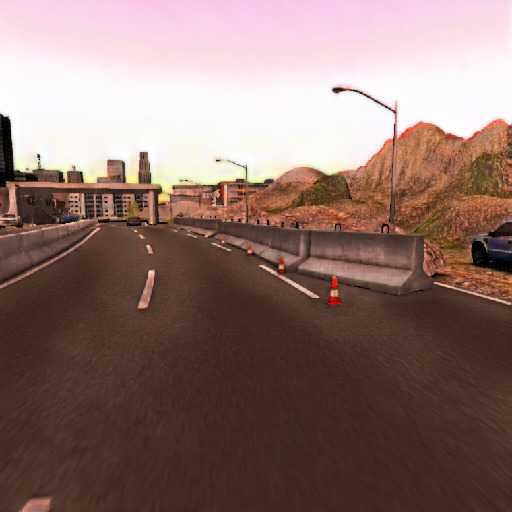} &
    \includegraphics[width=0.15\linewidth]{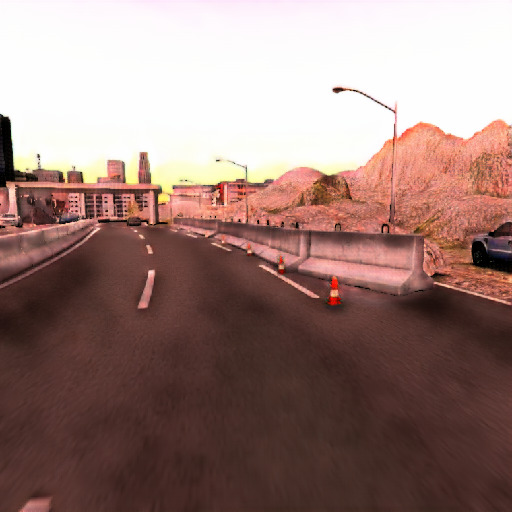}\\
    
    \includegraphics[width=0.15\linewidth]{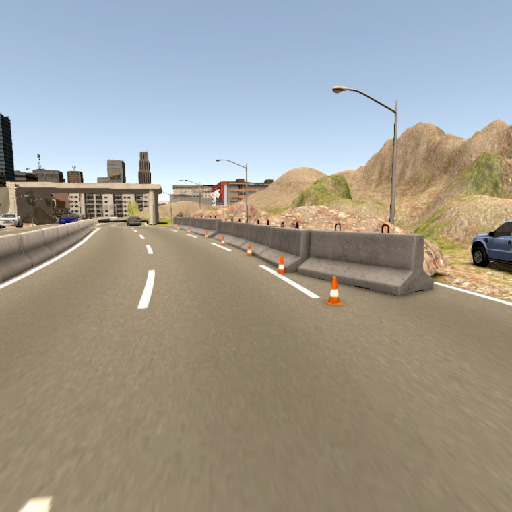} & \rotatebox{90}{\summertosunset} &
    \includegraphics[width=0.15\linewidth]{images/datasets/summer2d_s_s_n_disentangled-4/06-000867-0.00-0.00-0.00} &
    \includegraphics[width=0.15\linewidth]{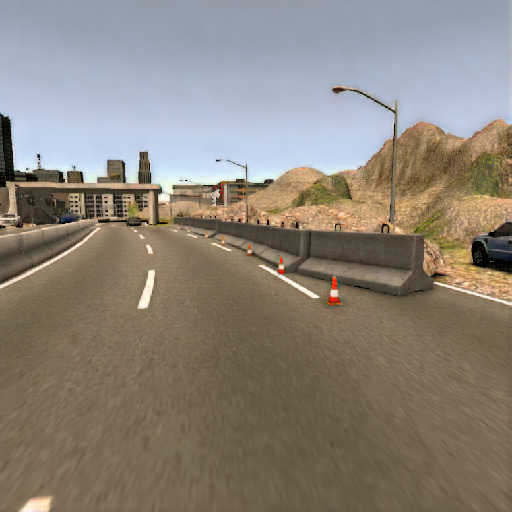} &
    \includegraphics[width=0.15\linewidth]{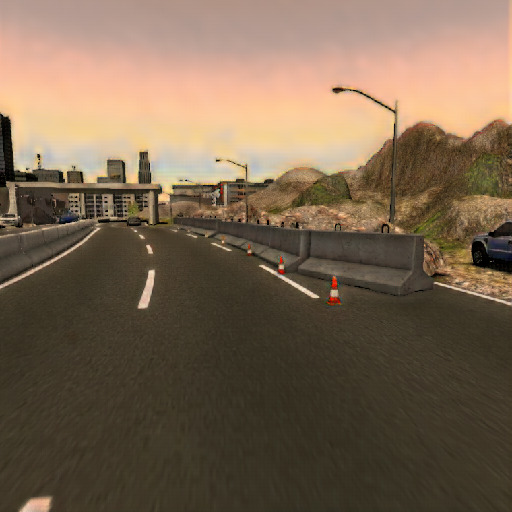} &
    \includegraphics[width=0.15\linewidth]{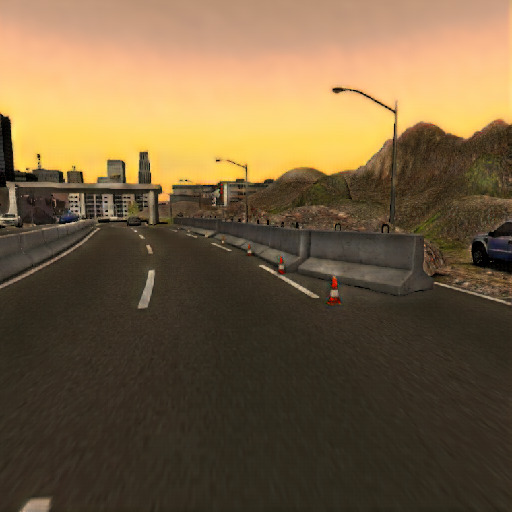} &
    \includegraphics[width=0.15\linewidth]{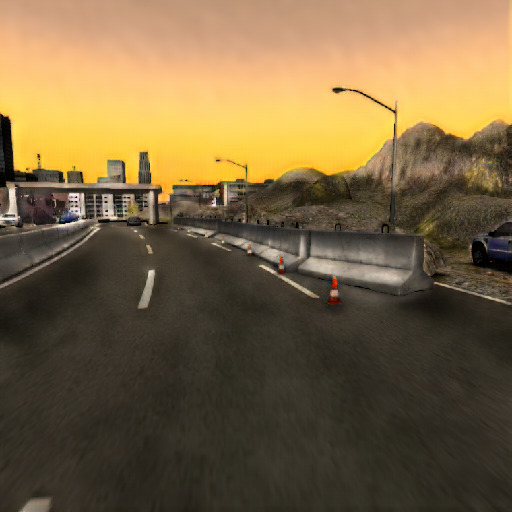}\\
    
    & \rotatebox{90}{\summertonight} &
    \includegraphics[width=0.15\linewidth]{images/datasets/summer2d_s_s_n_disentangled-4/06-000867-0.00-0.00-0.00} &
    \includegraphics[width=0.15\linewidth]{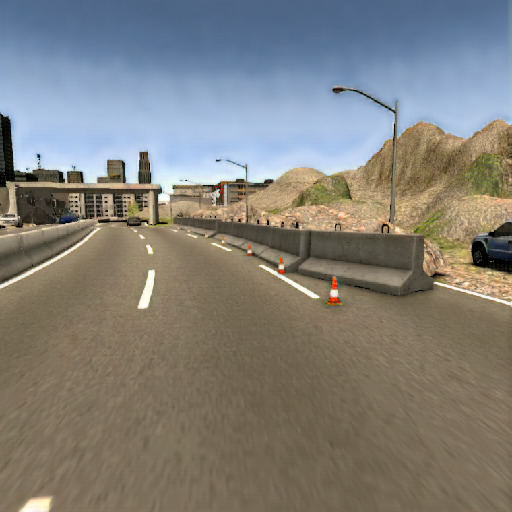} &
    \includegraphics[width=0.15\linewidth]{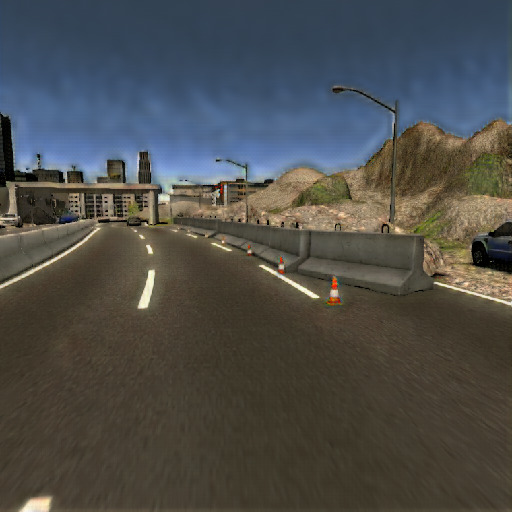} &
    \includegraphics[width=0.15\linewidth]{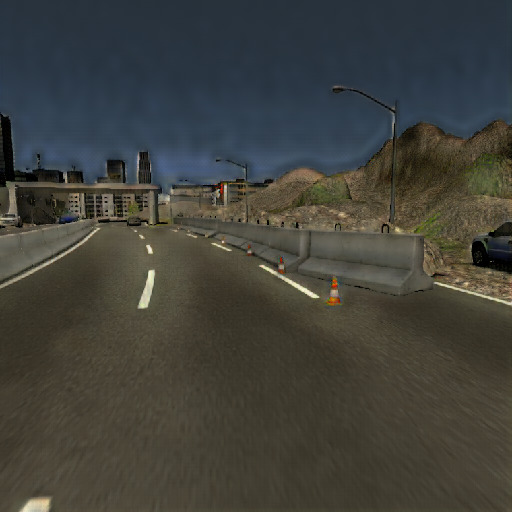} &
    \includegraphics[width=0.15\linewidth]{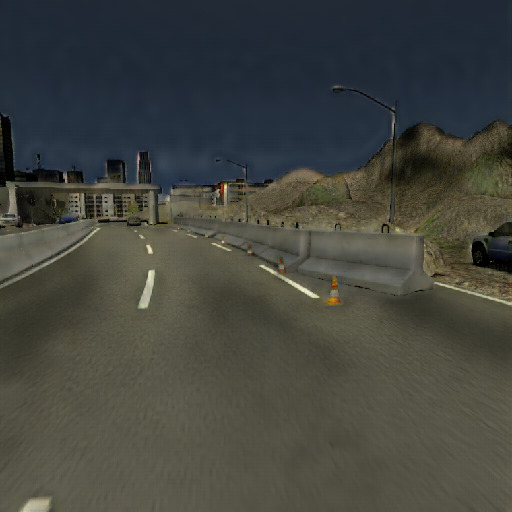}\\
    
    \midrule
    & \rotatebox{90}{\summertodawn} &
    \includegraphics[width=0.15\linewidth]{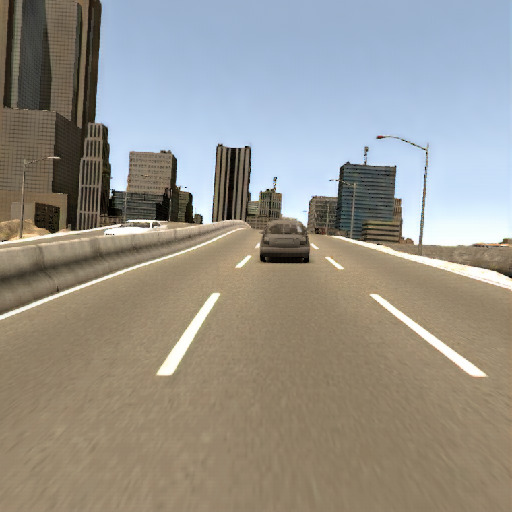} &
    \includegraphics[width=0.15\linewidth]{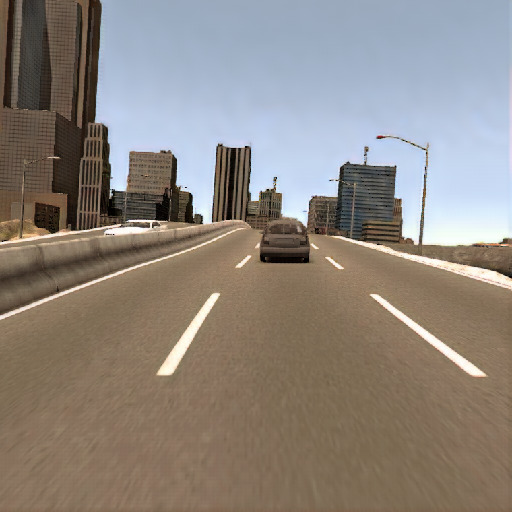} &
    \includegraphics[width=0.15\linewidth]{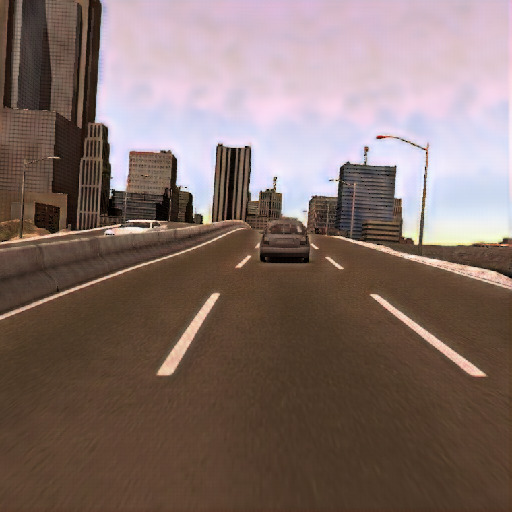} &
    \includegraphics[width=0.15\linewidth]{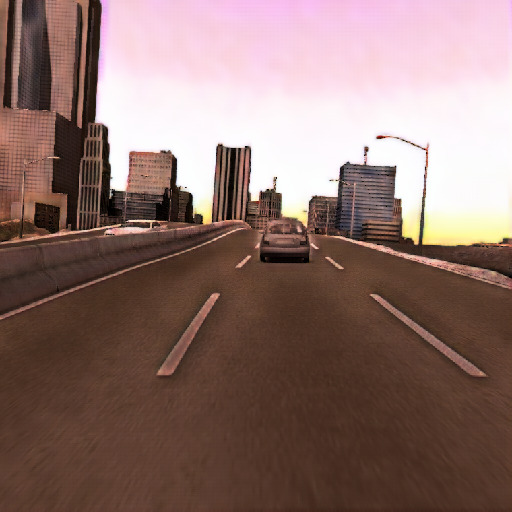} &
    \includegraphics[width=0.15\linewidth]{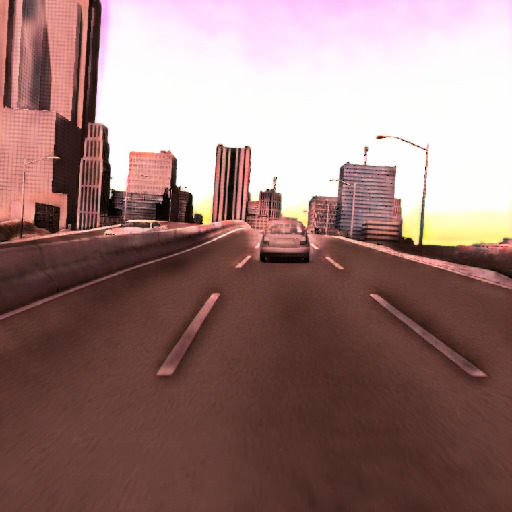}\\
    
    \includegraphics[width=0.15\linewidth]{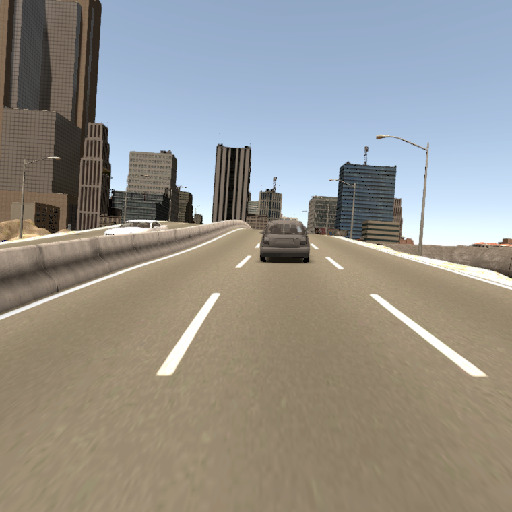} & \rotatebox{90}{\summertosunset} &
    \includegraphics[width=0.15\linewidth]{images/datasets/summer2d_s_s_n_disentangled-4/06-000582-0.00-0.00-0.00} &
    \includegraphics[width=0.15\linewidth]{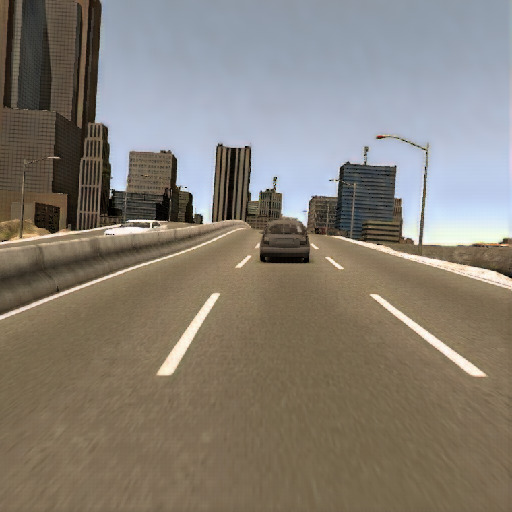} &
    \includegraphics[width=0.15\linewidth]{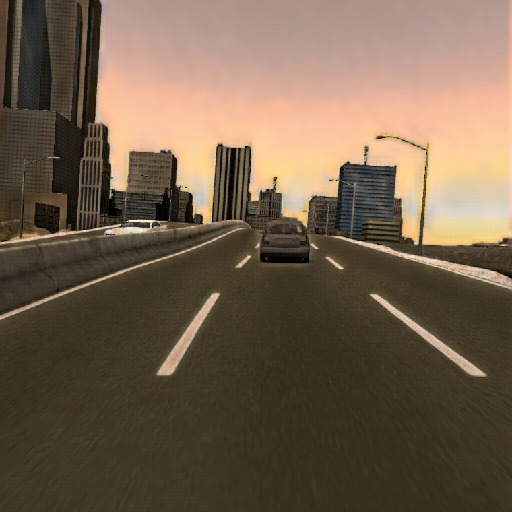} &
    \includegraphics[width=0.15\linewidth]{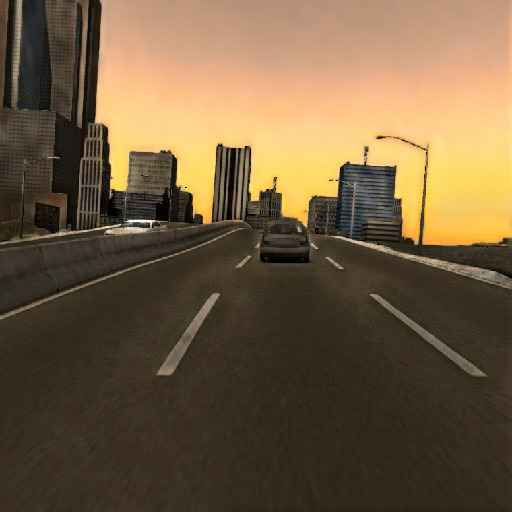} &
    \includegraphics[width=0.15\linewidth]{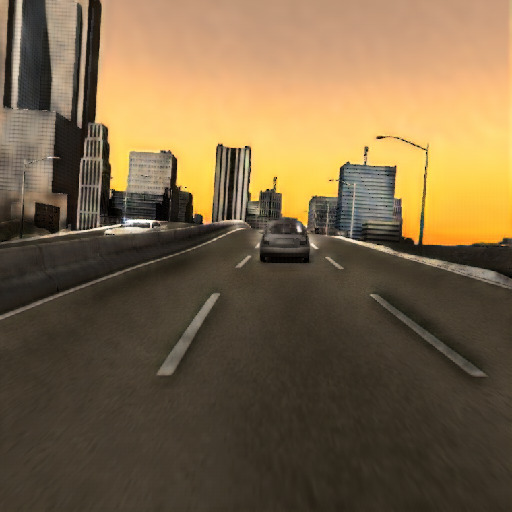}\\
    
    & \rotatebox{90}{\summertonight} &
    \includegraphics[width=0.15\linewidth]{images/datasets/summer2d_s_s_n_disentangled-4/06-000582-0.00-0.00-0.00} &
    \includegraphics[width=0.15\linewidth]{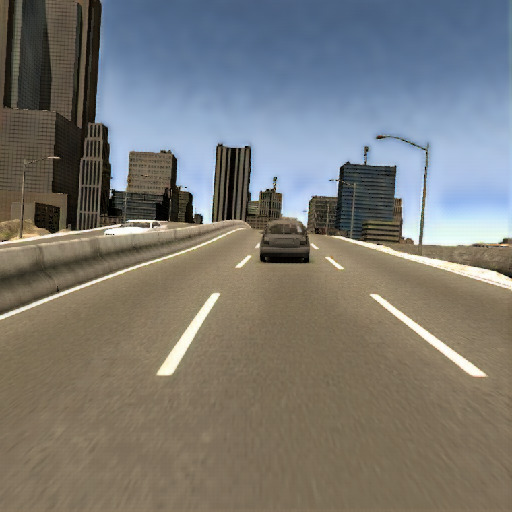} &
    \includegraphics[width=0.15\linewidth]{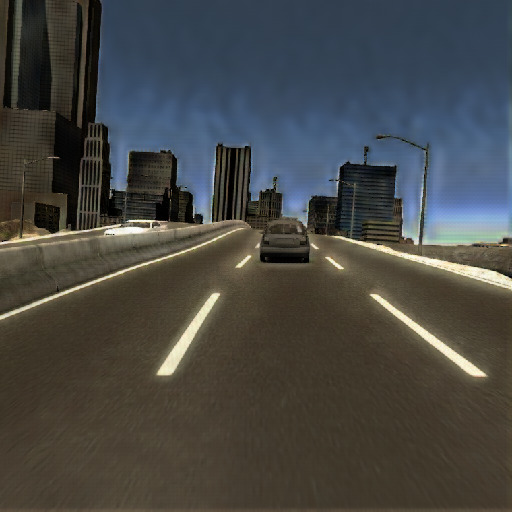} &
    \includegraphics[width=0.15\linewidth]{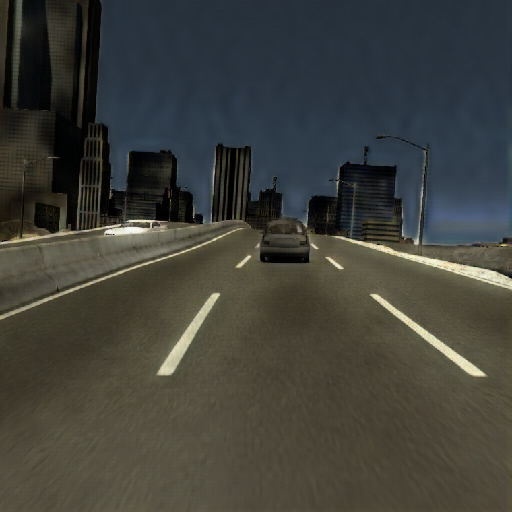} &
    \includegraphics[width=0.15\linewidth]{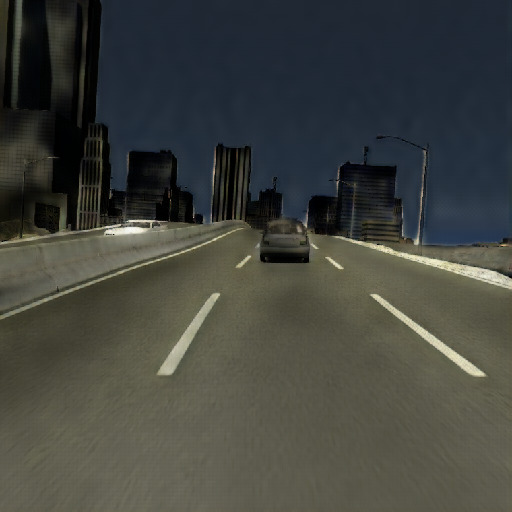}\\

\end{tabular}
\caption{Disentangled evaluation of the \summertodawnsunsetnight{} task.}
\label{fig:supplement_synthia_summertodawnsunsetnight}
\end{figure}

Similarly to the \bookCovers{} dataset, the network can learn multiple domain transformations simultaneously. 
We show this in \autoref{fig:supplement_synthia_summertodawnsunsetnight} for the \summertodawnsunsetnight{} task. 
A \emph{single} model is trained to learn three different domain mappings. 
No conflicts between the domains (such as artifacts) are noticeable.

We measure the quality of these transformations numerically using the \FID{} \cite{heusel2017gans} and \LPIPS{} \cite{zhang2018unreasonable} metrics. 
We evaluate the model with different values of $p$ on a random set of images from the source domain and measure the distance to different random subsets of both source and target images. 
Ideally, images transformed with $p=0$ have the lowest metric value when compared against the source domain, as the network simply performs an identity map. 

The opposite happens when comparing with the target images. 
The most similar image set is that with $p=1$. 
We report our findings on \autoref{tab:fid_lpips_multiplesynthia} for individual transformations and \autoref{tab:fid_lpips_summer2winter_night} for entangled ones.
The results show how the network is able to progressively make the image more dissimilar from the source domain and more similar to the target one.

\begin{table}[h!]
\center
\begin{tabular}{c@{\hskip 5mm}c}

\begin{tabular}{c|cc|cc}
\toprule
&\multicolumn{2}{c}{\textbf{Vs. source}} & \multicolumn{2}{c}{\textbf{Vs. target}}\\
\hline
\textbf{p} & \FID{} & \LPIPS{} & \FID{} & \LPIPS{}\\
\hline
0    & \cellcolor{metric_step0}77.560 & \cellcolor{metric_step0}0.583 & \cellcolor{metric_step4}258.961 & \cellcolor{metric_step4}0.807 \\
0.25 & \cellcolor{metric_step1}84.117 & \cellcolor{metric_step1}0.585   & \cellcolor{metric_step3}253.026 & \cellcolor{metric_step3}0.789  \\
0.5  & \cellcolor{metric_step2}99.404 & \cellcolor{metric_step2}0.614  & \cellcolor{metric_step2}240.897 & \cellcolor{metric_step2}0.749  \\
0.75 & \cellcolor{metric_step3}123.534 & \cellcolor{metric_step3}0.665   & \cellcolor{metric_step1}228.889 & \cellcolor{metric_step1}0.720  \\
1    & \cellcolor{metric_step4}185.267 & \cellcolor{metric_step4}0.703 & \cellcolor{metric_step0}210.790 & \cellcolor{metric_step0}0.701 \\
\bottomrule
\end{tabular} &
\begin{tabular}{c|cc|cc}
\toprule
&\multicolumn{2}{c}{\textbf{Vs. source}} & \multicolumn{2}{c}{\textbf{Vs. target}}\\
\hline
\textbf{p} & \FID{} & \LPIPS{} & \FID{} & \LPIPS{}\\
\hline
0    & \cellcolor{metric_step0}76.225 & \cellcolor{metric_step1}0.584  & \cellcolor{metric_step4}179.929 & \cellcolor{metric_step4}0.641 \\
0.25 & \cellcolor{metric_step1}85.276 & \cellcolor{metric_step0}0.583  & \cellcolor{metric_step3}178.206 & \cellcolor{metric_step3}0.614  \\
0.5  & \cellcolor{metric_step2}108.043 & \cellcolor{metric_step2}0.596  & \cellcolor{metric_step2}169.597 & \cellcolor{metric_step2}0.594 \\
0.75 & \cellcolor{metric_step3}125.122 & \cellcolor{metric_step3}0.611 & \cellcolor{metric_step1}160.732 & \cellcolor{metric_step1}0.584 \\
1    & \cellcolor{metric_step4}164.958 & \cellcolor{metric_step4}0.637  & \cellcolor{metric_step0}151.933 & \cellcolor{metric_step0}0.566 \\
\bottomrule
\end{tabular}\\
a) \summertosoftrain{} & b) \summertofog{}\\

\begin{tabular}{c|cc|cc}
\toprule
&\multicolumn{2}{c}{\textbf{Vs. source}} & \multicolumn{2}{c}{\textbf{Vs. target}}\\
\hline
\textbf{p} & \FID{} & \LPIPS{} & \FID{} & \LPIPS{}\\
\hline
0    & \cellcolor{metric_step0}77.359 & \cellcolor{metric_step0}0.585 & \cellcolor{metric_step4}162.797 & \cellcolor{metric_step4}0.633 \\
0.25 & \cellcolor{metric_step1}80.647 & \cellcolor{metric_step1}0.586  & \cellcolor{metric_step3}162.495 & \cellcolor{metric_step3}0.626  \\
0.5  & \cellcolor{metric_step2}87.051 & \cellcolor{metric_step2}0.589 & \cellcolor{metric_step2}161.036 & \cellcolor{metric_step2}0.621 \\
0.75 & \cellcolor{metric_step3}95.664  & \cellcolor{metric_step3}0.597 & \cellcolor{metric_step1}155.675 & \cellcolor{metric_step1}0.611  \\
1    & \cellcolor{metric_step4}113.402 & \cellcolor{metric_step4}0.608  & \cellcolor{metric_step0}155.786 & \cellcolor{metric_step0}0.608  \\
\bottomrule
\end{tabular} &
\begin{tabular}{c|cc|cc}
\toprule
&\multicolumn{2}{c}{\textbf{Vs. source}} & \multicolumn{2}{c}{\textbf{Vs. target}}\\
\hline
\textbf{p} & \FID{} & \LPIPS{} & \FID{} & \LPIPS{}\\
\hline
0    & \cellcolor{metric_step0}76.282  & \cellcolor{metric_step0}0.584  & \cellcolor{metric_step4}149.165 & \cellcolor{metric_step4}0.635    \\
0.25 & \cellcolor{metric_step1}77.612 & \cellcolor{metric_step1}0.584 & \cellcolor{metric_step3}148.333 & \cellcolor{metric_step3}0.628  \\
0.5  & \cellcolor{metric_step2}81.391 & \cellcolor{metric_step2}0.587  & \cellcolor{metric_step2}150.723 & \cellcolor{metric_step2}0.622  \\
0.75 & \cellcolor{metric_step3}85.515 & \cellcolor{metric_step3}0.590  & \cellcolor{metric_step1}148.327 & \cellcolor{metric_step1}0.617  \\
1    & \cellcolor{metric_step4}95.609 & \cellcolor{metric_step4}0.604  & \cellcolor{metric_step0}144.940 & \cellcolor{metric_step0}0.615 \\
\bottomrule
\end{tabular}\\
c) \summertofall{} & d) \summertospring{}\\

\begin{tabular}{c|cc|cc}
\toprule
&\multicolumn{2}{c}{\textbf{Vs. source}} & \multicolumn{2}{c}{\textbf{Vs. target}}\\
\hline
\textbf{p} & \FID{} & \LPIPS{} & \FID{} & \LPIPS{}\\
\hline
0    & \cellcolor{metric_step0}74.068 & \cellcolor{metric_step0}0.584 & \cellcolor{metric_step4}163.253  & \cellcolor{metric_step4}0.697 \\
0.25 & \cellcolor{metric_step1}75.020 & \cellcolor{metric_step1}0.589  & \cellcolor{metric_step3}162.684 & \cellcolor{metric_step3}0.685 \\
0.5  & \cellcolor{metric_step2}81.797 & \cellcolor{metric_step2}0.642   & \cellcolor{metric_step2}161.785 & \cellcolor{metric_step2}0.656  \\
0.75 & \cellcolor{metric_step3}92.5742 & \cellcolor{metric_step3}0.657  & \cellcolor{metric_step1}157.313 & \cellcolor{metric_step1}0.645 \\
1    & \cellcolor{metric_step4}121.722 & \cellcolor{metric_step4}0.664  & \cellcolor{metric_step0}155.643 & \cellcolor{metric_step0}0.631 \\
\bottomrule
\end{tabular} &
\begin{tabular}{c|cc|cc}
\toprule
&\multicolumn{2}{c}{\textbf{Vs. source}} & \multicolumn{2}{c}{\textbf{Vs. target}}\\
\hline
\textbf{p} & \FID{} & \LPIPS{} & \FID{} & \LPIPS{}\\
\hline
0    & \cellcolor{metric_step0}74.363 & \cellcolor{metric_step0}0.583 & \cellcolor{metric_step4}159.036 & \cellcolor{metric_step4}0.663  \\
0.25 & \cellcolor{metric_step1}76.965 & \cellcolor{metric_step1}0.583  & \cellcolor{metric_step3}157.723 & \cellcolor{metric_step3}0.657 \\
0.5  & \cellcolor{metric_step2}83.311  & \cellcolor{metric_step2}0.592  & \cellcolor{metric_step2}149.511 & \cellcolor{metric_step2}0.645  \\
0.75 & \cellcolor{metric_step3}103.049 & \cellcolor{metric_step3}0.614  & \cellcolor{metric_step1}142.353 & \cellcolor{metric_step1}0.627  \\
1    & \cellcolor{metric_step4}134.259 & \cellcolor{metric_step4}0.649  & \cellcolor{metric_step0}146.279 & \cellcolor{metric_step0}0.612  \\
\bottomrule
\end{tabular}\\
e) \summertodawn{} & f) \summertosunset{}\\
\end{tabular}
\caption{Average \FID{} and \LPIPS{} scores for multiple tasks using subsets of \synthia{} comparing the result image with randomly picked source and target images. A darker background color denotes a lower distance between the two sets of images.}
\label{tab:fid_lpips_multiplesynthia}
\end{table}

\begin{table}[h!]
    \centering
    \begin{tabular}{l|l|l|l|l}
        & \multicolumn{1}{c|}{\scriptsize \vtop{\hbox{\strut 0\% Winter}\hbox{\strut 0\% Night}}} 
        & \multicolumn{1}{c|}{\scriptsize \vtop{\hbox{\strut 100\% Winter}\hbox{\strut 0\% Night}}} 
        & \multicolumn{1}{c|}{\scriptsize \vtop{\hbox{\strut 0\% Winter}\hbox{\strut 100\% Night}}}
        & \multicolumn{1}{c}{\scriptsize \vtop{\hbox{\strut 50\% Winter}\hbox{\strut 0\% Night}}} \\\hline
        &
        \multicolumn{1}{c|}{\includegraphics[width=0.18\textwidth]{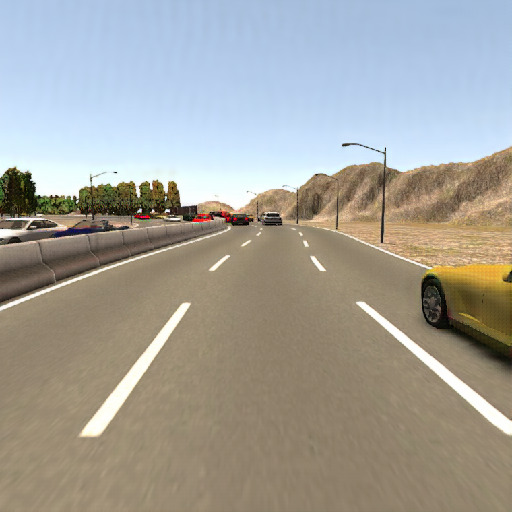}} &
        \multicolumn{1}{c|}{\includegraphics[width=0.18\textwidth]{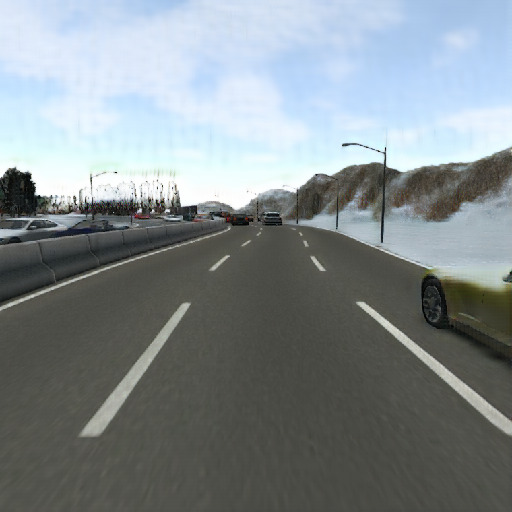}} &
        \multicolumn{1}{c|}{\includegraphics[width=0.18\textwidth]{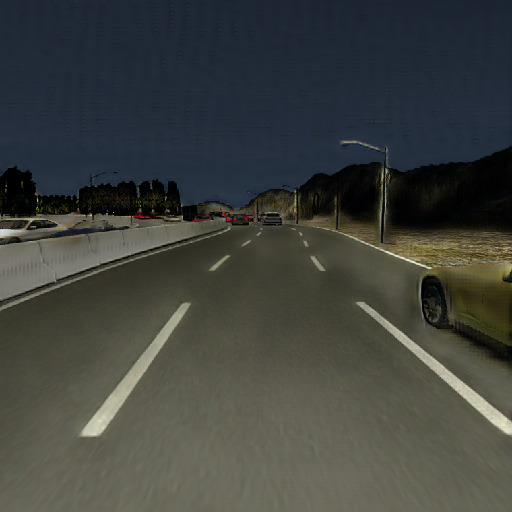}} &
        \multicolumn{1}{c}{\includegraphics[width=0.18\textwidth]{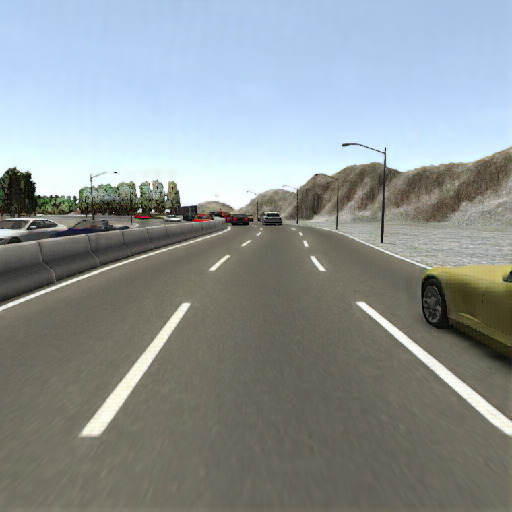}}
        \\\hline
        {\scriptsize FID w.r.t. to Summer} & {\scriptsize \textbf{20.423}} & {\scriptsize 90.128} & {\scriptsize 114.602} &{\scriptsize 58.436}\\
        {\scriptsize FID w.r.t. to Winter} & {\scriptsize 127.738} & {\scriptsize \textbf{116.896}} &  {\scriptsize 135.517} & {\scriptsize 119.847}\\
        {\scriptsize FID w.r.t. to Night} & {\scriptsize 173.572} & {\scriptsize 169.287} &  {\scriptsize \textbf{161.076}} & {\scriptsize 171.635}\\\hline
        {\scriptsize LPIPS w.r.t. to Summer} & {\scriptsize \textbf{0.022}} & {\scriptsize 0.155} &  {\scriptsize 0.242} &  {\scriptsize 0.0764}\\
        {\scriptsize LPIPS w.r.t. to Winter} & {\scriptsize 0.626} & {\scriptsize \textbf{0.599}} & {\scriptsize 0.621} & {\scriptsize 0.606}\\
        {\scriptsize LPIPS w.r.t. to Night} & {\scriptsize 0.623} & {\scriptsize 0.613} & {\scriptsize \textbf{0.567}} & {\scriptsize 0.614}\\\midrule
        
        &
        \multicolumn{1}{c|}{\scriptsize \vtop{\hbox{\strut 0\% Winter}\hbox{\strut 50\% Night}}} 
        & \multicolumn{1}{c|}{\scriptsize \vtop{\hbox{\strut 100\% Winter}\hbox{\strut 100\% Night}}}
        & \multicolumn{1}{c|}{\scriptsize \vtop{\hbox{\strut 75\% Winter}\hbox{\strut 0\% Night}}} 
        & \multicolumn{1}{c}{\scriptsize \vtop{\hbox{\strut 0\% Winter}\hbox{\strut 75\% Night}}}\\\hline
        &
        \multicolumn{1}{c|}{\includegraphics[width=0.18\textwidth]{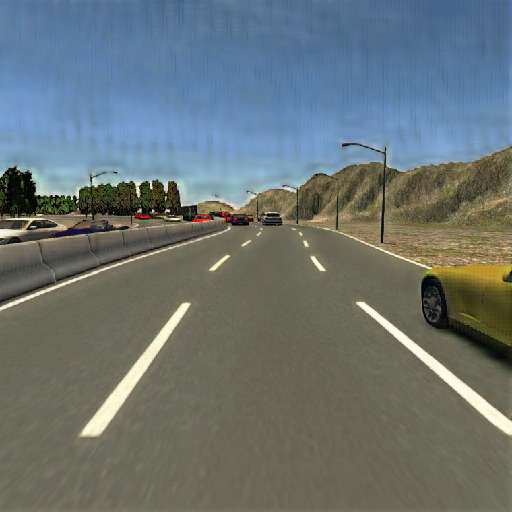}} &
        \multicolumn{1}{c|}{\includegraphics[width=0.18\textwidth]{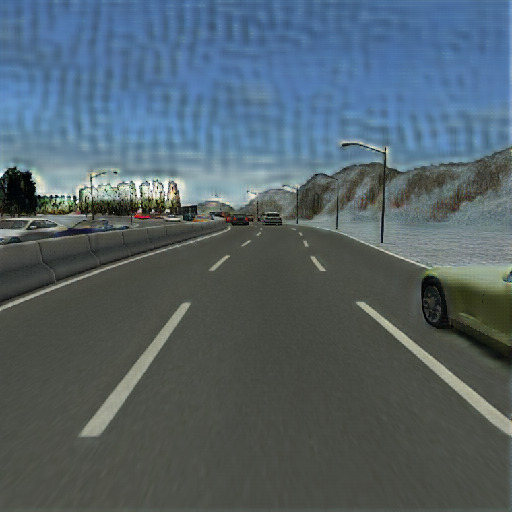}} &
        \multicolumn{1}{c|}{\includegraphics[width=0.18\textwidth]{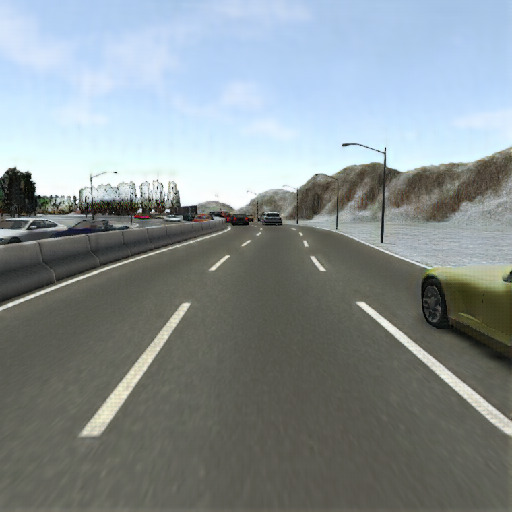}} &
        \multicolumn{1}{c}{\includegraphics[width=0.18\textwidth]{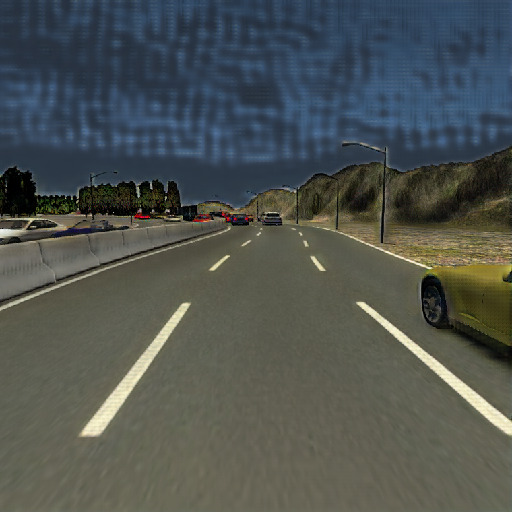}}\\\hline
        
        {\scriptsize FID w.r.t. to Summer} & {\scriptsize \textbf{52.557}} & {\scriptsize 136.331} & {\scriptsize 82.661} & {\scriptsize 90.585}\\
        {\scriptsize FID w.r.t. to Winter} & {\scriptsize 155.389} & {\scriptsize 146.648} &  {\scriptsize \textbf{120.264}} & {\scriptsize 133.728}\\
        {\scriptsize FID w.r.t. to Night} & {\scriptsize 163.074} & {\scriptsize 186.578} &  {\scriptsize 171.976} & {\scriptsize \textbf{160.286}}\\\hline
        {\scriptsize LPIPS w.r.t. to Summer} & {\scriptsize \textbf{0.090}} & {\scriptsize 0.199} &  {\scriptsize 0.111} &  {\scriptsize 0.178}\\
        {\scriptsize LPIPS w.r.t. to Winter} & {\scriptsize 0.620} & {\scriptsize 0.612} & {\scriptsize \textbf{0.602}} & {\scriptsize 0.621}\\
        {\scriptsize LPIPS w.r.t. to Night} & {\scriptsize 0.606} & {\scriptsize 0.600} & {\scriptsize 0.613} & {\scriptsize \textbf{0.582}}\\\hline

    \end{tabular}
    \caption{Average FID and L-PIPS distance for \summertowinternight{} transformations on arbitrary configurations. The images at the top are representatives of the transformation. In \textbf{bold} the set with the smallest distance row-wise is highlighted.}
    \label{tab:fid_lpips_summer2winter_night}
\end{table}

\clearpage
\subsection{Transformation quality} \label{supp:transformation_quality}
We are interested in determining whether the addition of the conditioning vector $p$ has a detrimental effect on the quality of the reconstruction. 
To this end, we measure the \FID{} and \LPIPS{} distances on images produced by CycleGAN \cite{CycleGAN2017} and our method parametrized with $p=1$, with respect to the target domain.
We train both models using the same data and hyperparameters.
In \autoref{tab:fid_lpips_vs_cyclegan} we show that the quality of our transformations is similar to that of existing formulations.

\begin{table}[h!]
\center
\begin{tabular}{c|cc|cc}
\toprule
&\multicolumn{2}{c|}{\textbf{CycleGAN \cite{CycleGAN2017}}} & \multicolumn{2}{c}{\textbf{Ours}}\\
\hline
\textit{Task} & \FID{} & \LPIPS{} $\pm \sigma$ & \FID{} & \LPIPS{} $\pm \sigma$\\
\hline
\synthia{} \summertowinter{} & \textbf{118.621} & 0.597 $\pm$ 0.091 & 132.208 & \textbf{0.590 $\pm$ 0.097}\\
\synthia{} \summertosoftrain{} & \textbf{201.941} & \textbf{0.693 $\pm$ 0.060} & 210.790 & 0.701 $\pm$ 0.055\\
\synthia{} \summertofog{} & 159.074 & 0.567 $\pm$ 0.082 & \textbf{151.934} & \textbf{0.566 $\pm$ 0.084}\\
\synthia{} \summertodawn{} & 171.494 & \textbf{0.628 $\pm$ 0.059} & \textbf{155.643} & 0.631 $\pm$ 0.076\\
\synthia{} \summertosunset{} & 154.975 & \textbf{0.601 $\pm$} 0.073 & \textbf{146.279} & 0.612 $\pm$ 0.074\\
\init{} \daytonight{} & \textbf{120.990} & \textbf{0.621 $\pm$} 0.075 & 149.868 & 0.679 $\pm$ 0.076\\\bottomrule
\end{tabular}
\caption{Comparison between the original CycleGAN formulation and our approach for full domain transformation ($p=1$). Both models have similar \FID{} and \LPIPS{} scores. The deviation from the mean measured on \LPIPS{} is also similar for both models.}
\label{tab:fid_lpips_vs_cyclegan}
\end{table}

\subsection{Mixing Transformations}
\label{supp:mixing}

In \autoref{fig:experimental_synthia_mixing} we also show generated images when all values of $p$ are nonzero, \ie, when asking to mix disjoint transformations.
The generator learns to \emph{mix} the two transformations, to some extent, generating images with a mixture of \textit{Night} and \textit{Winter} that are not part of the training sets (e.g., $p=[0.75, 0.75]$). 
In general, mixing styles can be considered a ``tug of war'' between the transformations, where each style ``pulls'' to bring the image closer to its target domain. Since multiple styles are doing this simultaneously, the result image shows several independent transformations that do not affect each other unless they alter the same area of the image (\eg adding snow on the side of the road and a night sky at the same time causes little conflict). The values of the different entries of $p$ determines the level of compromise to reach. When two styles collide (\eg two sky transformations) some artifacts appear in the affected region (see \autoref{fig:experimental_synthia_mixing} for $p=[0.5,0.75]$).
As expected, the higher the value of an individual element of $p$ is with respect to others, the more significant its presence. 
For example, for the $p=[0.25, 0.75]$ case in a \summertowinternight{} task, the amount of snow is fairly small and the image is significantly darkened. 
When $p_1$ increases, the night style loses its dominance and the images becomes slightly lighter and snowier, but nevertheless it still shows some darkness. 

\newlength{\synthiaMixingWidth}
\setlength{\synthiaMixingWidth}{0.18\linewidth}
\begin{figure}[h!]
    \centering
    \setlength{\tabcolsep}{2pt}
    \definecolor{Highlight}{RGB}{87,187,138}
    \begin{tabular}{c|ccccc}
    \diagbox[innerwidth=0.69cm]{\tiny{Night}}{\tiny{Winter}}& $p_1=0$ & $p_1=0.25$ & $p_1=0.5$ & $p_1=0.75$ & $p_1=1$ \\
    \midrule
    \rotatebox{90}{\hspace{4mm}$p_2=0$}
    &\cellcolor{Highlight}\includegraphics[width=\synthiaMixingWidth]{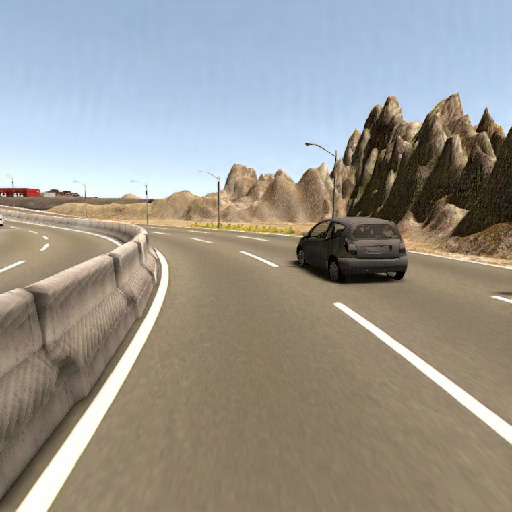}
    &\cellcolor{Highlight}\includegraphics[width=\synthiaMixingWidth]{images/datasets/summer2winter_night-4/06-000432-0.25-0.00}
    &\cellcolor{Highlight}\includegraphics[width=\synthiaMixingWidth]{images/datasets/summer2winter_night-4/06-000432-0.50-0.00}
    &\cellcolor{Highlight}\includegraphics[width=\synthiaMixingWidth]{images/datasets/summer2winter_night-4/06-000432-0.75-0.00}
    &\cellcolor{Highlight}\includegraphics[width=\synthiaMixingWidth]{images/datasets/summer2winter_night-4/06-000432-1.00-0.00}\\
    \rotatebox{90}{\hspace{4mm}$p_2=0.25$}
    &\cellcolor{Highlight}\includegraphics[width=\synthiaMixingWidth]{images/datasets/summer2winter_night-4/06-000432-0.00-0.25}
    &\includegraphics[width=\synthiaMixingWidth]{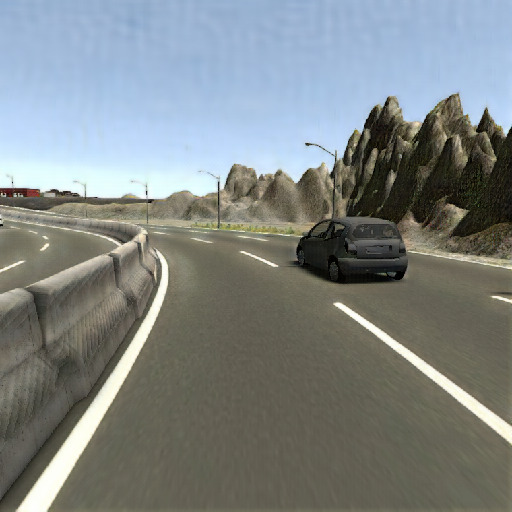}
    &\includegraphics[width=\synthiaMixingWidth]{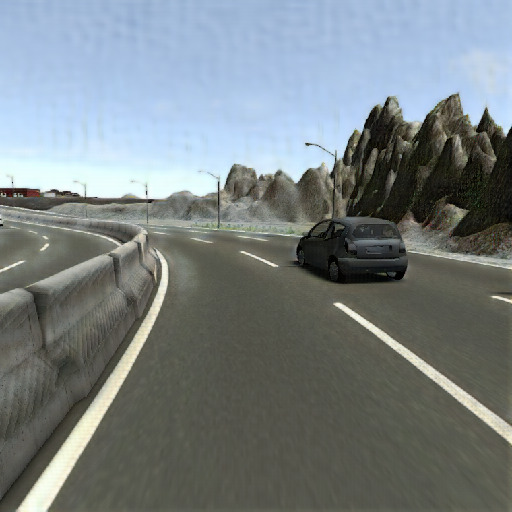}
    &\includegraphics[width=\synthiaMixingWidth]{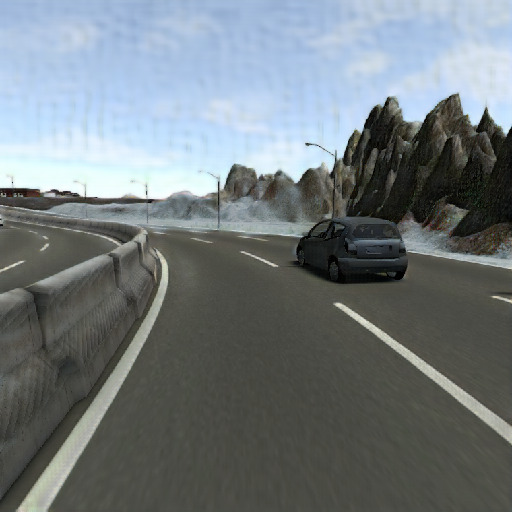}
    &\includegraphics[width=\synthiaMixingWidth]{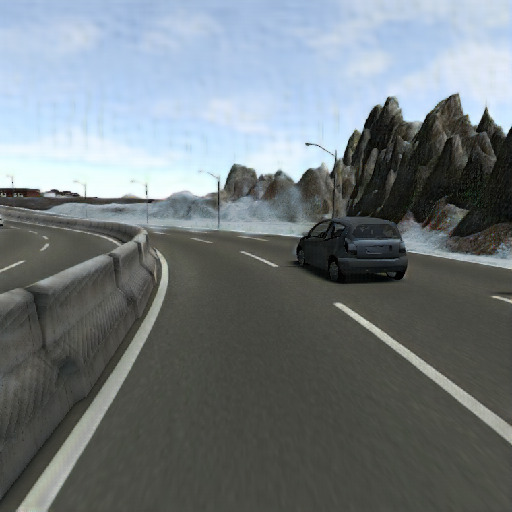}\\
    \rotatebox{90}{\hspace{4mm}$p_2=0.5$}
    &\cellcolor{Highlight}\includegraphics[width=\synthiaMixingWidth]{images/datasets/summer2winter_night-4/06-000432-0.00-0.50}
    &\includegraphics[width=\synthiaMixingWidth]{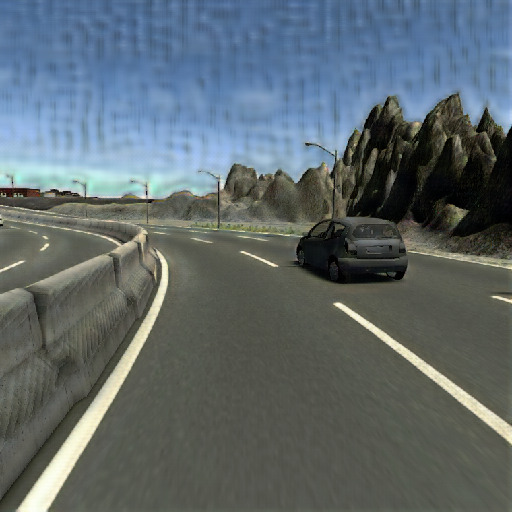}
    &\includegraphics[width=\synthiaMixingWidth]{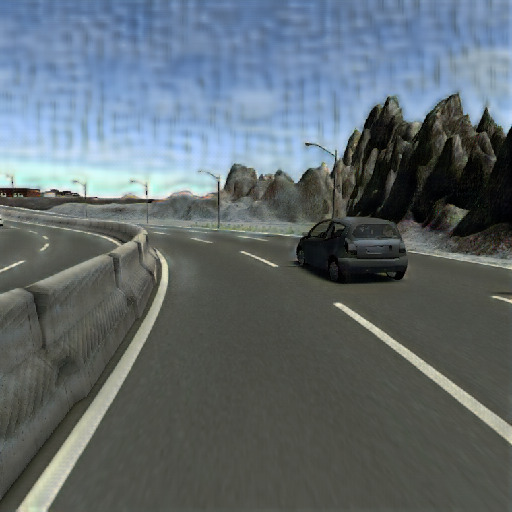}
    &\includegraphics[width=\synthiaMixingWidth]{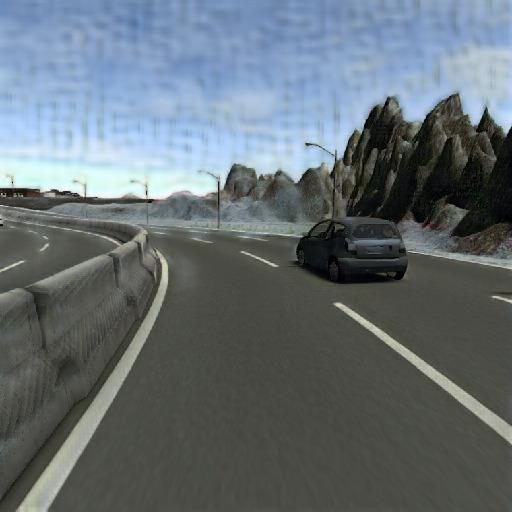}
    &\includegraphics[width=\synthiaMixingWidth]{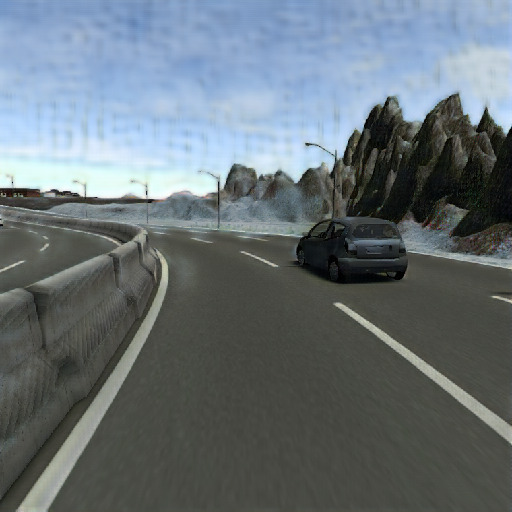}\\
    \rotatebox{90}{\hspace{4mm}$p_2=0.75$}
    &\cellcolor{Highlight}\includegraphics[width=\synthiaMixingWidth]{images/datasets/summer2winter_night-4/06-000432-0.00-0.75}
    &\includegraphics[width=\synthiaMixingWidth]{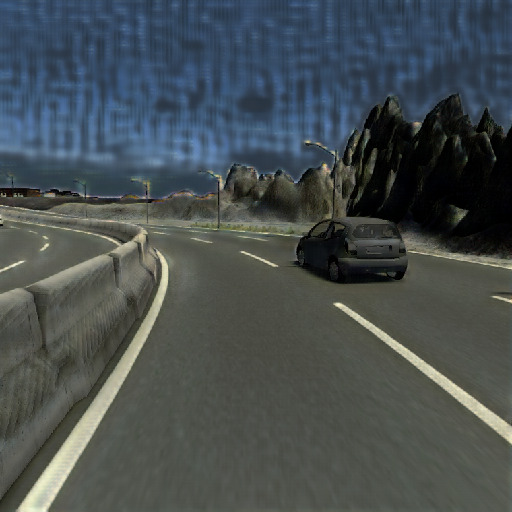}
    &\includegraphics[width=\synthiaMixingWidth]{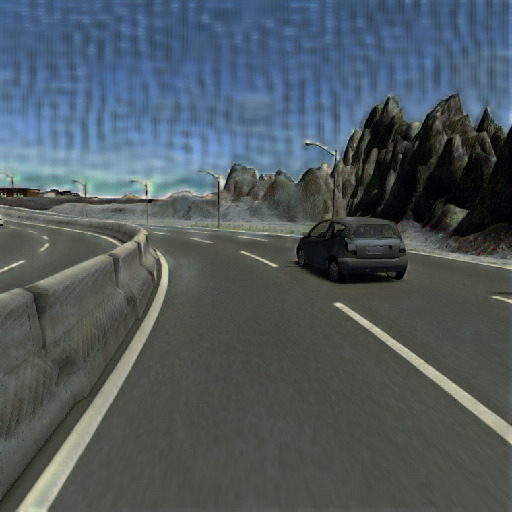}
    &\includegraphics[width=\synthiaMixingWidth]{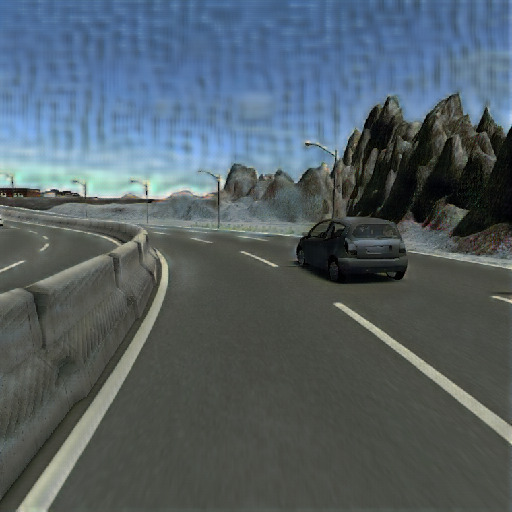}
    &\includegraphics[width=\synthiaMixingWidth]{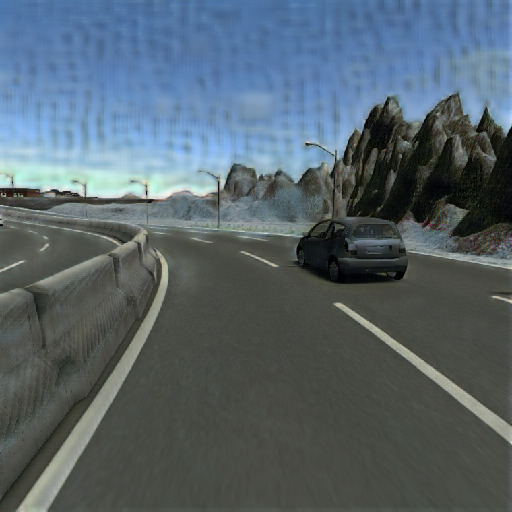}\\
    \rotatebox{90}{\hspace{4mm}$p_2=1$}
    &\cellcolor{Highlight}\includegraphics[width=\synthiaMixingWidth]{images/datasets/summer2winter_night-4/06-000432-0.00-1.00}
    &\includegraphics[width=\synthiaMixingWidth]{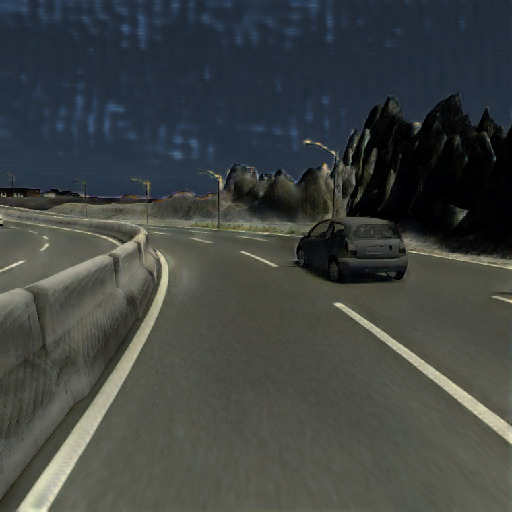}
    &\includegraphics[width=\synthiaMixingWidth]{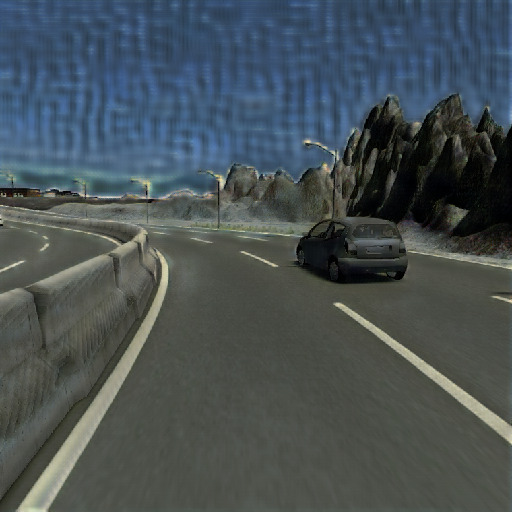}
    &\includegraphics[width=\synthiaMixingWidth]{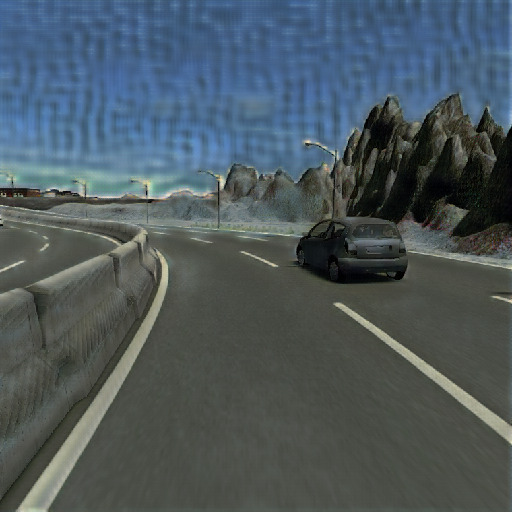}
    &\includegraphics[width=\synthiaMixingWidth]{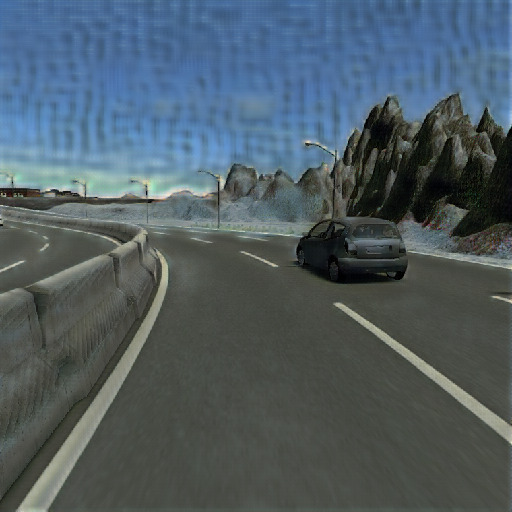}\\
    \end{tabular}
    \caption{Mixture of \textit{Winter} and \textit{Night} styles for \summertowinternight{}. The disentangled transformations (only one style applied) are {\color{Highlight} highlighted}.}
    \label{fig:experimental_synthia_mixing}
\end{figure}

\clearpage

\subsection{Feature-wise transformation} \label{supp:results_feature_wise}
An interesting effect of the interpolation is the fact that, contrary to a simple blending between two images, the network transforms the global style and different regions of the image at different steps (sky, terrain, snow, rain\dots). In \autoref{fig:feature_wise_transformations} we show this phenomenon for the \summertosoftrain{} and \summertowinter{} cases.

\newlength{\featureWiseLength}
\setlength{\featureWiseLength}{0.18\linewidth}
\begin{figure}[h!]
\centering
\begin{tabular}{ccccc}
    \setlength{\tabcolsep}{1pt}
    $p=0$ & $p=0.25$ & $p=0.5$ & $p=0.75$ & $p=1$ \\\toprule
    \includegraphics[width=\featureWiseLength]{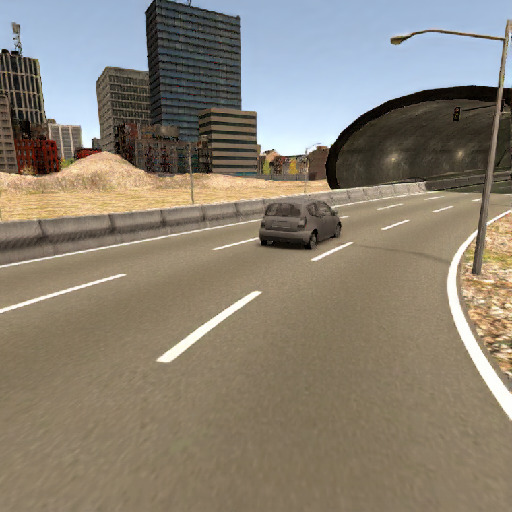} &
    \includegraphics[width=\featureWiseLength]{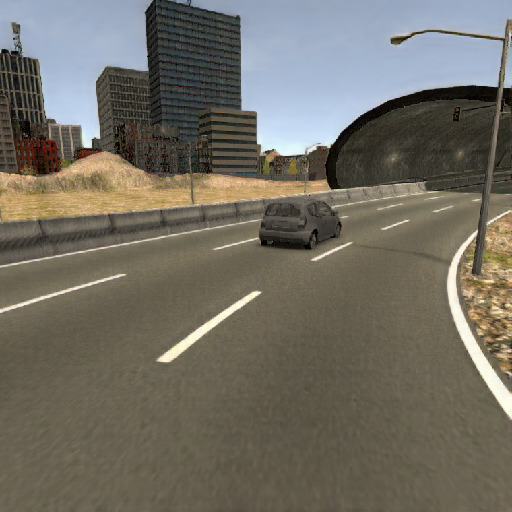} &
    \includegraphics[width=\featureWiseLength]{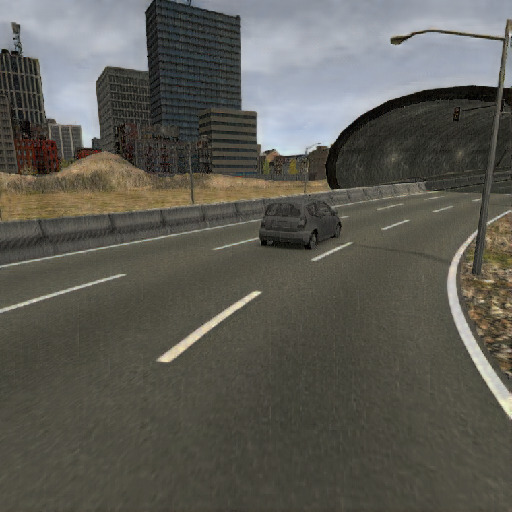} &
    \includegraphics[width=\featureWiseLength]{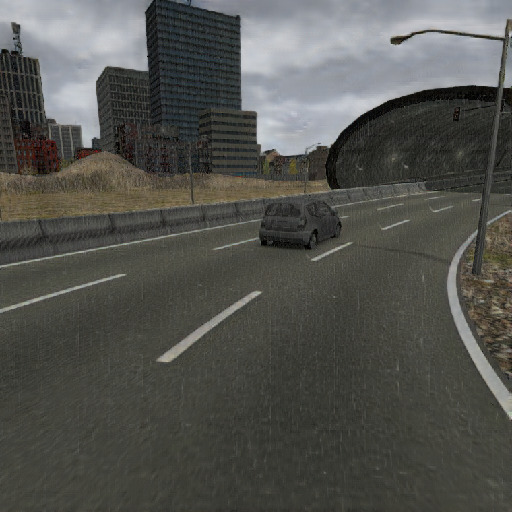} &
    \includegraphics[width=\featureWiseLength]{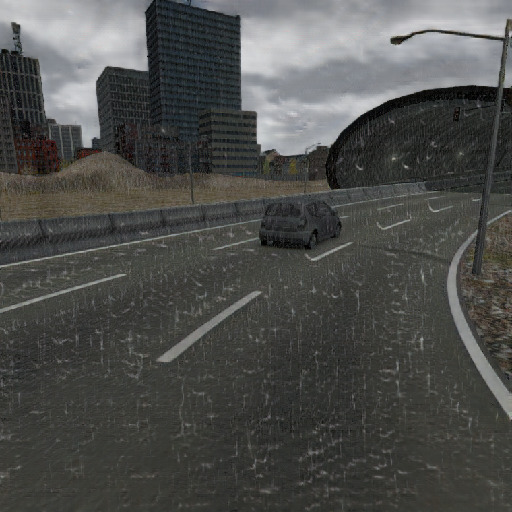} \\
    
    No changes & Luminosity & Clouds & Light rain & Heavy rain\\\midrule
    
    \includegraphics[width=\featureWiseLength]{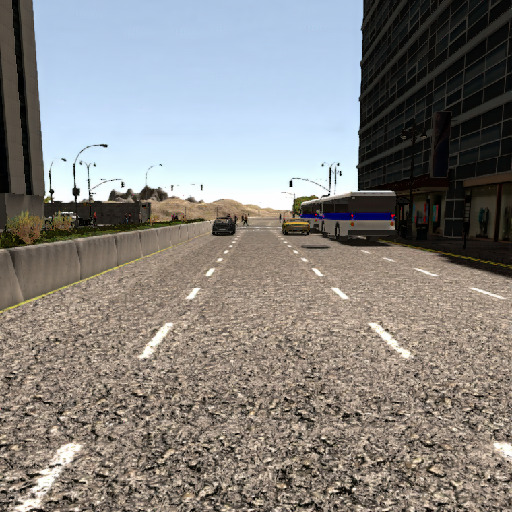} &
    \includegraphics[width=\featureWiseLength]{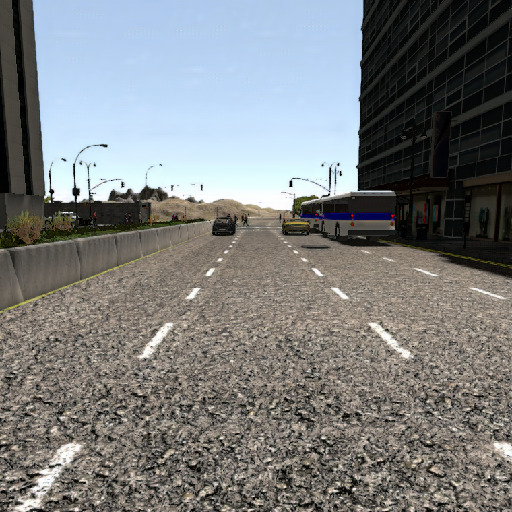} &
    \includegraphics[width=\featureWiseLength]{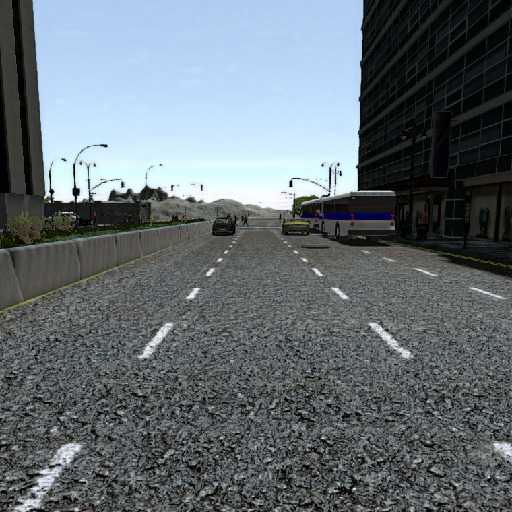} &
    \includegraphics[width=\featureWiseLength]{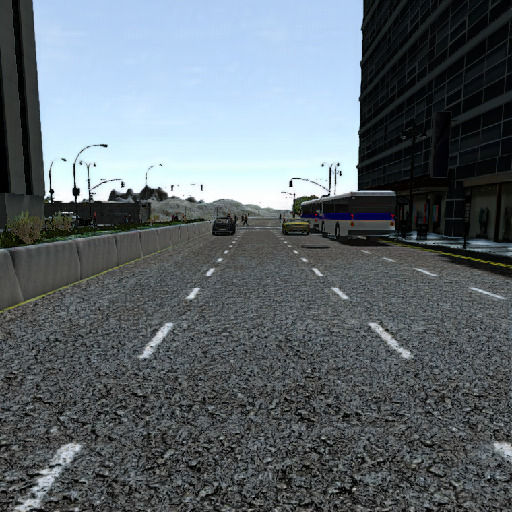} &
    \includegraphics[width=\featureWiseLength]{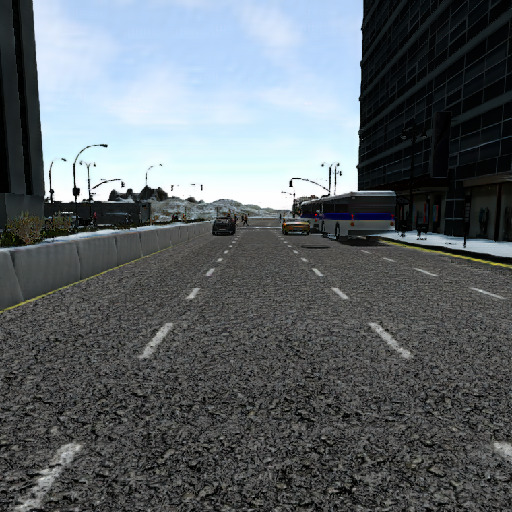} \\
    
    No changes & Luminosity & Clouds & Light snow & Clouds, heavy snow\\
\end{tabular}

\caption{Different changes at various stages of $p$. The network does not apply all aspects of the transformation at once, but rather it modifies progressively different characteristics of the input, see the caption under each image. \textbf{Top}: \summertosoftrain, \textbf{Bottom}: \summertowinter{}.}
\label{fig:feature_wise_transformations}
\end{figure}

\subsection{Latent space analysis}\label{supp:latent_space_analysis}

\begin{figure}
    \centering
    \includegraphics[width=.45\linewidth]{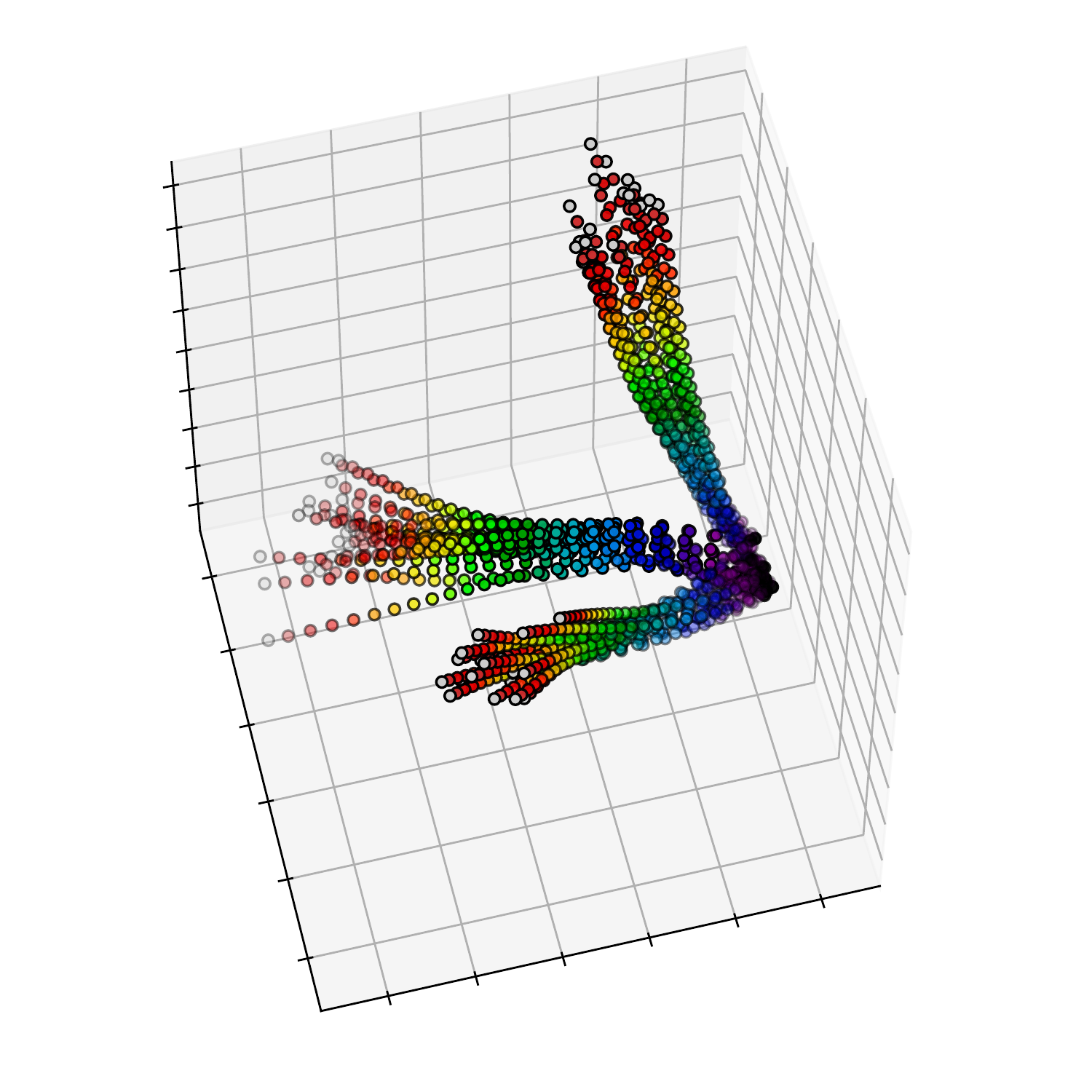}
    \includegraphics[width=.45\linewidth]{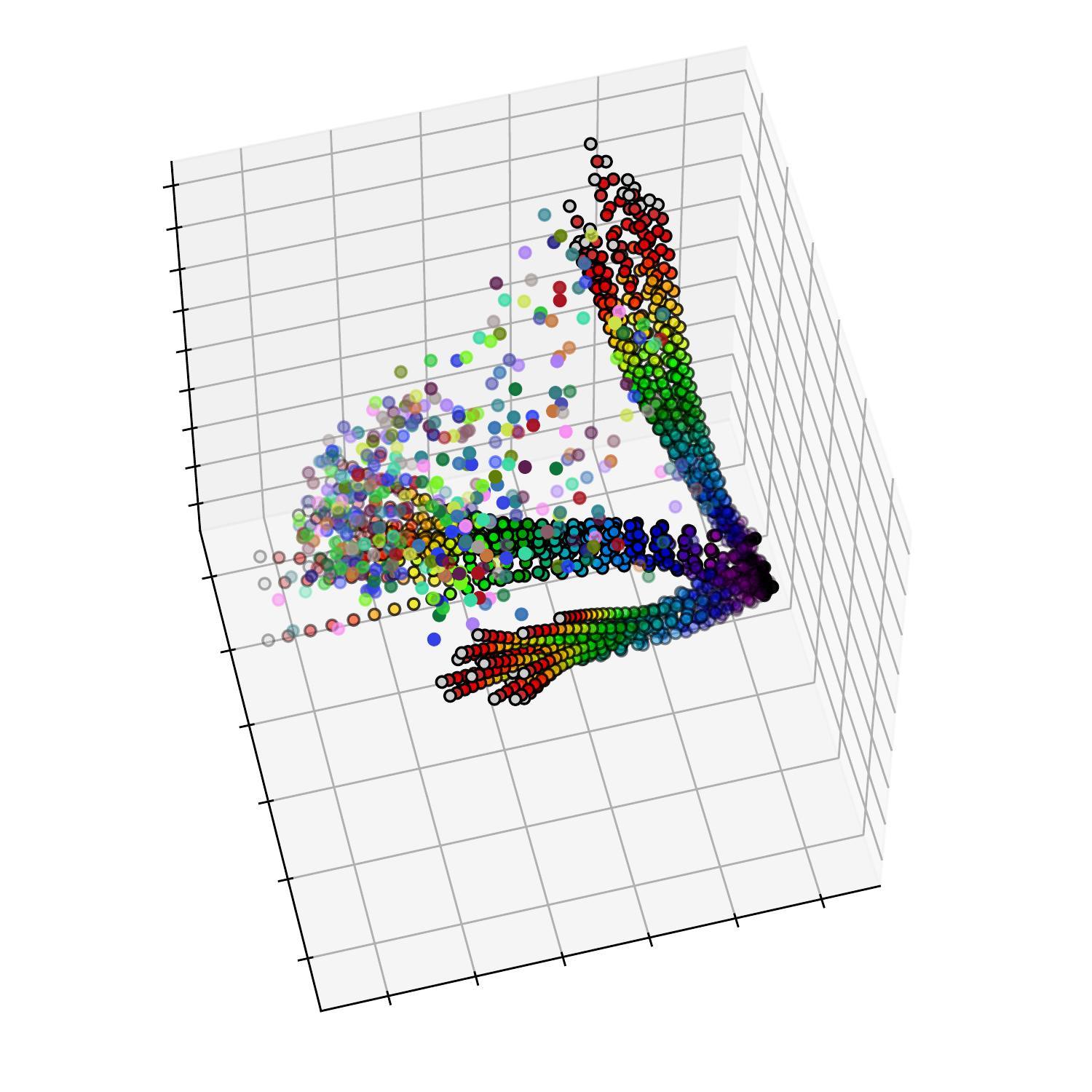}\\
    \caption{Visualizations of the latent space activations for the \summertowinternightsoftrain{} task. \textbf{Left}: Latent space activactions for disentangled transformations. \textbf{Right}: Same disentangled transformations plus entangled ones. The color code for each dimension of $p$  ranges from blue for $p=0$ to red ($p=1$). The transformations are laid out following three different trajectories (one per dimension of $p$, therefore one per style) and follow a clear trajectories away from $p=0$. The mixed vectors are all contained in the volume spanned by the disentangled trajectories.}
    \label{fig:latent_space_disentangled}
\end{figure}

As already mentioned, our method implicitly learns an interpolation between domains and a way of mixing them. 
We are interested in analyzing how this behavior reflects on the inner latent space learned by the network.
To this end, we collect the activations of the network right after the image and the conditioning vector information are mixed for different values of $p$ and plot them on a 3 dimensional space using PCA decomposition. 
In the left side of \autoref{fig:latent_space_disentangled} we show the results for \summertowinternightsoftrain{} when sampling different images from the same \synthia{} sequence and applying to each one disentangled values of $p$ ranging from $0$ to $1$.
Smooth changes in one of the parametrization dimensions make the learned image representations move smoothly in the latent space along well specified and pretty consistent trajectories. 
These results agree with the recent findings of \cite{DBLP:journals/corr/abs-1907-07171} that show how basic transformations are implicitly learned by generative models as trajectories on a latent space. 
We show that our model has similar properties while addressing an \imtoim{} task and providing an explicit control on how to move inside the learned trajectories, \ie, the parametrization vector.

On the right part of \autoref{fig:latent_space_disentangled} we add to the plot the latent space vectors obtained by mixing multiple transformations (e.g. $[0.25, 0.5, 1]$) to the same set of images. 
The newly added points falls within the three trajectories described by the disjoint transformations.
Therefore our model is able to remap the content of an image and the transformation to be applied in a shared latent space where individual domain to domain transformation remap to trajectories while mixing domains transformation remaps to points within these trajectories. 

\clearpage

\subsection{Domain generalization} \label{supp:domain_generalization}

\begin{figure}[h!]
    \centering
    \setlength{\tabcolsep}{2pt}
    \definecolor{Highlight}{RGB}{255,255,255}
    \begin{tabular}{c|ccccc}
    \diagbox[innerwidth=0.9cm]{\tiny{Rainy}}{\tiny{Night}}& $p_1=0$ & $p_1=0.25$ & $p_1=0.5$ & $p_1=0.75$ & $p_1=1$ \\
    \midrule
    \rotatebox{90}{\hspace{4mm}$p_2=0$}
    &\cellcolor{Highlight}\includegraphics[width=0.18\linewidth]{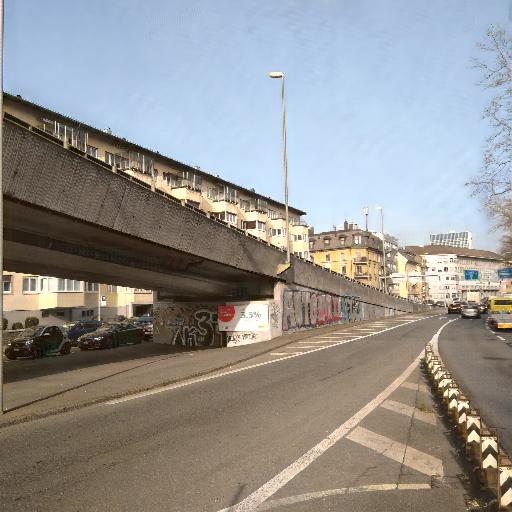}
    &\cellcolor{Highlight}\includegraphics[width=0.18\linewidth]{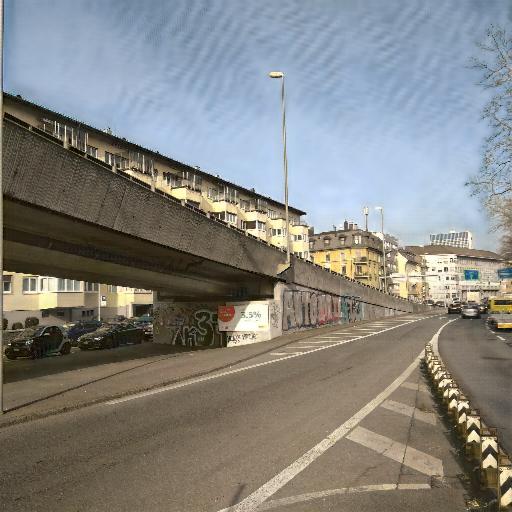}
    &\cellcolor{Highlight}\includegraphics[width=0.18\linewidth]{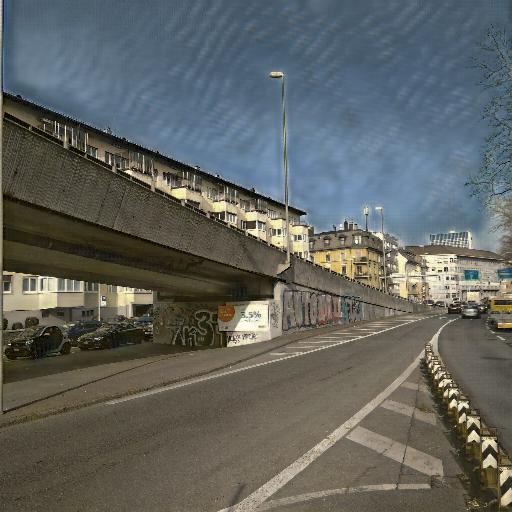}
    &\cellcolor{Highlight}\includegraphics[width=0.18\linewidth]{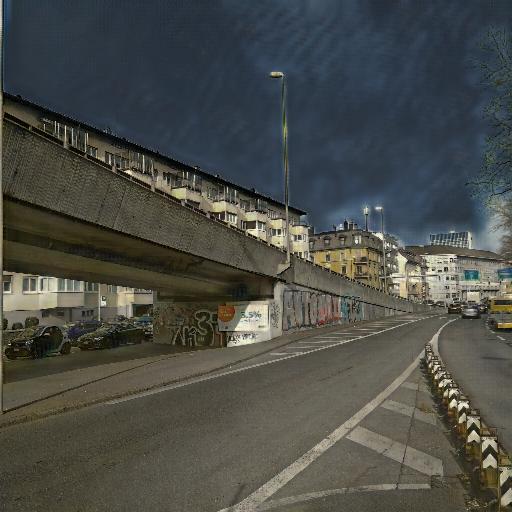}
    &\cellcolor{Highlight}\includegraphics[width=0.18\linewidth]{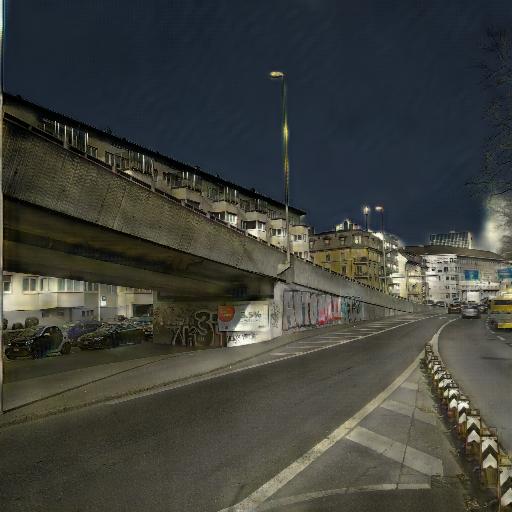}\\
    \rotatebox{90}{\hspace{4mm}$p_2=0.25$}
    &\cellcolor{Highlight}\includegraphics[width=0.18\linewidth]{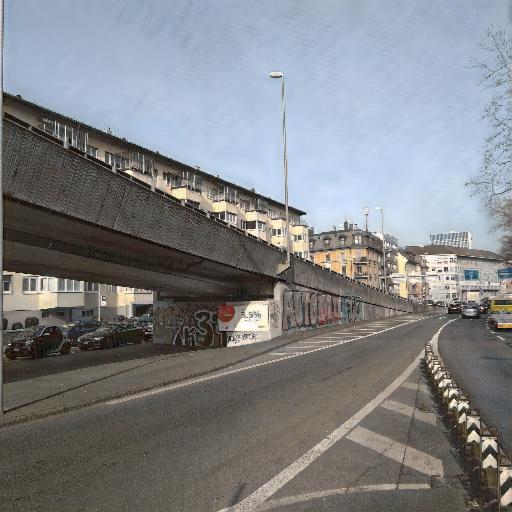}
    &\includegraphics[width=0.18\linewidth]{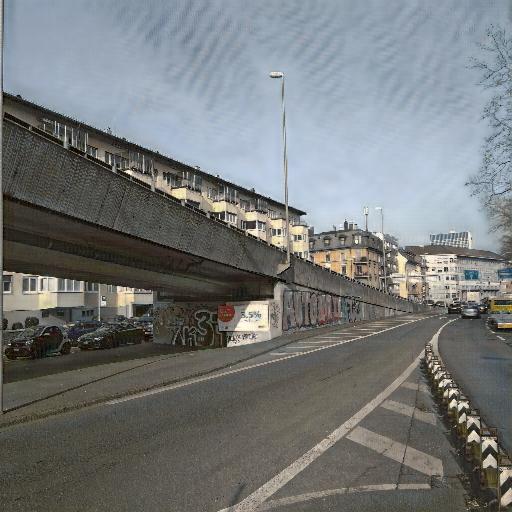}
    &\includegraphics[width=0.18\linewidth]{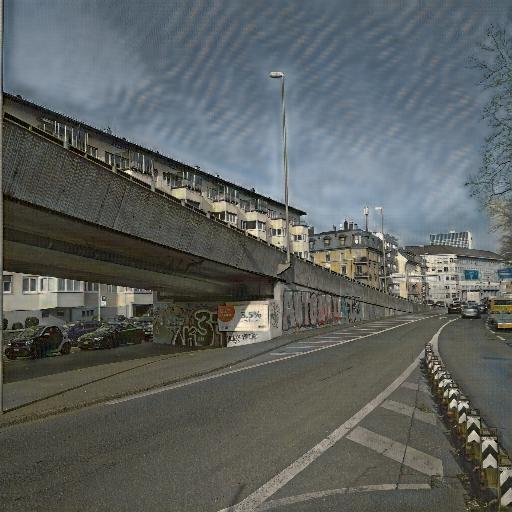}
    &\includegraphics[width=0.18\linewidth]{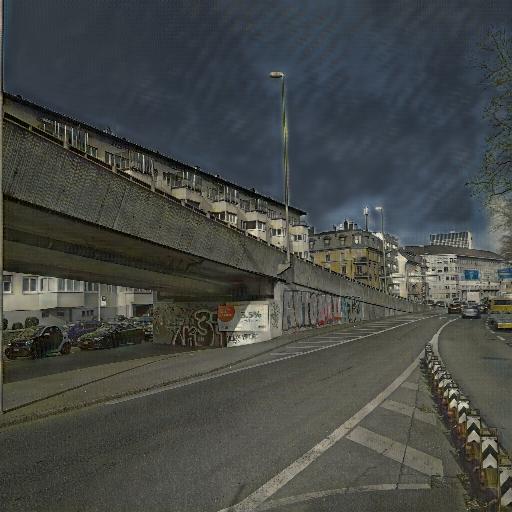}
    &\includegraphics[width=0.18\linewidth]{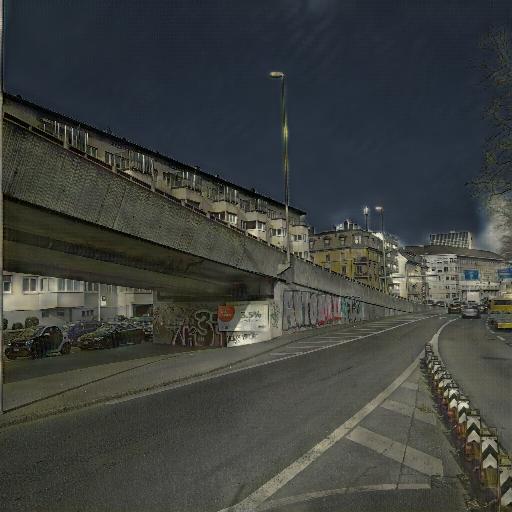}\\
    \rotatebox{90}{\hspace{4mm}$p_2=0.5$}
    &\cellcolor{Highlight}\includegraphics[width=0.18\linewidth]{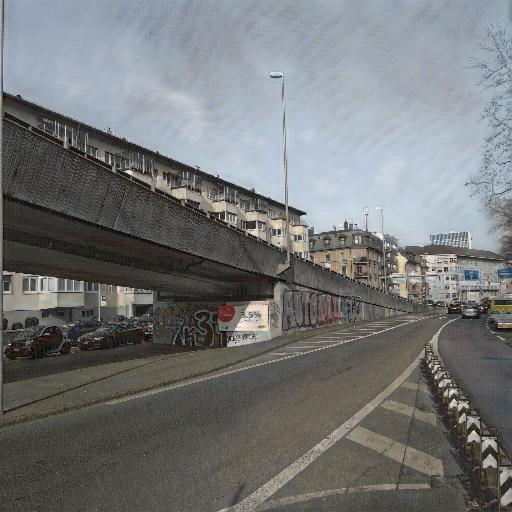}
    &\includegraphics[width=0.18\linewidth]{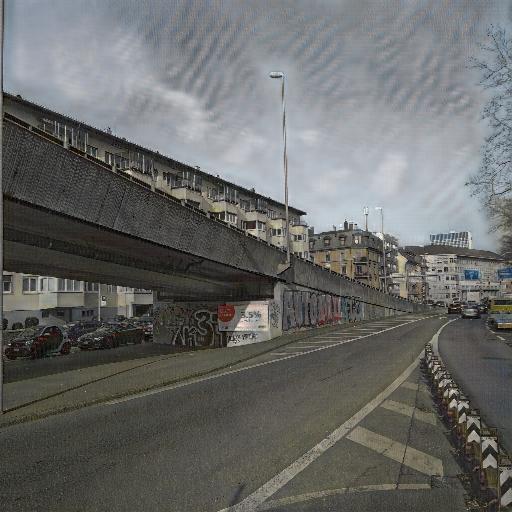}
    &\includegraphics[width=0.18\linewidth]{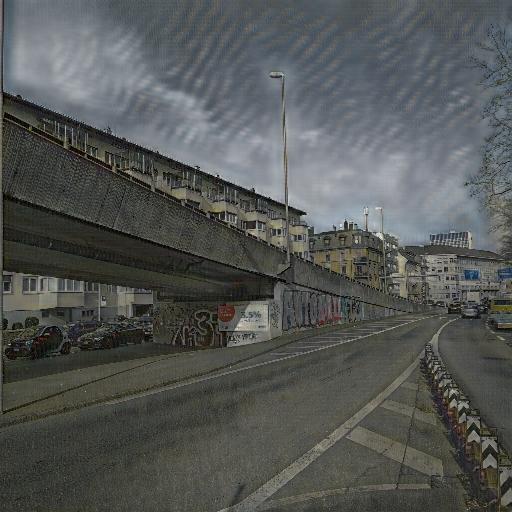}
    &\includegraphics[width=0.18\linewidth]{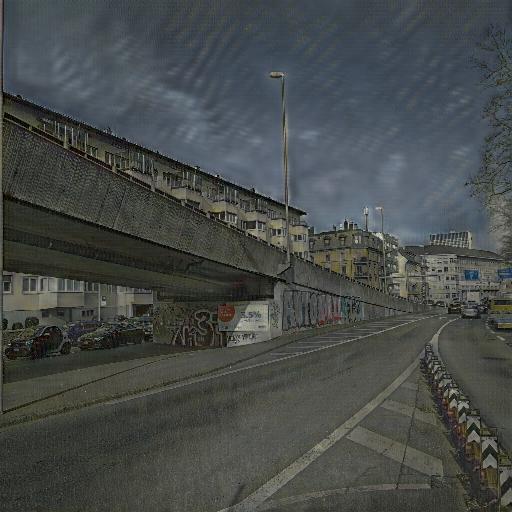}
    &\includegraphics[width=0.18\linewidth]{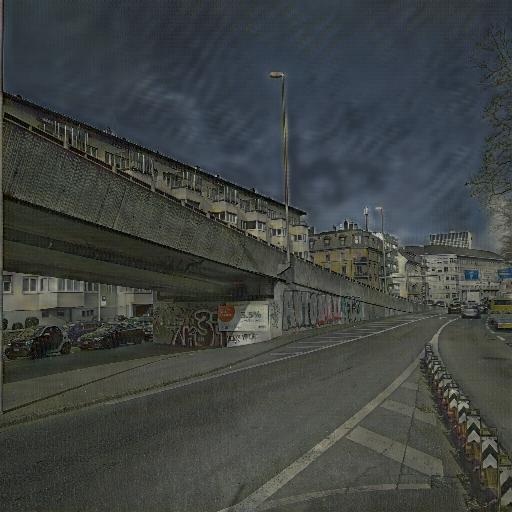}\\
    \rotatebox{90}{\hspace{4mm}$p_2=0.75$}
    &\cellcolor{Highlight}\includegraphics[width=0.18\linewidth]{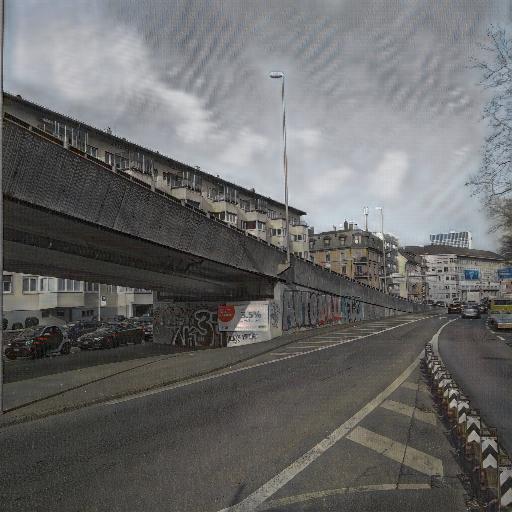}
    &\includegraphics[width=0.18\linewidth]{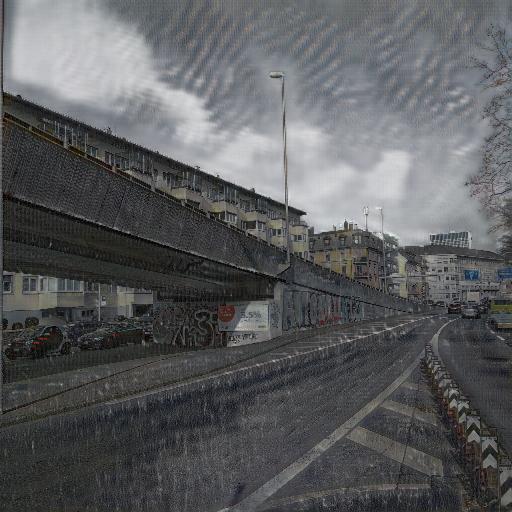}
    &\includegraphics[width=0.18\linewidth]{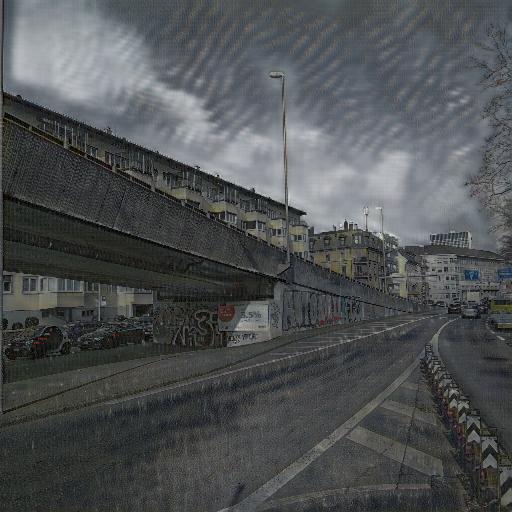}
    &\includegraphics[width=0.18\linewidth]{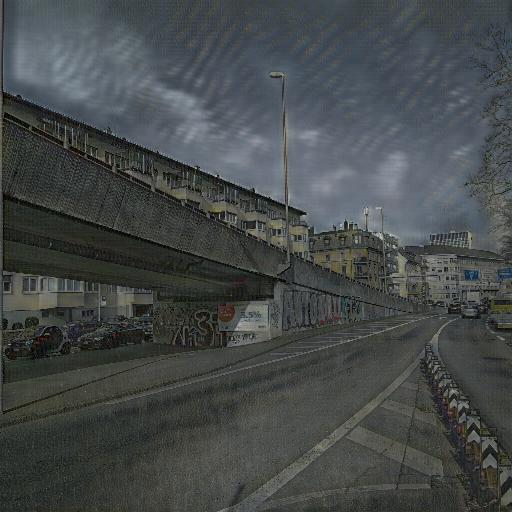}
    &\includegraphics[width=0.18\linewidth]{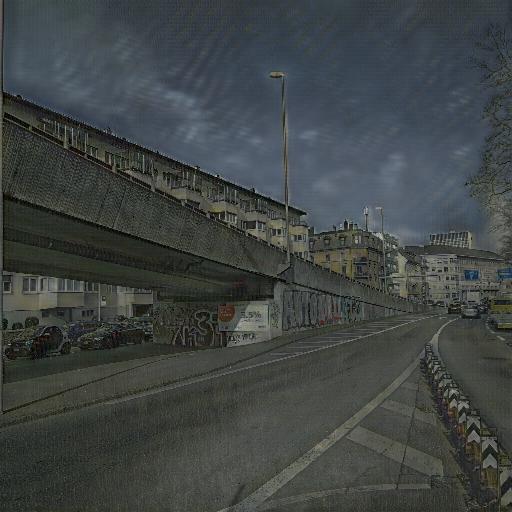}\\
    \rotatebox{90}{\hspace{4mm}$p_2=1$}
    &\cellcolor{Highlight}\includegraphics[width=0.18\linewidth]{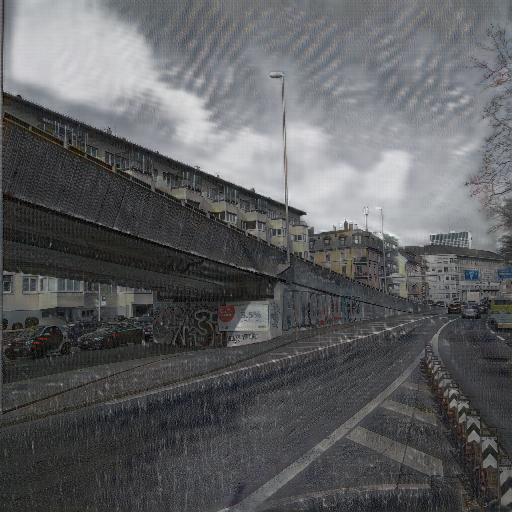}
    &\includegraphics[width=0.18\linewidth]{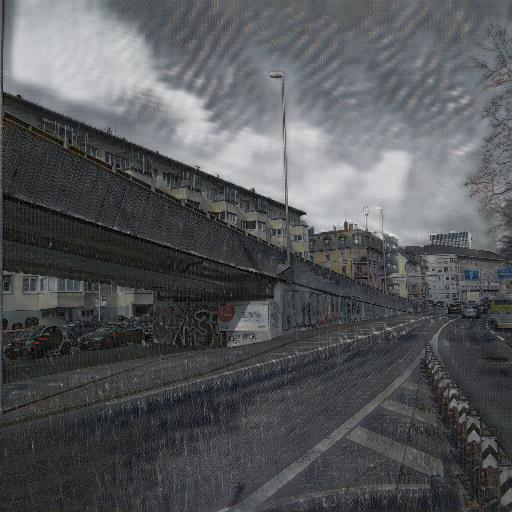}
    &\includegraphics[width=0.18\linewidth]{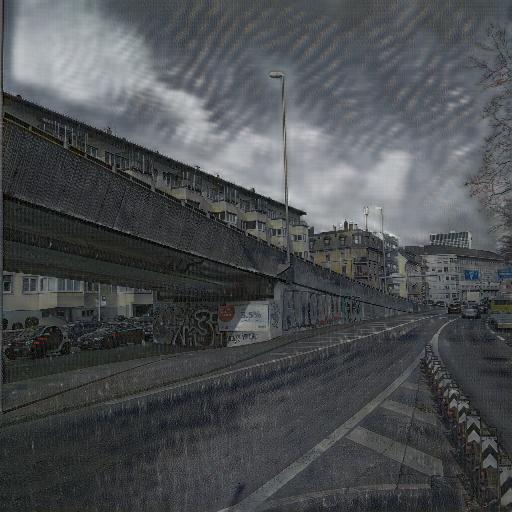}
    &\includegraphics[width=0.18\linewidth]{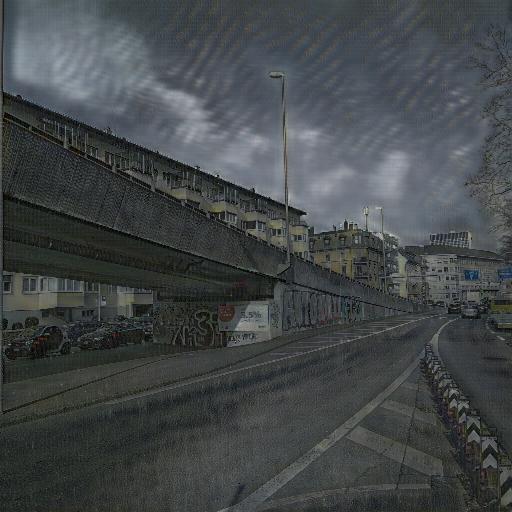}
    &\includegraphics[width=0.18\linewidth]{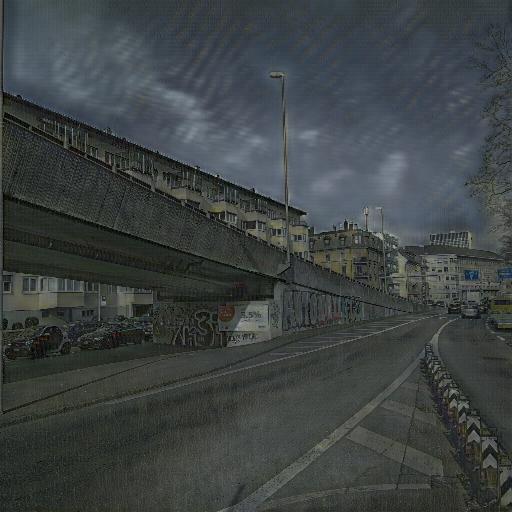}\\
    \end{tabular}
    \caption{Evaluation of the network trained on \synthia{} on real-world images.}
    \label{fig:manessestrasse_on_synthia}
\end{figure}

We also briefly explore the domain generalization capabilities of this method using real-world datasets for evaluation. In \autoref{fig:manessestrasse_on_synthia} we evaluate a pre-trained network for the \textit{summer$\rightarrow{}$night+rain} task trained with images from the \synthia{} dataset on real images acquired by a phone. The network correctly segments the different components of the image (sky, buildings) and adds the appropriate effects (clouds, lights).

\clearpage

\subsection{Implementation details} \label{supp:implementation_details}

Our model for soft-parameterization extends CycleGAN with a series of fully connected layers that transform the conditioning vector by upsampling it to a 64-dimensional vector. This is replicated and concatenated to the feature dimension of the output of the last ResNet block. In the \bookCovers{} case, we additionally use skip connections between the encoder and decoder parts of the generator, to which we also concatenate the upsampled conditioning vector, as well as multi-scale discriminators. We train for all experiments using a batch size of 1 and an image size of $512\times512$.
We implement our framework on TensorFlow 1 \cite{tensorflow2015-whitepaper} following the architecture of CycleGAN \cite{CycleGAN2017}\footnote{ \href{https://github.com/tensorflow/gan/tree/master/tensorflow\_gan/examples/cyclegan}{https://github.com/tensorflow/gan/tree/master/tensorflow\_gan/examples/cyclegan}.} with the additions already discussed in \autoref{ssec:architecture}. 
We use the Adam \cite{DBLP:journals/corr/KingmaB14} optimizer in all our experiments with a learning rate of 0.0001 and $\beta_1=0.5$.
Following the guidelines of the original CycleGAN architecture, we set $\lambda=10$ and use a batch size of 1.
The number of training iterations varies between different tests but roughly corresponds to $200$ epochs.

\subsection{Alternatives for the concatenation of \texorpdfstring{$p$}{\textit{p}}} \label{supp:concatenation}
The conditioning vector $p$ is transformed through a series of fully connected layers and finally concatenated at a certain step of the generators and discriminators. 
As additional ablation studies we analyze the effect of injecting it at different levels of the architectures and report the results in \autoref{supp:ablation}.
We train multiple variants of our architecture for the \summertowinter{} tasks and compare the different alternatives by measuring \FID{} with respect to a random source images (when conditioning with $p=0$) and with respect to a random target image (when conditioning with $p=1$). 
The performances of all the alternatives are fairly close, but injecting $p$ just once for both the generator and discriminator has the advantage of being the simpler strategy and providing the best performance. 
For this reason we selected this option and presented it in the main paper.

\begin{table}[]
\centering
\begin{tabular}{l|c|c|ll}
\toprule
& \multicolumn{2}{c|}{$p$ concatenation} & \multicolumn{2}{c}{FID} \\
\midrule
& Generator & Discriminator & vs source $p=0$ &  vs target $p=1$ \\
\midrule
Ablation 1 & once & $\forall$ conv. & 29.754 & 111.602 \\
Ablation 2 & $\forall$ encoder conv. & once & 13.285 & 113.442 \\ 
Ablation 3 & $\forall$ decoder conv. & once & 13.849 & 109.607 \\
Ablation 4 & $\forall$ encoder conv. & $\forall$ conv. & 16.217 & 116.461\\
Ablation 5 & $\forall$ decoder conv & $\forall$ conv. & 13.666  & 112.598 \\ 
\midrule
\textbf{ParGAN} & once & once & \textbf{13.183} & \textbf{109.25} \\
\bottomrule                  
\end{tabular}
\caption{Different configurations on how $p$ is included in the generator and discriminator and its effects on quality. We use the same measurement strategy as in \autoref{tab:fid_lpips_multiplesynthia}.}
\label{supp:ablation}
\end{table}

\end{document}